%% file: codemode_paper.tex
\documentclass[11pt]{article}

\usepackage{fancyvrb}
\usepackage[preprint]{emnlp}
\usepackage{xcolor}
\DefineVerbatimEnvironment{PromptVerb}{Verbatim}{%
  fontsize=\scriptsize,%
  xleftmargin=0pt,%
  samepage=false,%
}
\usepackage[utf8]{inputenc}
\usepackage[T1]{fontenc}
\usepackage{booktabs}
\usepackage{amsfonts}
\usepackage{amsmath}
\usepackage{microtype}
\usepackage{graphicx}
\usepackage{algorithm}
\usepackage{algorithmic}
\usepackage{tikz}
\usetikzlibrary{arrows.meta,positioning,fit,calc}

\DeclareRobustCommand*\circled[1]{\raisebox{.5pt}{\textcircled{\raisebox{-.9pt}{\scriptsize\bfseries #1}}}}
\DeclareRobustCommand*\circledw[1]{%
  \begin{tikzpicture}[baseline=(circ.base)]%
    \node[draw,circle,inner sep=0pt,minimum size=1.4em,line width=0.4pt,%
          font=\scriptsize\bfseries] (circ) {#1};%
  \end{tikzpicture}%
}

\title{RubricRefine:\\
Improving Tool-Use Agent Reliability with Training-Free Pre-Execution Refinement}

\author{Will LeVine, \, Brendan Raffel Evers, \, Sam Saltwick, \, Abhay Venkatesh \\
  {\normalfont Anduril Industries}}

\begin{document}
\maketitle

\begin{center}
\begin{minipage}{0.86\columnwidth}
\small\itshape
``People who have no hold over their process of thinking are likely to be ruined by liberty of thought. If thought is immature, liberty of thought becomes a method of converting men into animals.''
\upshape\raggedleft\\[2pt]--- Muhammad Iqbal
\end{minipage}
\end{center}
\vspace{0.5em}

\begin{abstract}

Iterative self-refinement is a popular inference-time reliability technique, but its effectiveness in code-mode tool use depends heavily on the structure of the feedback signal: unstructured critique helps inconsistently across models, and even revision with real execution feedback improves only modestly. The dominant failures are inter-tool contract violations (wrong output shape, incorrect tool routing, broken argument provenance) that run to completion without raising errors, making runtime feedback insufficient. We introduce \textbf{RubricRefine},\footnote{Full prompt templates and evaluation code: \url{https://anonymous.4open.science/r/RubricRefinePrivate-89FD/}} a training-free method for pre-execution \emph{contract checking} that generates task- and registry-specific rubrics, scores candidate code against explicit contract checks, and iteratively repairs failures before any execution occurs. RubricRefine reaches $0.86$, averaged across seven models, on M3ToolEval with zero execution attempts, improving over prior inference-time baselines at lower latency than rubric-guided reranking. Performance remains flat on the predominantly single-step API-Bank, consistent with the method's reliance on inter-tool contract structure. Results on AppWorld demonstrate that our method maintains an advantage in the multi-turn setting. Because the rubric is derived from the supplied tool documentation, the method's advantage survives \emph{incomplete} documentation but reverses under \emph{incorrect} documentation. A rubric-category ablation identifies which rules are load-bearing, and top-bin calibration enables early stopping even where aggregate calibration is poor.\end{abstract}

\section{Introduction}
\label{sec:intro}

Code-mode tool-use agents act by emitting executable programs rather than flat JSON function calls, a design that lets one turn express multiple tool calls, branching, intermediate state, and output formatting inside a single executable trace \citep{roucher2024smolagents,cloudflare2025codemode,wang2024executable}. This expressivity shifts the dominant failure mode away from action-format errors and toward inter-tool \emph{contract} failures (wrong output shape, broken argument provenance, incorrect tool routing) that run to completion without raising exceptions and are invisible to execution-based feedback. Existing reliability methods either require post-training for the target tool registry (e.g., CodeActAgent-style setups \citep{wang2024executable}) or depend on observed execution outcomes, leaving no deployment-time path to catch these contract failures before the first live action. This gap is operationally consequential: a failed attempt may already have altered external state in ways that are only partially observable, and execution is often rate-limited, paid, or safety constrained. This motivates the \textbf{single-attempt} setting, where all checking and repair must happen before a task's one allowed live action. What is missing is a deployment-time method that adapts to a new registry and improves execution readiness without any training or execution.

The question is what form that pre-execution control signal should take. Prior work has shown that unstructured iterative self-revision without external grounding is unreliable \citep{huang2024selfcorrect,snell2024scaling}, and our results confirm this for Self-Refine \citep{madaan2023selfrefine}: it underperforms single-pass generation on five of the seven models we test, helping only on the two open-weight models. The loop is not the problem; the feedback quality is. Free-form critique, lacking a structured account of what failed and why, does not reliably diagnose contract violations in code-mode programs. Rubrics supply exactly what is missing: decomposing task correctness into individually checkable criteria yields feedback that is both \emph{structured} (criteria rather than free-form critique) and, when generated per instance, \emph{task-specific} --- the two properties iterative repair needs. Section~\ref{sec:related} situates both against prior rubric work.

We introduce \textbf{RubricRefine}, a method for pre-execution contract checking in multi-step code-mode agents that generates a registry-conditioned rubric, scores candidate code against explicit contract checks, and iteratively repairs failures before runtime.

Our central contribution is \textbf{pre-execution contract checking} as a reliability primitive for code-mode agents: a characterization of a failure class (silent contract violations invisible to execution feedback) and an operationalization (contract-structured instance-specific rubrics) that detects many of these failures before any live action. On \textbf{M3ToolEval}~\citep{wang2024executable}, RubricRefine improves over prior inference-time baselines by $+0.14$ to $+0.38$ across seven models; on the single-step \textbf{API-Bank}~\citep{li2023apibank} it is flat, because the method helps in proportion to the inter-tool contract structure a task contains. Stress-testing documentation quality (Section~\ref{sec:doc-robustness}), the gain survives truncated descriptions but reverses when documented parameter names are wrong. On the multi-turn \textbf{AppWorld}~\citep{trivedi2024appworld}, RubricRefine beats CodeAct at every interaction budget tested; the gap narrows as turns grow plentiful but does not close, mirroring the Self-Debug result at a different scale (Section~\ref{sec:appworld}). Additional analyses show top-bin calibration enables early stopping even with poor aggregate calibration, and a rubric-category ablation reveals which rules are load-bearing.

\section{Related Work}
\label{sec:related}

\paragraph{Inference-Time Self-Correction}
Self-Refine \citep{madaan2023selfrefine} uses the same generate-critique-revise loop but with unstructured critique, which prior work has shown to be unreliable without external grounding \citep{huang2024selfcorrect,snell2024scaling}. Self-Debug~\citep{chen2023selfdebug} and Reflexion~\citep{shinn2023reflexion} revise iteratively from observed execution outcomes; we evaluate Self-Debug with real execution feedback as a direct baseline (Section~\ref{sec:main-results}). Test-execution methods, and additional execution-feedback and learned-verifier work, are discussed in Appendix~\ref{app:rw-main}.

\paragraph{Rubric-Based Evaluation}
RubricRefine inherits the rubric-based judging perspective of LLM-as-a-Judge and related rubric-conditioned evaluators \citep{zheng2023judging,kim2024prometheus}, but uses it for iterative repair rather than post hoc scoring. Two especially relevant precedents are Rubrics-as-Rewards, which uses sample-dependent rubrics generated from human-written reference answers for reinforcement-learning post-training \citep{gunjal2025rubricsrewards}, and Agentic Rubrics, which generates rubrics from pull-request descriptions to evaluate code changes in software-engineering tasks \citep{raghavendra2026agenticrubrics}. REBEL~\citep{levine2025rebel} applies sample-dependent rubric scoring to document reranking rather than iterative code repair. RubricRefine adopts the same sample-dependent rubric intuition, but conditions the rubric on the current task prompt and tool registry rather than on a pull-request workflow or a human-written reference answer and uses these signals for online refinement.

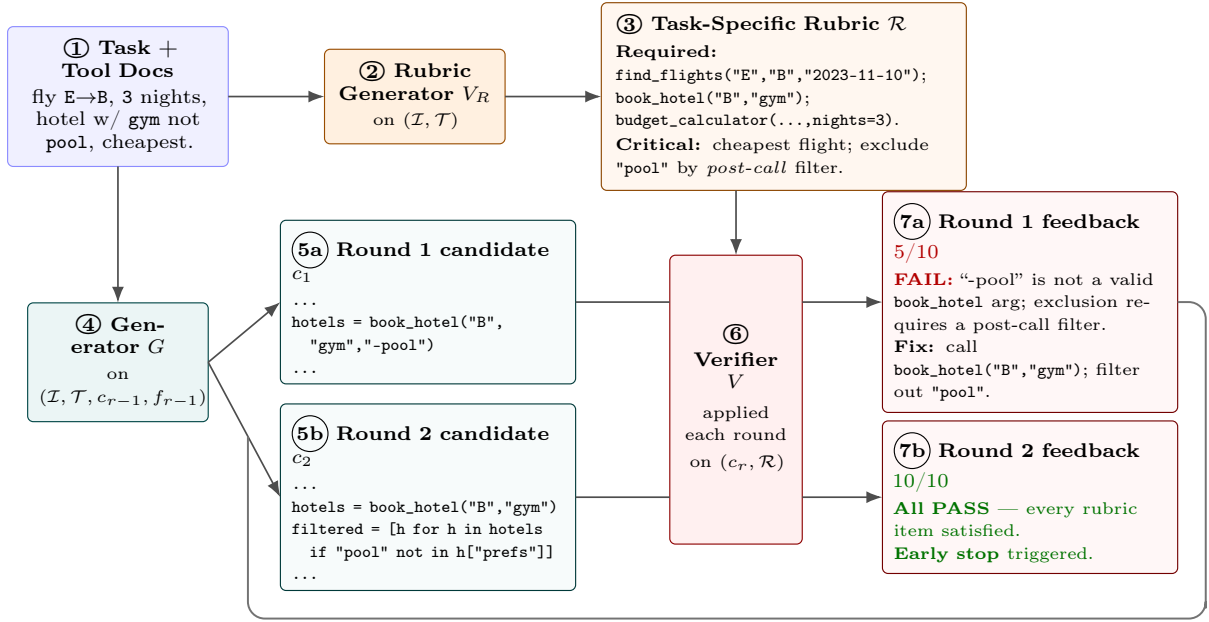
\begin{figure*}[t]
\centering
\resizebox{\textwidth}{!}{%
\begin{tikzpicture}[
    font=\scriptsize,
    >=Latex,
    node distance=1.3cm,
    box/.style={
        draw,
        rounded corners=2pt,
        align=center,
        inner sep=5pt,
        outer sep=0pt,
        line width=0.4pt,
        minimum height=0.9cm
    },
    taskbox/.style={box, fill=blue!6, draw=blue!45, text width=2.4cm},
    rgenbox/.style={box, fill=orange!12, draw=orange!60!black, text width=1.9cm},
    rubricbox/.style={box, fill=orange!6, draw=orange!55!black, text width=4.2cm, align=left, inner sep=5pt, minimum height=1.4cm},
    genbox/.style={box, fill=teal!8, draw=teal!55!black, text width=1.9cm},
    candbox/.style={box, fill=teal!3, draw=teal!45!black, text width=3.4cm, align=left, inner sep=4pt},
    verifbox/.style={box, fill=red!6, draw=red!55!black, text width=1.9cm},
    fbbox/.style={box, fill=red!3, draw=red!45!black, text width=3.4cm, align=left, inner sep=4pt},
    flow/.style={->, line width=0.6pt, draw=black!70},
    looparrow/.style={line width=0.7pt, draw=black!55, rounded corners=6pt},
    lbl/.style={font=\tiny\itshape, text=black!65},
]

\node[taskbox] (task) at (0,0) {%
{\bfseries\circled{1} Task + Tool Docs}\\[0.15em]
{\scriptsize fly \texttt{E}$\to$\texttt{B}, \texttt{3} nights, hotel w/ \texttt{gym} not \texttt{pool}, cheapest.}
};

\node[rgenbox, right=1.2cm of task] (rgen) {%
{\bfseries\circled{2} Rubric Generator $V_R$}\\[0.15em]
{\tiny on $(\mathcal{I},\mathcal{T})$}
};

\node[rubricbox, right=1.2cm of rgen] (rubric) {%
{\bfseries\scriptsize\circled{3} Task-Specific Rubric $\mathcal{R}$}\\[0.2em]
{\tiny\textbf{Required:} \texttt{find\_flights("E","B","2023-11-10")}; \texttt{book\_hotel("B","gym")}; \texttt{budget\_calculator(...,nights=3)}.}\\[0.1em]
{\tiny\textbf{Critical:} cheapest flight; exclude \texttt{"pool"} by \emph{post-call} filter.}
};

\node[genbox, below=1.7cm of task] (gen) {%
{\bfseries\circled{4} Generator $G$}\\[0.15em]
{\tiny on $(\mathcal{I},\mathcal{T}, c_{r-1}, f_{r-1})$}
};

\node[candbox, right=0.9cm of gen, yshift=0.75cm] (cand1) {%
{\bfseries\scriptsize\circledw{5a} Round 1 candidate $c_1$}\\[0.15em]
{\ttfamily\tiny \textellipsis}\\
{\ttfamily\tiny hotels = book\_hotel("B",}\\
{\ttfamily\tiny\hspace*{1em}"gym",\textbf{"-pool"})}\\
{\ttfamily\tiny \textellipsis}
};

\node[candbox, below=0.25cm of cand1] (cand2) {%
{\bfseries\scriptsize\circledw{5b} Round 2 candidate $c_2$}\\[0.15em]
{\ttfamily\tiny \textellipsis}\\
{\ttfamily\tiny hotels = book\_hotel("B","gym")}\\
{\ttfamily\tiny filtered = [h for h in hotels}\\
{\ttfamily\tiny\hspace*{1em}if "pool" not in h["prefs"]]}\\
{\ttfamily\tiny \textellipsis}
};

\node[box, fill=red!6, draw=red!55!black, text width=1.3cm, align=center,
      minimum height=3.6cm]
      (verif) at ($(cand1.east)!0.5!(cand2.east) + (2.0cm, 0)$) {%
{\bfseries\circled{6}}\\[0.15em]
{\bfseries\scriptsize Verifier $V$}\\[0.3em]
{\tiny applied}\\
{\tiny each round}\\[0.2em]
{\tiny on $(c_r,\mathcal{R})$}
};

\coordinate (fbx) at ($(verif.east) + (1.0cm, 0)$);
\node[fbbox, anchor=west] (fb1) at (fbx |- cand1) {%
{\bfseries\scriptsize\circledw{7a} Round 1 feedback}\hfill{\scriptsize\color{red!70!black} 5/10}\\[0.2em]
{\tiny\textbf{\color{red!70!black}FAIL:} ``-pool'' is not a valid \texttt{book\_hotel} arg; exclusion requires a post-call filter.}\\
{\tiny\textbf{Fix:} call \texttt{book\_hotel("B","gym")}; filter out \texttt{"pool"}.}
};

\node[fbbox, anchor=west] (fb2) at (fbx |- cand2) {%
{\bfseries\scriptsize\circledw{7b} Round 2 feedback}\hfill{\scriptsize\color{green!45!black} 10/10}\\[0.2em]
{\tiny\color{green!45!black}\textbf{All PASS} --- every rubric item satisfied.}\\
{\tiny\color{green!45!black}\textbf{Early stop} triggered.}
};

\draw[flow] (task) -- (rgen);
\draw[flow] (rgen) -- (rubric);
\draw[flow] (task.south) -- (gen.north);
\draw[flow] (verif.north |- rubric.south) -- (verif.north);
\draw[flow] (gen.east) -- (cand1.west);
\draw[flow] (gen.east) -- (cand2.west);
\draw[line width=0.6pt, draw=black!70] (cand1.east) -- (cand1.east -| verif.west);
\draw[flow] (cand1.east -| verif.east) -- (fb1.west);
\draw[line width=0.6pt, draw=black!70] (cand2.east) -- (cand2.east -| verif.west);
\draw[flow] (cand2.east -| verif.east) -- (fb2.west);
\coordinate (merge) at ($(gen.east)!0.55!(cand2.west)$);
\coordinate (loopY) at ($(cand2.south) + (0, -0.35cm)$);
\coordinate (loopX) at ($(fb2.east) + (0.35cm, 0)$);
\draw[looparrow]
    (fb1.east)                
    -- (fb1.east -| loopX)    
    |- (loopY -| loopX)       
    -- (loopY -| merge)       
    -- (merge);               

\end{tikzpicture}
}
\caption{RubricRefine overview. \emph{Setup (top row):} the task instruction and tool documentation (\circled{1}) are passed to the rubric generator $V_R$ (\circled{2}), which produces a task-specific rubric $\mathcal{R}$ (\circled{3}) of itemized contract checks. \emph{Refinement loop (bottom row):} the generator $G$ (\circled{4}) produces a candidate $c_r$ each round; the candidate flows through the verifier $V$ (\circled{6}), which scores it against $\mathcal{R}$ and emits that round's score, item-level PASS/FAIL directives, and revision suggestions. Worked example: round-1's $c_1$ (\circledw{5a}) passes \texttt{"-pool"} as a hotel-preference argument (a contract violation that would run without raising an exception), which $V$ flags as a FAIL in round-1 feedback (\circledw{7a}), along with a fix (``call \texttt{book\_hotel("B","gym")} and filter out \texttt{"pool"} via a post-call filter''). This round-1 feedback feeds back into the generator as the revision prompt driving round~2. Round-2's $c_2$ (\circledw{5b}) applies the suggested fix, replacing the invalid argument with a post-call filter. The verifier grades $c_2$ at 10/10 against the rubric in round-2 feedback (\circledw{7b}), triggering early stopping and execution of $c_2$. Each round's candidate is drawn with surrounding ``\texttt{\textellipsis}'' to indicate omitted code.}
\label{fig:rubricrefine-example}
\end{figure*}

\section{Method}
\label{sec:method}

\subsection{Problem Formulation (Single-Attempt Execution)}

We adopt the executable-action interaction setting introduced in CodeAct~\citep{wang2024executable}. Each task instance provides an instruction $\mathcal{I}$ and a tool registry $\mathcal{T} = \{t_1,\ldots,t_K\}$, where every tool exposes a name, signature, and documentation string. The agent's action is a piece of executable code $c$, and our deployment assumption is \textbf{single-attempt execution}: the agent may spend arbitrary inference-time compute before acting, but the environment is executed at most once per instance.

\subsection{RubricRefine: Pre-Execution Algorithm}

RubricRefine is a control loop over a generator and a verifier\footnote{We use \emph{verifier} for the scoring role in the sense standard in the process-supervision literature \citep{uesato2022process,lightman2024verify}: a model that grades intermediate work. Unlike the trained process reward models of that line, our verifier is a prompted LLM grader in the sense of \citet{zheng2023judging}, \citet{kim2024prometheus}, and \citet{kwok2026llmverifier}. We do not claim verification in the formal sense, since a passing score is evidence of contract compliance rather than a proof of it.}, not a single prompted call. It maintains explicit state --- an instance-specific rubric $\mathcal{R}$, held fixed for the task so the optimisation target cannot drift --- and defines a scoring function over that state, a gating policy mapping item-level verdicts to an ordinal score, a patience-based termination rule, and a selection rule over the candidates produced. Figure~\ref{fig:rubricrefine-example} illustrates a complete trajectory.

\subsubsection{Rubric Generation}
\label{sec:phase-a}

Given $(\mathcal{I}, \mathcal{T})$, a rubric generator $V_R$ produces a structured rubric $\mathcal{R}$ containing itemized checks. In our implementation $V_R$ is the same underlying LLM as the verifier $V$ (Section~\ref{sec:phase-b}), invoked with a distinct rubric-construction prompt; they remain two separate functional roles. This rubric is \emph{sample-dependent}: its criteria are generated conditioned on the specific task instance and tool registry rather than drawn from a fixed template. Sample dependence is the variable this paper isolates: Section~\ref{sec:main-results} holds the loop, the PASS/FAIL format, and the score aggregation constant while varying only whether criteria are instance-conditioned, so the reported gap over Fixed RubricRefine is attributable to the rubric's conditioning rather than to the presence of a critique loop. The rubric criteria fall into four categories that cover distinct dimensions of correctness arising broadly in code-mode tool use:
\begin{itemize}
    \item \textbf{Tool-choice rules}: whether the agent dispatches to the correct documented API rather than reimplementing equivalent logic in the host language.
    \item \textbf{Output-contract rules}: whether the program's final emitted value conforms to the expected shape and type.
    \item \textbf{Call-signature rules}: whether each tool invocation matches its documented signature in argument names, keyword usage, and arity.
    \item \textbf{Data-provenance rules}: whether values consumed by downstream calls trace back to tool return values or task-specified literals rather than being fabricated.
\end{itemize}

These four categories delimit the failure space pre-execution checking must cover, and only one is reliably visible to prior mechanisms: an immediate type mismatch in a call signature raises at runtime. Wrong tool selection, wrong output shape, and broken argument provenance run to completion without raising --- the silent contract violations that motivate checking all four categories before execution. Two caveats bound this. The checks are LLM judgments, so a PASS is evidence of plausibility rather than proof. And all four are evaluated \emph{against the supplied documentation}, so they inherit its errors --- a dependence we quantify in Section~\ref{sec:doc-robustness}.

\subsubsection{Scoring and Gating}
\label{sec:phase-b}

At refinement round $r$, candidate code $c_r$ is scored against $\mathcal{R}$:
\begin{equation}
(s_r, f_r) = V(c_r, \mathcal{R}), \quad s_r \in \{1,\ldots,10\}.
\end{equation}

$f_r$ is structured item-level feedback consisting of PASS/FAIL judgments, reasons, and revision directives.

\paragraph{Scoring Policy and Failure Gating}
RubricRefine uses an ordinal 1--10 score with explicit gating: low bands for missing core calls, middle bands when critical contract errors remain, and 10 only for fully grounded, contract-compliant code. Appendix~\ref{app:method-impl} details how these caps are enforced at the prompt level.

\subsubsection{Iterative Repair}
\label{sec:phase-c}

At each round, generator $G$ receives $(\mathcal{I},\mathcal{T},\mathcal{R},c_{r-1},f_{r-1})$ and emits a revised candidate $c_r$. The loop terminates when the max score is reached, patience $P$ is exhausted, or the round budget runs out; the best-scoring candidate is then selected as the task's single executable action.

Key implementation details, including the pseudocode (Algorithm~\ref{alg:rubricrefine}), prompt structure, generator/verifier configuration, decoding settings, and the values of $N$, $R$, and $P$, are provided in Appendix~\ref{app:method-impl}. The verbatim prompt templates for rubric generation, scoring, and repair are reproduced in Appendix~\ref{app:prompts}.

\section{Experiments}
\label{sec:experiments}

\begin{table*}[!t]
\centering
\footnotesize
\setlength{\tabcolsep}{4pt}
\resizebox{\textwidth}{!}{%
\begin{tabular}{lcccccccc}
\toprule
Method & \texttt{GPT-4.1-mini} & \texttt{GPT-4o} & \texttt{o3-mini} & \texttt{GPT-4.1} & \texttt{Gemma-4-26B} & \texttt{Qwen3.6-27B} & \texttt{Sonnet-4.6} & Mean \\
\midrule
CodeAct (Baseline) & $.64 \pm .02$ & $.67 \pm .01$ & $.66 \pm .01$ & $.65 \pm .02$ & $.50 \pm .01$ & $.46 \pm .01$ & $.74 \pm .00$ & $.62$ \\
Self-Refine & $.60 \pm .02$ & $.62 \pm .02$ & $.60 \pm .02$ & $.60 \pm .02$ & $.71 \pm .01$ & $.67 \pm .01$ & $.71 \pm .01$ & $.64$ \\
Self-Debug & $.75 \pm .02$ & $.73 \pm .01$ & $.75 \pm .01$ & $.75 \pm .01$ & $.65 \pm .01$ & $.76 \pm .01$ & $.79 \pm .01$ & $.74$ \\
Best-of-$N$ & $.65 \pm .02$ & $.65 \pm .02$ & $.61 \pm .01$ & $.62 \pm .01$ & $.53 \pm .01$ & $.51 \pm .00$ & $.75 \pm .00$ & $.62$ \\
\midrule
BoN+fixed rubric (Ours, Ctrl) & $.74 \pm .01$ & $.75 \pm .01$ & $.75 \pm .02$ & $.73 \pm .01$ & $.52 \pm .01$ & $.51 \pm .02$ & $.75 \pm .00$ & $.68$ \\
BoN+rubric (Ours, Ctrl) & $.75 \pm .01$ & $.75 \pm .01$ & $.77 \pm .01$ & $.76 \pm .01$ & $.50 \pm .01$ & $.51 \pm .01$ & $.76 \pm .01$ & $.69$ \\
Fixed RubricRefine (Ours, Ctrl) & $.80 \pm .01$ & $.75 \pm .01$ & $.78 \pm .01$ & $.80 \pm .01$ & $.73 \pm .01$ & $.76 \pm .01$ & $.83 \pm .00$ & $.78$ \\
RubricRefine (Ours) & $\mathbf{.86 \pm .01}$ & $\mathbf{.86 \pm .01}$ & $\mathbf{.85 \pm .01}$ & $\mathbf{.85 \pm .01}$ & $\mathbf{.85 \pm .01}$ & $\mathbf{.84 \pm .01}$ & $\mathbf{.88 \pm .01}$ & $\mathbf{.86}$ \\
\bottomrule
\end{tabular}%
}
\caption{M3ToolEval success rates (mean $\pm$ SE across trials) for seven models. M3ToolEval tasks require multi-step tool composition with dataflow between coordinated API calls. Bolded cells mark each model's best method; RubricRefine is best on every model tested. Rows marked ``(Ours, Ctrl)'' are our own control methods for factorial decomposition.  Per-model paired $t$-test $p$-values, Wilcoxon robustness checks, trial SDs, and minimum gaps are reported in Appendix~\ref{app:eval-config}. Logprob-weighted scoring variants are reported separately in Table~\ref{tab:logprob-m3} (Appendix~\ref{app:logprob-results}).}
\label{tab:main-m3-success}
\end{table*}

\subsection{Experimental Setup}
\label{sec:eval}

We evaluate under the single-attempt protocol on two complementary CodeAct-aligned benchmarks. To estimate variance, we run 10 independent trials per model--method pair; cross-trial variation comes from the generator's sampling temperature ($T = 0.7$). We evaluate seven models spanning four frontier OpenAI APIs (\texttt{GPT-4.1-mini}, \texttt{GPT-4o}, \texttt{o3-mini}, \texttt{GPT-4.1}), two recent open-weight systems (\texttt{Gemma-4-26B-A4B-it}~\citep{googledeepmind2026gemma4}, abbreviated \texttt{Gemma-4-26B}; and \texttt{Qwen3.6-27B-FP8}~\citep{qwen3.6-27b}, abbreviated \texttt{Qwen3.6-27B}), and a recent Anthropic Claude model (\texttt{Claude-Sonnet-4.6}, abbreviated \texttt{Sonnet-4.6} in tables) to test whether findings transfer across backbones. \textbf{M3ToolEval} is the main-text benchmark because it directly measures end-to-end executable task success in code-as-action settings. Specifically, on M3ToolEval, success is the fraction of instances where the code-mode agent produced code that yields the ground truth value when executed. We record wall-clock latency, LM calls, and tokens alongside success rate on M3ToolEval. \textbf{API-Bank} provides a complementary step-level view by measuring exact API-call correctness (Section~\ref{sec:apibank-results}, Appendix~\ref{app:apibank}).

\subsection{Baseline and Competitive Methods}
All methods share the same code-as-action format and underlying model; they differ only in inference-time strategy. \textbf{CodeAct (Baseline)}~\citep{wang2024executable} is single-pass code-mode generation without any CodeAct-specific post-training. \textbf{Self-Refine}~\citep{madaan2023selfrefine} iteratively critiques and revises in free-form natural language (the same loop as RubricRefine but without a rubric). \textbf{Self-Debug}~\citep{chen2023selfdebug} executes the candidate, observes the output or error, and revises iteratively, using real environment interaction that RubricRefine is not allowed. \textbf{Best-of-$N$} samples $N$ candidates and selects the highest self-rated one. \textbf{Best-of-$N$+fixed rubric} and \textbf{Best-of-$N$+rubric} (ours) select via a fixed or sample-dependent rubric score respectively, isolating the effect of sample dependence in the selection setting. \textbf{Fixed RubricRefine} (ours) uses the iterative repair loop with a fixed rubric; comparing it against \textbf{RubricRefine} (ours, full method from Section~\ref{sec:method}) isolates the effect of sample-dependent rubric generation in the refinement setting. All iterative methods (Self-Refine, Self-Debug, Fixed RubricRefine, RubricRefine) are allowed up to $5$ rounds; Best-of-$N$ variants use $N = 5$ candidates. Logprob-weighted variants~\citep{kwok2026llmverifier} are evaluated in Appendix~\ref{app:logprob-results}.

\subsection{Main Results (M3ToolEval)}
\label{sec:main-results}

Table~\ref{tab:main-m3-success} reports M3ToolEval success rates over $10$ independent trials on each of seven models. RubricRefine is best on every model, and post-method rates cluster tightly in $[0.85, 0.88]$ despite CodeAct baselines spanning $0.46$--$0.74$: the method saturates near $0.86$ regardless of baseline, so the lift scales with baseline weakness. Averaged across models it improves over CodeAct by $+0.24$ ($0.86$ vs.\ $0.62$), with per-model two-sided paired $t$-tests on the 10 matched trials at $p < 0.001$ throughout (SDs, ranges, and minimum gaps in Appendix~\ref{app:eval-config}), and over Self-Refine on every model ($\alpha = 0.01$). Self-Debug, which executes real code and revises from the observed output or error, also beats single-pass CodeAct ($0.74$ vs.\ $0.62$; all $p < 0.02$) --- execution feedback is genuinely informative --- yet RubricRefine with zero execution beats \emph{it} by $+0.12$ (all $p < 0.001$). The gap is explained by what triggers a retry: Self-Debug engages only on runtime exceptions, while the dominant failures are inter-tool contract violations that run to completion without raising. A worked example is in Appendix~\ref{app:qualitative-example}.

The per-task breakdown makes the mechanism concrete (Appendix~\ref{app:frontier-per-task}). Gains scale with how much inter-tool contract structure a family contains. Travel itinerary planning --- three coordinated calls with cross-call argument provenance --- improves by $+0.41$ to $+0.48$ on the OpenAI models and $+0.77$ to $+0.81$ on the two open-weight models, but only $+0.06$ on \texttt{Sonnet-4.6}, whose baseline is already $0.73$. Message decoder and DNA sequencer, with lighter composition, improve moderately ($-0.02$ to $+0.48$). Trade calculator, arithmetic-heavy with minimal inter-tool dataflow, is flat on the OpenAI models ($-0.01$ to $+0.02$) though positive on \texttt{Gemma-4-26B} ($+0.12$), \texttt{Qwen3.6-27B} ($+0.11$), and \texttt{Sonnet-4.6} ($+0.27$). This dataflow-to-gain relationship is the paper's sharpest mechanism evidence, and its corollary is that RubricRefine should \emph{not} help where no inter-tool contracts exist: on the single-step API-Bank it is flat or within noise on all four OpenAI models\label{sec:apibank-results} (Appendix~\ref{app:apibank}), and sample-dependent rubrics can even hurt by over-specifying API choices from dialogue context, which Fixed RubricRefine's generic rubric avoids (Appendix~\ref{app:apibank-regression}).

\paragraph{Structured contract checking beats unstructured critique.}
Fixed RubricRefine improves over Self-Refine on every model, by $+0.09$ to $+0.20$ on six of seven (all per-model paired $t$-tests $p < 0.001$): contract-structured feedback substantially outperforms free-form critique in the same generate-verify-revise loop. The exception is \texttt{Gemma-4-26B} ($+0.02$, $p = 0.14$), where free-form self-critique already extracts enough revision signal to narrow the incremental gain from adding structure.

\paragraph{Sample-dependent rubrics beat fixed rubrics for repair.}
RubricRefine improves over Fixed RubricRefine on every model ($+0.04$ to $+0.12$; all $p < 0.02$). Both use the same iterative loop, PASS/FAIL format, and score aggregation, differing only in whether the checks are conditioned on the current task instance. The consistency of this gap isolates \emph{sample dependence in the rubric criteria themselves} as the component that matters for revision signal rather than for scalar ranking.

\paragraph{Rubric structure helps selection on frontier models but not on smaller open-weight models.}
Adding a rubric to Best-of-$N$ selection (BoN $\rightarrow$ BoN+rubric) improves success by $+0.10$--$+0.14$ on the four OpenAI models and by $+0.01$ on the Anthropic model, but produces no gain on \texttt{Gemma-4-26B} ($-0.03$) or \texttt{Qwen3.6-27B} ($0.00$). The two null cases are the smallest open-weight models we evaluate: selection requires rubric scores to track correctness across the full range, which frontier verifiers satisfy and these two do not. Iterative repair instead consumes the itemized PASS/FAIL structure as a revision directive, and its only scalar dependence is the stopping test at the top bin --- a weaker requirement that both open-weight verifiers do meet (Section~\ref{sec:calibration}, Appendix~\ref{app:middle-bin-calibration}). Sample dependence adds little in the selection setting (BoN+fixed rubric vs.\ BoN+rubric differs by $\le 0.02$ on six of seven models, $0.03$ on \texttt{GPT-4.1}): it pays off in iterative repair, not scalar reranking. Logprob-weighting further improves selection (Appendix~\ref{app:logprob-results}, consistent with \citealp{kwok2026llmverifier}), but iterative repair remains strongest on every model.

\subsection{Cross-Provider Verifier}
\label{sec:cross-verifier}

Our main results instantiate the generator and verifier from the same backbone, so the gain could in principle reflect self-confirmation rather than informative feedback. To test this, we pair a \texttt{GPT-4.1} generator with a \texttt{Claude-Sonnet-4.6} verifier from a different provider, holding the prompts, budgets, and evaluation semantics fixed. The cross-provider pairing yields $0.861$ (SD $0.032$, $3$ trials) versus $0.852$ for the same-model verifier. The gain is preserved rather than reduced --- the direction that matters, since self-confirmation predicts degradation. Appendix~\ref{app:cross-verifier} states the scope of this check.

\subsection{Efficiency Analysis}
\label{sec:efficiency}

RubricRefine's gain is not merely bought with additional inference compute; it converts that compute into success more efficiently. On \texttt{GPT-4.1} it averages $30.0$s per task, a $2.6\times$ wall-clock reduction over Best-of-$N$+rubric at higher accuracy, and at latency comparable to Self-Refine it reaches $0.85$ against $0.60$. Self-Debug, our execution-aware baseline, is faster ($3.1$s) but converts that budget into half the gain over CodeAct on this model ($+0.10$ vs.\ $+0.20$). The mechanism is \textbf{early stopping}: the loop terminates once the rubric score saturates, which $\approx\!90\%$ of tasks do within two rounds (Appendix~\ref{app:round-stopping}), yielding lower LM-call usage than rubric-guided reranking on all seven models (Appendix~\ref{app:early-stopping}). Nearly all of the accuracy gain materializes in the first refinement round, whereas Best-of-$N$+rubric climbs gradually and has not caught up at $N=5$ (Appendix~\ref{app:budget-scaling}). Wall-clock latency is measured on \texttt{GPT-4.1} (Table~\ref{tab:efficiency}) and, in a locally served regime, on \texttt{Gemma-4-26B}, where the reduction against rubric-guided reranking is $2.3\times$ in wall-clock and ${\approx}49\%$ in tokens (Appendix~\ref{app:gemma4-tradeoffs}) --- so the efficiency advantage is not an artifact of API serving latency. The token and LM-call reductions do generalize: RubricRefine uses $43$--$50\%$ fewer tokens than rubric-guided reranking on all five API-served models and fewer LM calls on all seven (Appendix~\ref{app:early-stopping}); per-model tradeoff curves are in Appendix~\ref{app:scaling}.

\subsection{Calibration and Early Stopping}
\label{sec:calibration}

RubricRefine's early stopping depends on rubric scores being reliable enough that a high score indicates an execution-ready candidate.

\begin{figure}[t]
\centering
\begin{minipage}[t]{0.48\linewidth}
\centering
\includegraphics[width=\linewidth]{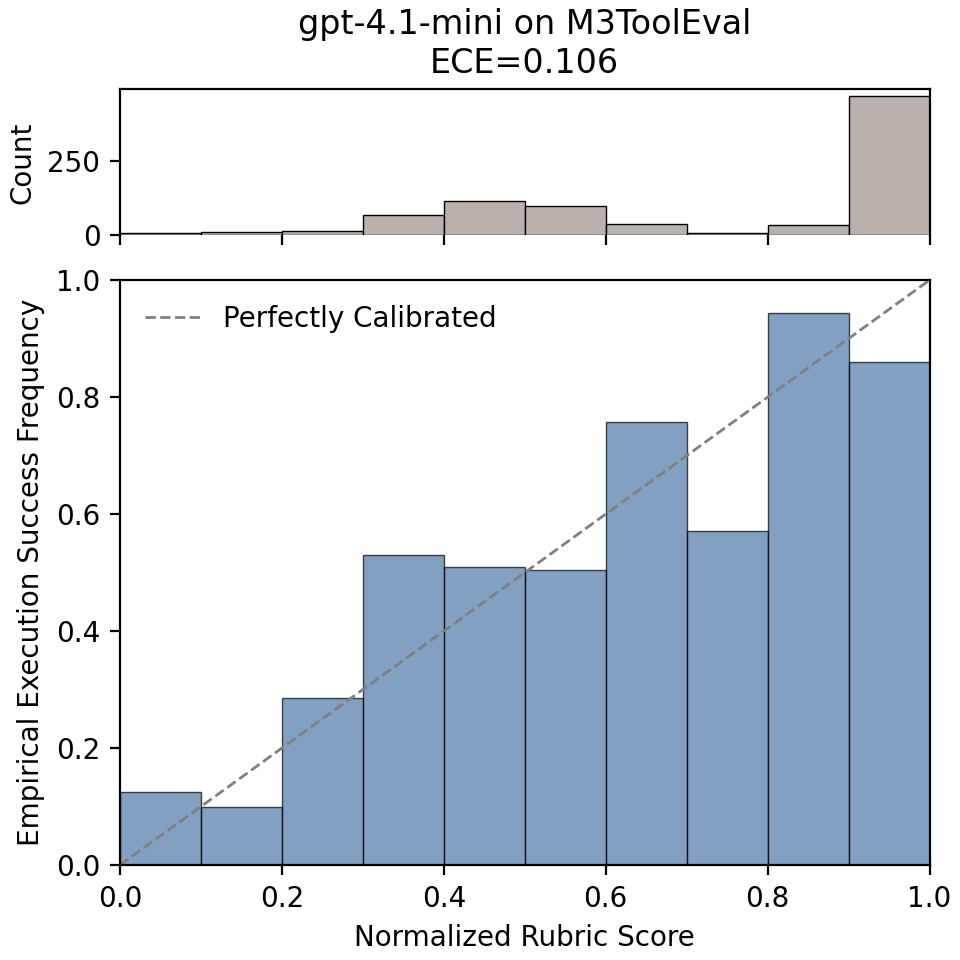}
\end{minipage}
\hfill
\begin{minipage}[t]{0.48\linewidth}
\centering
\includegraphics[width=\linewidth]{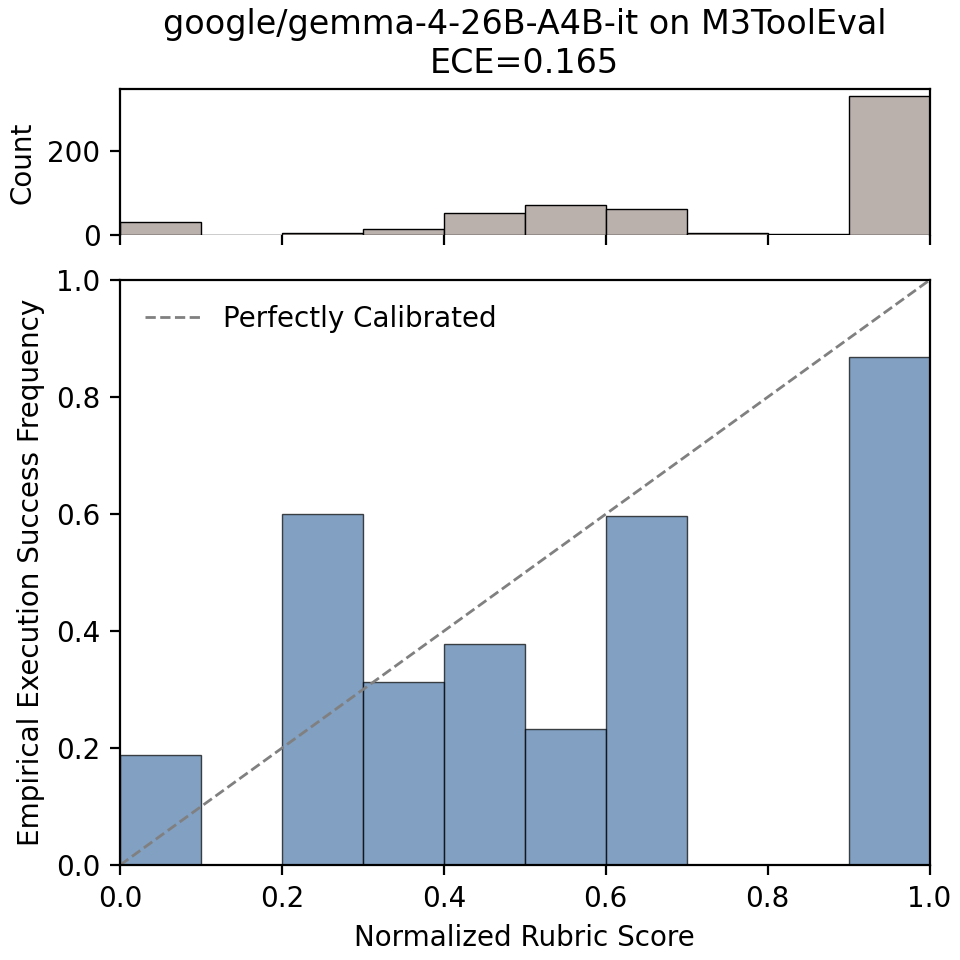}
\end{minipage}
\caption{Reliability diagrams for normalized rubric scores on M3ToolEval. Left: \texttt{GPT-4.1-mini} ($\text{ECE}=0.106$), well-calibrated across all bins. Right: \texttt{Gemma-4-26B} ($\text{ECE}=0.165$), poorly calibrated in the middle bins but retaining meaningful top-bin separation (accuracy $0.87$ at score $= 10$, $n=329$). RubricRefine's early stopping depends only on the top bin, so the method remains effective on Gemma despite the verifier being unreliable as a scalar ranker due to poor calibration in the middle bins. The \texttt{GPT-4.1} reliability diagram, also well-calibrated, is in Appendix~\ref{app:gpt41-calibration}.}
\label{fig:reliability-main}
\end{figure}

\paragraph{Globally calibrated verifiers.}
On held-out RubricRefine trajectories, the normalized rubric score (assigned score divided by maximum) predicts binary execution success with AUROC $0.796$/$0.764$ and ECE~\citep{niculescumizil2005predicting,guo2017calibration} $0.106$/$0.115$ for \texttt{GPT-4.1-mini}/\texttt{GPT-4.1} --- reasonably calibrated in every bin with no post hoc transformation, consistent with prior findings that LLMs can produce calibrated self-assessments~\citep{kadavath2022language}. Figure~\ref{fig:reliability-main} (left) shows \texttt{GPT-4.1-mini}'s top bin (score $= 10$, the early-stopping trigger) aligned with the diagonal: candidates reaching the maximum score are execution-ready at the rate the score implies. The \texttt{GPT-4.1} diagram in Appendix~\ref{app:gpt41-calibration} shows the same finding.

\paragraph{Top-bin-only calibrated verifiers.}
Global calibration is sufficient but not necessary: only the score $=10$ bin triggers termination, so only it must be reliable. \texttt{Gemma-4-26B}, the smallest model we evaluate, satisfies this weaker property despite worse aggregate calibration (ECE $0.165$, Figure~\ref{fig:reliability-main} right), consistent with prior findings that smaller models produce less calibrated self-assessments~\citep{kadavath2022language}: iterative repair drives candidates toward the top bin, where accuracy is $0.87$ --- a $0.13$ deviation, below the global ECE. That reliability is joint rather than a property of the verifier alone --- the same verifier degrades on cold parallel samples (AUROC $0.795 \rightarrow 0.700$) --- an argument developed in Appendices~\ref{app:calibration} and~\ref{app:gemma4-calibration}.

\subsection{Rubric Category Ablation}
\label{sec:ablation}

To identify which rubric categories drive the gains, we remove each of the four thematic groups from Section~\ref{sec:phase-a} from all three system prompts, keeping the rest intact, and run $10$ matched-seed trials per condition on two models of differing capability (Table~\ref{tab:ablation-multi}).

The clearest pattern is a \textbf{capability gradient: stronger models need less explicit scaffolding.} Every ablation costs the weaker \texttt{GPT-4.1-mini} more than \texttt{GPT-4o}, and the gap is large where it matters most --- removing call-signature rules costs $-0.146$ ($p<10^{-5}$) against $-0.058$ ($p<0.001$), and removing tool-choice rules costs $-0.158$ ($p<10^{-6}$) against a null $-0.012$ ($p=0.30$). Enumerating a contract explicitly buys most when the generator cannot reliably infer it from documentation alone; at higher capability the same rules become partly redundant. \emph{Output-contract} removals are within noise on both ($p=0.33$ and $p=0.15$), so we do not claim all four categories contribute reliably. \emph{Call-signature} is the only category load-bearing on both models ($p<0.001$ and $p<10^{-5}$), consistent with call shape being the dimension least recoverable from prose description. That dimension is also the one a deterministic checker can most plausibly implement, yet substituting an AST-based signature checker for the LLM verifier underperforms even unrefined CodeAct (Appendix~\ref{app:static-checker}).

\subsection{Dependence on Documentation Quality}
\label{sec:doc-robustness}

Because the rubric derives entirely from the task instruction and tool documentation, its value depends on that documentation's quality. We stress-test this on \texttt{GPT-4o} across all four M3ToolEval families ($10$ trials per condition) under two perturbations: \textbf{truncation}, dropping the last $25/50/75\%$ of each tool description, and \textbf{parameter scrambling}, replacing every documented parameter name with a generic one (\texttt{arg0}, \texttt{arg1}, \ldots) to simulate documentation that is confidently \emph{wrong} rather than merely incomplete.

\begin{table}[t]
\centering
\footnotesize
\setlength{\tabcolsep}{5pt}
\resizebox{\columnwidth}{!}{%
\begin{tabular}{lcccc}
\toprule
Condition & CodeAct & RubricRefine & $\Delta$ & 95\% CI \\
\midrule
Clean               & $0.713$ & $0.831$ & $+0.119$ & $[+.084,+.153]$ \\
Truncate $25\%$     & $0.377$ & $0.592$ & $+0.215$ & $[+.153,+.276]$ \\
Truncate $50\%$     & $0.365$ & $0.550$ & $+0.185$ & $[+.141,+.230]$ \\
Truncate $75\%$     & $0.352$ & $0.560$ & $+0.208$ & $[+.181,+.236]$ \\
Scramble parameters & $0.600$ & $0.421$ & $\mathbf{-0.179}$ & $[-.237,-.122]$ \\
\bottomrule
\end{tabular}%
}
\caption{Degraded-documentation robustness on M3ToolEval (\texttt{GPT-4o}, $10$ trials per condition, $R=5$). $\Delta$ is the mean paired RubricRefine $-$ CodeAct difference; CIs are $95\%$ intervals on the paired difference. Truncating descriptions preserves RubricRefine's advantage at every severity; scrambling documented parameter names reverses it.}
\label{tab:doc-robustness}
\end{table}

The result is asymmetric (Table~\ref{tab:doc-robustness}). Truncation degrades both methods sharply in absolute terms yet preserves RubricRefine's advantage at every severity ($+0.19$ to $+0.22$, every CI excluding zero), because function \emph{signatures} --- which carry the structural contract --- survive even when prose descriptions are cut. Scrambling reverses the advantage ($-0.179$). The mechanism is direct: the rubric transcribes the wrong names into binding criteria and the repair loop rewrites correct calls to satisfy them, converting a documentation error into a code error.

This is the method's most important boundary condition: \textbf{a high rubric score is not evidence of correctness when the supplied contract may itself be wrong}, only of agreement with the documentation. Where tool docs may be stale or auto-generated, a rubric score should not be a release gate on its own; Appendix~\ref{app:doc-robustness} discusses safeguards, none evaluated here.

\subsection{Interaction Budgets in a Multi-Turn Setting}
\label{sec:appworld}

Our protocol so far permits one live action per task. To test whether pre-execution checking still pays when an agent may act repeatedly, we evaluate on \textbf{AppWorld}~\citep{trivedi2024appworld}: nine applications, $457$ APIs, and tasks requiring several coordinated calls against live state. The agent interacts over up to eight turns, and RubricRefine refines \emph{each} turn's block before that block executes, so refinement remains pre-execution but applies at turn rather than task granularity. Both arms receive an identical task-scoped registry from the benchmark's own documentation search, differing only in whether each block is checked before it runs. Adapting the method to a stateful interpreter required exposing the live variable namespace to the verifier (Appendix~\ref{app:appworld}).

\begin{figure}[t]
\centering
\includegraphics[width=0.60\columnwidth]{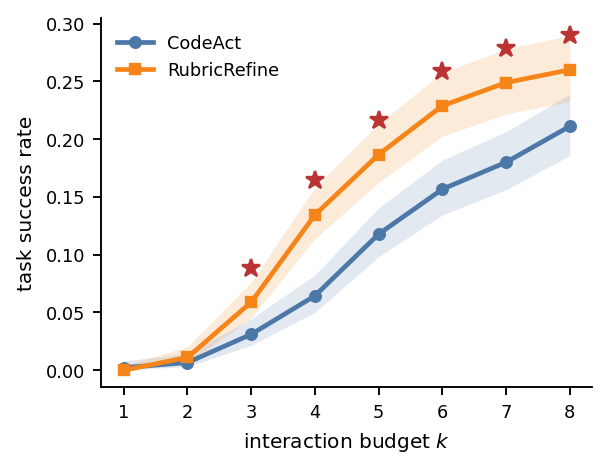}
\caption{Task success against interaction budget $k$ on AppWorld. A point at $k$ is the fraction of tasks completed within $k$ executed turns and passing evaluation. Bands are $95\%$ Wilson intervals; stars mark paired McNemar tests at $p<0.05$ over 90 tasks with 10 trials each ($900$ paired episodes). The paired-test rationale and protocol details are in Appendix~\ref{app:appworld}.}
\label{fig:appworld}
\end{figure}

Figure~\ref{fig:appworld} reports success against the interaction budget $k$. RubricRefine leads significantly from $k=3$ onward. Extra turns help the baseline substantially --- probing live state recovers failures that surface as errors --- but never close the gap: pre-execution contract checking catches what no amount of execution can surface. This mirrors the Self-Debug comparison of Section~\ref{sec:main-results}.

\section{Conclusion}
\label{sec:conclusion}

Iterative self-refinement is only as good as its feedback signal. If RubricRefine effectively structures its feedback as pre-execution contract checking, four results follow, and we observe all four: it beats unstructured critique (Self-Refine) on every model; it beats execution feedback (Self-Debug) precisely where the retry loop never engages because the failing code returned a wrong answer without raising; per-task lifts scale with inter-tool dataflow richness; gains vanish on the predominantly single-step API-Bank; and on a multi-turn benchmark the gap narrows but survives an eight-turn interaction budget, as it must if the failures are invisible to execution. The key finding is not the iterative loop but the contract structure: sample-dependent rubrics that decompose tool-use correctness into individually checkable criteria turn an unreliable revision process into an effective one.

The same mechanism sets the boundary: because every check is evaluated against the supplied documentation, the method inverts when that documentation is wrong. RubricRefine belongs where the contract is trustworthy and a failed action is costly. Validating docs against executable tool definitions, extending to larger and messier registries, and evaluating rubric quality directly rather than only through downstream success are natural next steps.

\section*{Limitations}
\label{sec:limitations}

RubricRefine improves pre-execution reliability, but it does not eliminate all sources of failure. Because the method reasons over the task prompt, tool documentation, and generated code before execution, it cannot fully anticipate failures that depend on latent runtime state or external environment properties that are not observable at generation time. Examples include stale sessions, hidden preconditions, changing world state, or tool behaviors that are underspecified in the documentation. In such cases, even a strong rubric may certify code that later fails when exposed to the live environment.

The method is also bounded by verifier quality. If the generated rubric omits an important constraint, encodes a misleading decomposition, or scores candidates inaccurately, the resulting feedback can push revisions in the wrong direction. RubricRefine is an LLM-based heuristic for contract checking, not a sound verification procedure: it produces no guarantee, and a passing rubric score is not a proof of correctness. We also do not evaluate rubric quality directly. All our evidence is downstream task success, so we cannot separate a rubric that captures the right constraints from one that happens to steer the generator usefully for other reasons; human or expert assessment of rubric coverage would test this and remains future work.

A sharper bound comes from documentation quality. Section~\ref{sec:doc-robustness} shows the method tolerates incomplete documentation but inverts under incorrect documentation: with scrambled parameter names, RubricRefine falls $0.179$ \emph{below} CodeAct, because the rubric transcribes the erroneous contract into binding criteria and the repair loop enforces it. A rubric score therefore measures agreement with the supplied contract, not correctness, and should not be treated as a release gate where tool docs may be stale, auto-generated, or drifting from implementation. We propose but do not evaluate two safeguards: validating documented signatures against executable tool definitions before rubric generation, and treating divergence between refined and unrefined candidates as an abstention signal.

RubricRefine currently assumes the code-mode setting with an explicit tool registry, and is not intended as a drop-in replacement for broader free-form generation. The core generate-verify-revise loop is not inherently tied to code-mode, but adapting it to other settings would require replacing the code-mode-specific system prompts with domain-appropriate rubric criteria encoding the relevant correctness dimensions for each target domain. Our registries are also small (a handful of documented tools per task); how rubric generation behaves with hundreds of overlapping or near-duplicate tools, where the dominant difficulty shifts from contract compliance to tool disambiguation, is untested.

RubricRefine additionally requires extra inference-time compute for rubric generation, scoring, and iterative repair. In settings where execution is cheap and reversible, that preflight cost may not be justified; the tradeoff is attractive when failed executions are expensive, stateful, or safety sensitive.

Finally, our empirical validation covers M3ToolEval, API-Bank, and AppWorld. The first two capture two complementary views of reliability, namely whole-program executable success and exact API-call fidelity, and AppWorld adds a large registry under repeated interaction, where the persistence of the gap at the full budget is consistent with the single-attempt findings rather than an artifact of the one-action constraint. These do not exhaust the broader space of real-world tool-use constraints. Other tool-use benchmarks such as ToolBench~\citep{qin2024toolllm} and BFCL~\citep{berkeley-function-calling-leaderboard} evaluate JSON function-calling rather than code-as-action, so they do not directly apply to the code-mode setting this paper targets; the scarcity of multi-step code-mode benchmarks that admit single-attempt evaluation is a limitation of the current evaluation ecosystem, not only of this paper. Results should therefore be interpreted as evidence for the value of pre-execution rubric-guided refinement in these executable tool-use settings, not as a complete characterization of deployment behavior in all agent environments.

\clearpage
\input{references}

\clearpage
\appendix

\section{Extended Background and Motivation}
\label{app:background}

\subsection{Code Mode and CodeAct Foundation}
\label{sec:codeact-foundation}

\subsubsection{What ``Code Mode'' Means}

In this paper, \emph{code mode} means that the model emits executable code as the action payload. That code can call tools directly, branch on observed values, maintain intermediate state, and compute derived quantities before producing an answer. The resulting action semantics are therefore substantially richer than those of pure natural-language plans or single-call JSON actions. Recent open-source agent frameworks and production documentation use this same idea as a first-class implementation pattern \citep{roucher2024smolagents,cloudflare2025codemode}.

\subsubsection{What CodeAct Established}

CodeAct made three foundational contributions that this paper builds on \citep{wang2024executable}:
\begin{itemize}
    \item \textbf{Action-format evidence:} code actions can outperform text and JSON action formats on tool-use tasks.
    \item \textbf{Evaluation substrate:} executable-action evaluation was operationalized on benchmarks including M3ToolEval and API-Bank.
    \item \textbf{Data and agent artifacts:} CodeActInstruct and CodeActAgent provide a concrete training-and-evaluation workflow for code-mode agents.
\end{itemize}

Taken together, these contributions did more than propose a new interface: they established code mode as an experimentally grounded setting with data, agents, and benchmarks. The paper and repository materials describe this ecosystem at scale, including broad model benchmarking and a CodeActInstruct corpus of roughly 7k multi-turn trajectories spanning hundreds of tasks and APIs \citep{wang2024executable}.

\subsubsection{Why Reliability Becomes the Next Bottleneck}

Code mode improves expressivity and often improves success rates, but it also changes where failures concentrate. Once the action representation is powerful enough, many remaining errors are no longer high-level planning failures; they are \emph{contract-execution} failures. A model can be ``almost right'' in ways that still cause the program to fail:
\begin{itemize}
    \item selecting the right tool but grounding to the wrong real-world entity,
    \item using parameters with mismatched semantics (units, timezones, currencies, enums),
    \item making invalid state assumptions (missing preconditions, stale sessions, expired auth),
    \item violating policy, safety, or compliance constraints specified at inference time but not present in the training set.
\end{itemize}

These are near-miss failures that can look perfectly plausible in text while remaining wrong in execution. Reliability therefore becomes the next bottleneck: the system often discovers the problem only after an execution attempt has already consumed interaction budget, API quota, or side-effect risk.

A common mitigation is to retry after observing an execution error. That can recover some failures, but it is not a free repair mechanism. For stateful actions, the first attempt may already have changed the world before failing later in the program. Re-execution can therefore duplicate effects, create inconsistent state, or compound an error that was initially only partial. Worse, once the environment has been modified, later generations may be reasoning from an incomplete picture of the new state: not all relevant consequences are available in the prompt, and recovering them may require additional tool calls, time, or human intervention. In some settings, full recovery is impossible, and full observability into the post-execution state is also impossible. In high-consequence settings such as medicine, legal workflows, or systems that interact with the physical world, that combination of irreversibility and partial observability is operationally serious. This motivates the single-attempt setting we evaluate: the system should do as much checking and repair as possible \emph{before} the first live action.

Another way to address the bottleneck is post-training, as in the broader CodeAct line of work built around CodeActInstruct and CodeActAgent \citep{wang2024executable}. That path is important, but it requires collecting and curating trajectories, annotating or filtering training data, and rerunning training as registries evolve. Those steps are expensive, slow to refresh, and often impractical when the deployment problem is adaptation to a specific new registry rather than improvement of a base model in the abstract.

RubricRefine targets the same bottleneck from the deployment side. Its goal is to move failure detection earlier through registry-conditioned, pre-execution checks, so that fewer errors reach state-changing execution in the first place. This is particularly attractive in high-cost environments such as clinical operations, legal and financial workflows, and systems that interact with the physical world, where the marginal cost of an incorrect action can easily exceed the marginal cost of additional pre-execution reasoning. We cite these domains as motivation for reliability-first design, not as claims of direct deployment validation.

\section{Extended Related Work}
\label{app:rw-main}

\paragraph{Inference-Time Refinement}
Self-Refine~\citep{madaan2023selfrefine} is a canonical execution-free self-correction baseline in which the same model generates an answer, critiques it in natural language, and revises it. That baseline is directly relevant here because RubricRefine also spends inference-time compute on revision without relying on execution feedback, but replaces free-form critique with registry-conditioned, rubric-structured scoring. We also compare against parallel sampling because RubricRefine spends inference-time compute on sequential revision with a verifier, whereas a natural alternative is to spend that same budget on multiple independent samples and then select the best one. \citet{snell2024scaling} show that this tradeoff depends strongly on verifier quality, which makes Best-of-$N$+rubric the most direct control for whether RubricRefine's gains come from targeted repair rather than from extra test-time compute alone. \emph{CRITIC}~\citep{gou2024critic} is an especially relevant adjacent method: it uses tool-interactive critique rather than registry-conditioned, execution-free program verification, but it shares the core intuition that critique is more useful when it is grounded in external task-relevant information than when it is based on self-reflection alone. Tree of Thoughts~\citep{yao2023tree} is another approach to structured inference-time search, but uses branching exploration rather than a generate-verify-repair loop.

\paragraph{Rubric-Based Rewards in Post-Training}
Concurrently with this work, \citet{zhang2026chasingtail} show that rubric-based rewards mitigate reward over-optimization during post-training. That line and ours share the intuition that decomposed, itemized criteria give a more robust signal than a scalar objective, but apply it at different stages: theirs shapes a model during training, whereas RubricRefine leaves the model fixed and spends the criteria at inference time.

\paragraph{Tool-Use Post-Training and Unseen-Tool Adaptation}
Recent competitive function-calling systems usually improve tool use through better synthetic data, reflection tuning, reinforcement learning, or retrieval and planning mechanisms for unseen-tool adaptation rather than through inference-time checking alone~\citep{liu2024toolace,chen2025button,ma2025toolmvr,hao2025funreason,zhang2025tooln1,feng2025retool,lu2024gear,wu2025chainoftools,lumer2025graphragtoolfusion}. RubricRefine targets a different problem: deployment-time adaptation to a \emph{specific} tool registry without any additional training. This also makes the method directly applicable to API-only proprietary models, whereas registry adaptation through post-training is less directly available when the deployed model is closed-weight or exposed only through an inference API.

\paragraph{LLM-Based Code Verification and Execution Feedback}
A substantial body of work uses LLMs or learned models to verify or filter generated code, but the dominant paradigm relies on execution outcomes rather than pre-execution static checking. CodeRL~\citep{le2022coderl} trains an actor-critic framework for code generation where a critic model scores candidate programs; the critic is trained on execution outcomes and used to guide beam search. LEVER~\citep{ni2023lever} verifies LLM-generated programs by executing them on sampled inputs and using a learned verifier to aggregate execution signals. AlphaCode~\citep{li2022competition} uses large-scale sampling followed by clustering and execution-based filtering to select final outputs. CodeT~\citep{chen2022codet} and AlphaCodium~\citep{ridnik2024alphacodium} iterate a candidate against generated tests, which is pre-deployment in pure-code software-engineering tasks but not in tool use, where running a test is itself a live tool invocation that can affect the environment and is therefore incompatible with the single-attempt constraint. These methods achieve verification through execution: they require live interaction with the environment to obtain the signal used for selection or repair. RubricRefine operates before any execution, making it applicable precisely in settings where these methods cannot be used: when execution is rate-limited, stateful, costly, or unsafe to attempt on candidate code.

\paragraph{Execution-Feedback Refinement}
Self-Debug~\citep{chen2023selfdebug} and Reflexion~\citep{shinn2023reflexion} use execution error messages as feedback for iterative code revision. LATS~\citep{zhou2023language} performs tree search over agent trajectories using execution outcomes as the evaluation signal. We evaluate Self-Debug with real execution feedback as a direct baseline (Section~\ref{sec:main-results}): it improves modestly over single-pass generation but RubricRefine with zero execution outperforms it by a further $+0.10$ absolute, suggesting that the dominant contract failures in this setting do not produce informative runtime errors.

\paragraph{Static and Hybrid Verification}
PreFlect~\citep{wang2026preflect} uses prospective reflection to anticipate failures before acting, which is the closest conceptual precedent for pre-execution checking in agent settings. Unlike RubricRefine, PreFlect operates on natural-language plans rather than executable code and does not use a structured rubric conditioned on tool documentation. A natural question is whether such checking must be model-based at all, since call-shape conformance is mechanically decidable from an AST. Our results suggest the mode of checking matters more than the dimension: substituting a deterministic signature checker for the LLM verifier on precisely its strongest dimension underperforms the unrefined baseline in the selection setting (Section~\ref{sec:ablation}, Appendix~\ref{app:static-checker}).

\section{Extended Evaluation Design}
\label{app:eval-detail}

\subsection{Benchmark Rationale}

\paragraph{M3ToolEval (CodeAct-Aligned Executable Actions)}
M3ToolEval evaluates end-to-end tool-use completion in executable-action mode. In the repository configuration used here, each task is rendered with tool documentation and an output contract, and correctness is calculated by exact matching between the execution output and the annotated ground truth, where the match status is recorded as \texttt{is\_correct}.

\paragraph{API-Bank (Step-Level API Correctness)}
Whereas M3ToolEval evaluates an entire code-mode program at once, API-Bank provides exact string matching at the single API-call level, with strict checks on API identity and parameter correctness under the task context.

\paragraph{Why Both Are Needed}
M3 and API-Bank are not co-equal "win on both" benchmarks; they sit at opposite ends of the contract-complexity axis on which our mechanism is hypothesized to operate. M3ToolEval evaluates whole programs with rich inter-tool dataflow, where output contracts, call signatures, tool routing, and argument provenance interact across multiple steps. API-Bank evaluates predominantly single-step API calls drawn from multi-turn dialogues, where the dominant correctness question is identifying the next call and getting its arguments right rather than composing several tools. RubricRefine is designed to repair inter-tool contract violations, so the predicted pattern is a sizable lift on M3 and a flat (or worse) result on API-Bank. Reporting both therefore tests the mechanism rather than averaging two views of "reliability": the absence of a lift on API-Bank is itself evidence that the gains on M3 come from contract-structured checking rather than from extra inference compute generically improving outputs. Detailed benchmark semantics, the API-Bank regression analysis, and criterion differences are provided in Appendix~\ref{app:m3}, Appendix~\ref{app:apibank}, and Appendix~\ref{app:criteria}.

\subsection{Additional Metrics and Baselines}

\paragraph{Dual-View Reporting}
We report M3 and API-Bank separately and do not average across them. The two benchmarks evaluate the mechanism at opposite ends of the inter-tool contract-complexity axis (Appendix~\ref{app:criteria}); collapsing them into a single mean would obscure the mechanism evidence that comes from the asymmetric pattern of results.

\paragraph{Pre-Execution Efficiency Metrics}
Because RubricRefine emphasizes pre-execution reliability, we report pre-execution efficiency statistics including wall-clock latency, total LM calls, and total token usage, and we additionally compare observed LM-call usage against rubric-guided reranking in Appendix~\ref{app:early-stopping}.

\paragraph{Baselines}
Our comparison set includes:
\begin{itemize}
    \item single-pass code mode,
    \item Self-Refine \citep{madaan2023selfrefine},
    \item Best-of-$N$,
    \item Best-of-$N$ with a fixed rubric,
    \item Best-of-$N$ with a sample-dependent rubric,
    \item Fixed RubricRefine,
    \item RubricRefine.
\end{itemize}
Three additional logprob-weighted scoring variants are reported in Appendix~\ref{app:logprob-results}.

\paragraph{Single-Attempt Deployment Regime}
We intentionally do not evaluate post-execution retries in the main setting. The target use case is the single-attempt regime motivated in Section~\ref{sec:intro}, where runtime correction is too costly because even a partial execution may have already consumed scarce budget, triggered side effects, or altered external state in ways that are expensive or impossible to fully audit, reverse, or observe. All methods are therefore compared under the same constraint: unlimited pre-execution reasoning within the method's allotted inference budget, but only one live execution attempt per task.

\subsection{Calibration of Rubric Scores}
\label{app:calibration}

The calibration experiment is run on candidate trajectories produced by RubricRefine on M3ToolEval. Confidence is defined as the normalized rubric score, i.e., the assigned rubric score divided by the maximum possible rubric score for that task. Accuracy is the binary indicator of whether execution produced the correct ground-truth answer. To avoid leakage across sibling candidates, all candidates from the same task instance are kept in the same split. We compute AUROC for ranking quality and ECE using 10 confidence bins. For \texttt{GPT-4.1-mini} and \texttt{GPT-4.1}, the calibration analysis uses $240$ scored candidate trajectories each; for \texttt{Gemma-4-26B} it uses $580$ scored candidate trajectories pooled across 10 trials. The main text reports the resulting AUROC and ECE values.

\paragraph{Reliability Diagrams}
We visualize calibration through \textbf{reliability diagrams}~\citep{degroot1983comparing,niculescumizil2005predicting}. Following the exposition of~\citet{levine2023calibration}, these diagrams group points by their predicted confidence scores into $M$ equally spaced bins, and then compute the true and estimated accuracies in each bin as follows: let $B_m$ be the test samples whose confidence (i.e.\ estimated accuracy) falls into the interval $I_m = \big((m-1)/M,\; m/M\big]$ for $m = 2, \dots, M$, and $I_1 = [0, 1/M]$. The true accuracy is $\text{acc}(\hat{f}, B_m)$ and the \textit{estimated} accuracy (i.e.\ average confidence) within $B_m$ is $\hat{p}(\hat{f}, B_m) = \frac{1}{|B_m|}\sum_{i \in B_m}\hat{p}_i$. The reliability diagram plots the difference between true accuracy and estimated accuracy for all $M$ bins, and deviations from the line $f(x)=x$ represent miscalibrations: areas where there is a significant difference between the estimated and true accuracy.

\paragraph{Expected Calibration Error (ECE)}
We quantify miscalibration with the \textbf{Expected Calibration Error (ECE)}~\citep{10.5555/2888116.2888120}. ECE, aimed at summarizing the miscalibration visualized in reliability diagrams, is calculated as
\begin{equation}
    \text{ECE} = \sum_{m=1}^{M} \frac{|B_m|}{|D|}\,\bigl\lvert\hat{p}(\hat{f}, B_m) - \text{acc}(\hat{f}, B_m)\bigr\rvert.
\end{equation}
Lower ECE means the reported confidence is more interpretable as a probability of success. For all experiments we use $M=10$ bins, as is standard~\citep{guo2017calibration}.

\begin{figure}[t]
\centering
\includegraphics[width=0.6\linewidth]{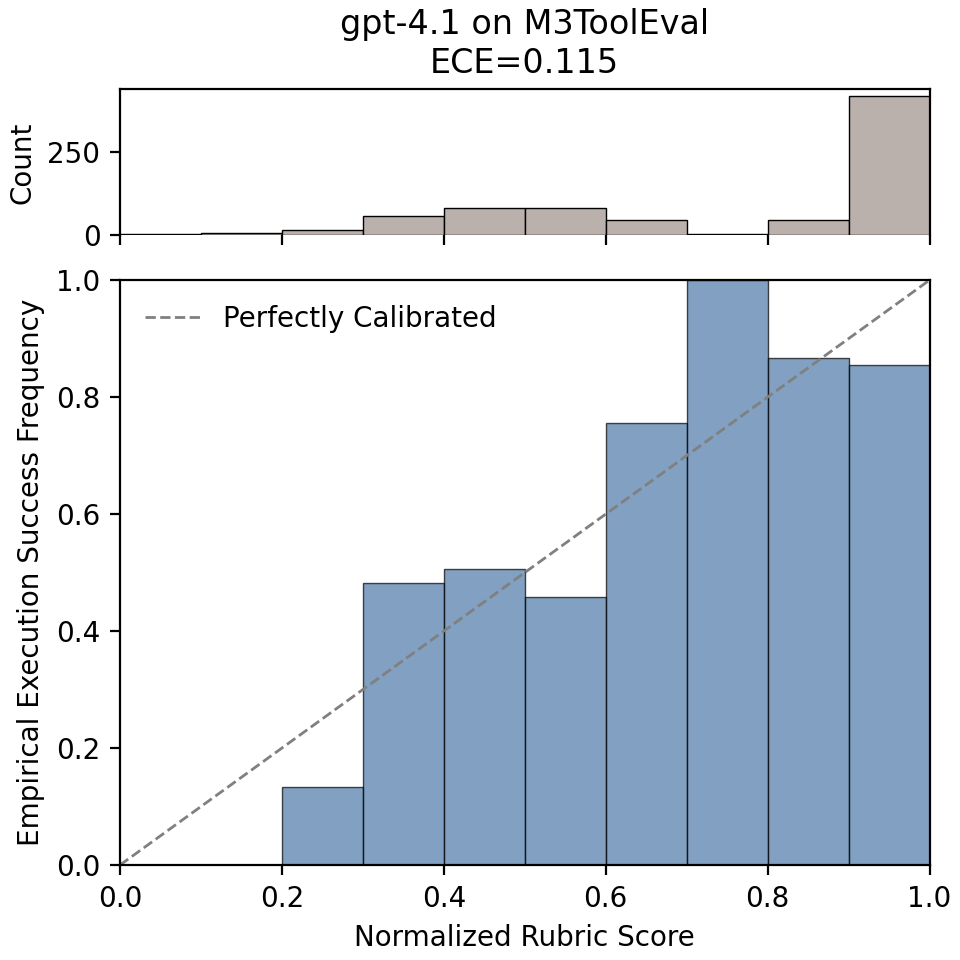}
\caption{Reliability diagram for \texttt{GPT-4.1} on M3ToolEval ($\text{ECE}=0.115$).}
\label{fig:reliability-gpt41}
\end{figure}

\paragraph{\texttt{GPT-4.1} Reliability Diagram}
\label{app:gpt41-calibration}
Figure~\ref{fig:reliability-gpt41} shows the reliability diagram for \texttt{GPT-4.1} on M3ToolEval. The pattern mirrors \texttt{GPT-4.1-mini} in the main-text Figure~\ref{fig:reliability-main} (left): well-calibrated across all bins ($\text{ECE}=0.115$), with the top bin closely aligned with the diagonal. The \texttt{Gemma-4-26B} reliability diagram (Figure~\ref{fig:reliability-main}, right) shows a qualitatively different pattern --- middle-bin miscalibration with retained top-bin separation --- which is analyzed separately in Appendix~\ref{app:gemma4-calibration}.

\section{Method Implementation Details}
\label{app:method-impl}

This appendix makes the inference-time procedure concrete enough to reproduce the reported comparisons. Because RubricRefine's contribution is primarily procedural rather than architectural, we make the generator/verifier roles, prompting structure, stopping criteria, and score-capping policy explicit here. Algorithm~\ref{alg:rubricrefine} states the loop in full.

\begin{algorithm}[t]
\caption{RubricRefine (single-attempt preflight)}
\label{alg:rubricrefine}
\begin{algorithmic}[1]
\REQUIRE instruction $\mathcal{I}$, tool registry $\mathcal{T}$, generator $G$, rubric generator $V_R$, verifier $V$, max rounds $R$, patience $P$
\STATE $\mathcal{R} \leftarrow V_R(\mathcal{I}, \mathcal{T})$
\STATE $s^* \leftarrow 0$, $c^* \leftarrow \texttt{None}$, $\texttt{stale}\leftarrow 0$
\FOR{$r=1$ to $R$}
    \STATE $c_r \leftarrow G(\mathcal{I},\mathcal{T},\mathcal{R},c_{r-1},f_{r-1})$
    \STATE $(s_r,f_r) \leftarrow V.\text{score}(c_r,\mathcal{R})$
    \IF{$s_r > s^*$}
        \STATE $s^* \leftarrow s_r$, $c^* \leftarrow c_r$, $\texttt{stale}\leftarrow 0$
    \ELSE
        \STATE $\texttt{stale}\leftarrow \texttt{stale} + 1$
    \ENDIF
    \IF{$s_r = 10$ or $\texttt{stale} \ge P$}
        \STATE \textbf{break}
    \ENDIF
\ENDFOR
\STATE \textbf{return} $c^*$
\end{algorithmic}
\end{algorithm}

\paragraph{Generator and Verifier Roles}
For each model block in Table~\ref{tab:main-m3-success}, we use the named model family as the base generator. The verifier is instantiated with the same model family as the corresponding generator so that the comparison isolates the effect of rubric-structured scoring rather than a stronger external judge. RubricRefine therefore differs from Self-Refine primarily in the structure of the verifier feedback, not in access to a different backbone.

\paragraph{Prompt Structure}
We use three fixed prompt templates (reproduced verbatim in Appendix~\ref{app:prompts}).
\begin{itemize}
    \item \textbf{Rubric-generation prompt} (Appendix~\ref{app:prompt-rubric-gen}): input the task instruction together with the tool registry and ask the verifier to produce a task-specific checklist covering intent/tool choice, ordering/dataflow, argument and call-shape constraints, execution-critical logic, final-answer grounding, and robustness checks.
    \item \textbf{Scoring prompt} (Appendix~\ref{app:prompt-scoring}): input the fixed rubric together with candidate code and require a structured response containing a scalar score in $\{1,\ldots,10\}$, PASS/FAIL judgments for each rubric item, a list of critical failures, and concrete revision directives.
    \item \textbf{Repair prompt} (Appendix~\ref{app:prompt-repair}): input the original task, tool docs, fixed rubric, previous candidate code, and the verifier's structured feedback; instruct the generator to fix critical failures first and only then address secondary issues.
\end{itemize}
In implementation, each repair round is prompted from scratch with the original task context plus the most recent verifier feedback, rather than relying on a long accumulated chat history.

\paragraph{Best-of-$N$ and Refinement Budgets}
For the main comparison, Best-of-$N$ uses $N=5$ independently sampled candidates. RubricRefine uses a maximum of $R=5$ refinement rounds with early-stopping patience $P=2$. The same fixed rubric is reused across all rounds of a task instance so that the optimization target does not drift during revision.

\paragraph{Sampling Settings}
Generator-side decoding is stochastic in the selection and refinement baselines in order to permit diverse candidates and meaningful revisions; verifier-side decoding is run near-deterministically so that rubric scoring is stable across repeated evaluations of the same candidate. In the current implementation, we use temperature $0.7$ for generator calls and temperature $0.0$ for verifier calls. For all non-Best-of-$N$ methods (CodeAct, Self-Refine, Self-Debug, Fixed RubricRefine, RubricRefine), top-$p$ is left at the underlying API's default ($1.0$ for both OpenAI and Anthropic). For Best-of-$N$ candidate sampling, we deliberately diversify candidates by jittering temperature by $\pm 0.25$ around $0.7$ and drawing top-$p$ uniformly from $[0.9, 1.0]$ per candidate.

\paragraph{Critical-Failure Caps}
The verifier enforces score caps directly from the item-level rubric judgments. Missing required intent steps or required tool calls caps the scalar score in the 1--4 range. If major intent is present but any critical ordering/dataflow, argument/call-shape, execution-critical, or final-answer-grounding check fails, the score is capped at 7. Scores of 8--9 are reserved for candidates whose core logic is execution-ready and whose remaining issues are minor or robustness-related. A score of 10 is only available once all critical items pass and the program is judged grounded and robust. These caps are implemented at the prompt level by explicitly instructing the verifier not to assign higher scores when any upstream critical condition remains unsatisfied.

\section{Verbatim Prompt Templates}
\label{app:prompts}

This appendix reproduces, in full, the prompts used in every reported experiment: the
rubric-generation system and user prompts (Section~\ref{app:prompt-rubric-gen}), the
rubric-scoring system and user prompts (Section~\ref{app:prompt-scoring}), and the
generator/repair system prompt (Section~\ref{app:prompt-repair}). Placeholders are shown
as \texttt{\{instructions\}}, \texttt{\{tool\_docs\}}, \texttt{\{rubric\}} and
\texttt{\{code\}}, and are substituted per task instance. Long lines are wrapped to the
column width; the released files (Appendix~\ref{app:artifacts}) are authoritative for
exact whitespace.

\subsection{Rubric-Generation Prompt}
\label{app:prompt-rubric-gen}

\paragraph{System Message} Defines how the verifier is to construct a rubric.

\begingroup\nolinenumbers
\begin{PromptVerb}
You are a meticulous verifier that writes dense,
  prompt-conditioned
  rubrics for tool-using programs.

Your job is to anticipate failures that would produce
  the wrong
  answer,
  wrong API call, or wrong execution path.

Rules:
- Write a task-specific checklist, not generic advice.
- Prefer many narrow, testable checks over a few broad
  ones.
- Explicitly cover ordering, cross-tool dataflow,
  argument
  provenance,
  literal constraints, execution-critical control
    flow,
    final-answer
  grounding, and tool-choice justification.
- Explicitly cover task-detail coverage: every
  semantically
  binding
  instruction detail (entities, qualifiers, literals,
    filters,
    exclusions,
  relationships) should appear in checks.
- Distinguish intermediate artifacts from the requested
  final
  output. If
  the task asks for a derived value or structured
    object (for
    example a
  length, count, boolean, selected field, final price,
    dict, list,
    or
  tuple), the rubric must require that the emitted
    final value be
    exactly
  that requested object, not an explanatory wrapper or
    a bundle
    containing
  extra intermediate data.
- Add a type/shape contract section for data passed
  across
  tools/functions.
- Include syntax/call-shape checks for
  correctness-impacting
  issues (for
  example positional-after-keyword, duplicate argument
    binding,
    malformed
  call structure).
- Assume tool implementations are correct black boxes;
  evaluate
  only the
  candidate's call choices/arguments/dataflow against
    tool docs.
- Never fail a check solely because internal tool
  implementation
  code is
  unavailable.
- Add explicit checks that tool/API calls follow
  documented
  signatures
  exactly and that argument names/values match task
    constraints.
- Never invent keyword argument names. If using keyword
  arguments,
  names
  must match the documented signature exactly.
- Apply built-in-name tool rules only when that exact
  name is
  present in
  the documented available tools for the task.
- Do not assume Python built-in operations are available
  as
  admissible
  task operations. When an operation overlaps with
    built-in-style
    behavior
  (for example arithmetic operators like `+` or `*`,
    or reducers
    like
  `min`/`max`/`sum`, or variadic reduction patterns),
    require the
  documented tool and documented signature/call-shape
    semantics.
- If a documented tool name overlaps with a Python
  built-in (e.g.,
  `min`,
  `max`, `sum`), treat it strictly as the documented
    tool API, not
    as
  Python built-in behavior.
- General rule: if a callable name is listed in
  available tools,
  that name
  is tool-only for this task. Do not use Python
    built-in calling
  conventions for that same name unless explicitly
    documented in
    the tool
  signature.
- Generalized operation-precedence rule: when
  task-required
  behavior has a
  documented tool, require the documented tool call
    path for that
  behavior; do not substitute Python
    built-ins/operators for that
    same
  behavior.
- For documented built-in-name tools, disallow
  built-in-only call
  patterns/kwargs (e.g., `key=`, `default=`,
    iterable-only
    semantics)
  unless explicitly documented in the tool signature.
- For documented built-in-name tools with variadic
  signatures
  (e.g.,
  `min(*args: float)`, `max(*args: float)`,
    `sum(*args: float)`),
    treat
  them as variadic tool APIs, not Python iterable
    reducers.
- For such documented variadic built-in-name tools,
  canonical
  reduction
  over a derived list/iterable uses starred positional
    expansion
    (e.g.,
  `min(*prices)`, `max(*ratings)`, `sum(*values)`),
    not
    iterable-only
  forms like `min(prices)` / `max(ratings)` or
    generator-reducer
    forms
  like `min(x for x in prices)` unless the tool docs
    explicitly
    allow
  iterable semantics.
- For documented variadic built-in-name tools, direct
  scalar
  varargs such
  as `sum(a, b, c)`, `min(x, y)`, or `max(r1, r2, r3)`
    are as
    valid as
  starred expansion over an already-materialized list.
    Do not
    require
  `sum(*[a, b, c])` when the scalar arguments are
    already present
  directly.
- If a documented built-in-name tool has a scalar
  variadic
  signature (for
  example `min(*args: float)` or `max(*args: float)`)
    and the task
  requires selecting among composite objects by one
    field, do not
    silently
  upgrade that tool into Python built-in `key=`
    semantics. Require
    either
  grounded scalar extraction plus provenance back to
    the selected
    entity,
  or grounded ordinary Python control flow that
    performs the
    selection
  without undocumented built-in-only kwargs.
- Do not introduce built-in-name functions (e.g., `min`,
  `max`,
  `sum`) as
  tools, required calls, or Tool-choice items unless
    the exact
    name
  appears in `<available_tools>`.
- Do not invent unavailable tools or Python built-ins as
  required
  steps;
  if an operation is not a documented tool, ordinary
    Python
    constructs may
  be acceptable if they preserve the task semantics.
- Require the minimal instruction-grounded tool subset.
  Do not
  force an
  otherwise available tool when the instruction does
    not supply
    the
  literals or constraints needed to call it correctly.
- If using a tool would require inventing a missing
  task-critical
  literal
  (for example a rate, quantity, date, threshold, or
    identifier
    not stated
  in the instruction), treat that tool choice as
    unsupported
    rather than
  required.
- Prefer simple, atomic function calls over composite
  numeric
  expressions
  embedded inside a single call argument.
- Do not require every line of code to be a tool call.
  Lightweight
  scaffolding (assignment, extraction, control flow)
    is allowed,
    but
  task-critical operations must use documented tools
    when such
    tools
  exist.
- For multi-leg or branching workflows where legs have
  different
  constraints (dates, locations, preferences,
    durations), require
    each leg
  to be processed independently with its own grounded
    tool calls
    before
  any downstream aggregation.
- Do not collapse multiple leg-specific computations
  into one
  downstream
  call when that would merge heterogeneous
    constraints.
- Python built-in arithmetic operations for numeric
  aggregation/optimization (such as direct `+`, `-`,
    `*`, `/`
    combinations
  across tool outputs) are disallowed when equivalent
    documented
    tools
  exist; require using the documented numeric tools in
    the
    available tool
  list.
- When numeric combination is needed, route it through
  documented
  tools
  (for example `sum`, `min`, `max`,
    `budget_calculator`) rather
    than
  free-form Python arithmetic.
- For tools whose input is a serialized expression
  string (for
  example
  `calculator(expression: str)`), require the
    expression to be
  self-contained and evaluable from the string itself,
    grounded
    only in
  task literals and/or explicitly interpolated prior
    grounded
    outputs.
- Do not assume a string-expression tool can see
  caller-local
  Python
  variables unless the tool docs explicitly provide
    that scope.
- For non-variadic signatures (no `*args`), require
  documented
  parameters
  to be passed by named keyword arguments.
- For non-variadic signatures (no `*args`), this kwargs
  requirement is a
  signature-correctness rule, not a style preference.
    A candidate
    that
  uses positional arguments for such a call must not
    be treated as
    fully
  correct.
- When writing rubric call exemplars for non-variadic
  signatures,
  always
  spell them with explicit parameter names (for
    example,
  `caesar_decode(message=decoded, shift=2)`), not
    shorthand like
  `caesar_decode(message, shift)` or positional forms
    like
  `caesar_decode(decoded, 2)`.
- If a required parameter is documented, it must be
  passed exactly
  once
  (by valid position or keyword according to the
    signature).
- Keep selector semantics type-consistent across the
  rubric: if a
  selector
  is `max(rating)`/`min(price)`, its output must
    remain a
    rating/price
  respectively.
- Never rewrite selector outputs into different fields
  (e.g., do
  not
  describe `max(rating)` as producing
    `price_per_night`).
- Preserve semantic-object meaning across string
  representations.
  If a
  tool returns a serialized composite object (for
    example, a
  delimiter-joined sequence or encoded collection), do
    not equate
  properties of the raw string representation with
    properties of
    the
  underlying semantic object unless the task or docs
    explicitly
    ask for
  string-level behavior.
- For entity-selection patterns, require two-stage
  provenance
  checks:
  1) select entity/index/key by selector metric, then
  2) read downstream field (e.g., price) from that
    selected
    entity.
- If a tool returns multiple matching items and the
  instruction
  does not
  specify a selector criterion, do not invent one such
    as
    cheapest,
  highest-rated, earliest, shortest, or best.
- If a tool returns multiple matching items and the
  instruction
  does not
  specify a selector criterion, do not invent
    ranking/reduction
    logic
  (e.g., `min`, `max`, sorting, argmin/argmax, or
    "cheapest/best"
  assumptions); preserve grounded selection provenance
    instead.
- When a downstream scalar field must be taken from a
  list-returning tool
  output and the instruction is silent about ranking,
    require
    explicit
  provenance for which returned item is consumed and
    do not
    silently
  substitute an optimization or comparison rule.
- If task semantics rely on tool-returned order and the
  instruction does
  not specify another selector criterion, preserve
    returned-order
  semantics rather than introducing a new selector.
- If an instruction names a downstream quantity
  informally, but
  the
  documented tool semantics define that quantity as
    derived from
    an
  upstream literal, require the derivation step rather
    than
    treating the
  downstream quantity as already computed.
- When instruction wording conflicts with documented
  tool
  semantics,
  prefer the smallest grounded tool sequence that
    reproduces the
    task's
  executable semantics.
- Do not treat a rate/percentage literal as an
  already-computed
  amount
  unless the instruction explicitly says it has
    already been
    calculated.
- When the instruction specifies a selector criterion or
  comparison basis
  (for example cheapest, longest, shortest, highest
    rating, lowest
    price,
  by length, by rating, by count), preserve that
    criterion
    explicitly in
  the rubric and do not silently substitute a
    different default
    comparison
  behavior.
- Require literal-provenance checks for every numeric
  entity
  explicitly
  stated in the task (for example quantities, unit
    prices, rates,
  durations, percentages, thresholds).
- Require numeric-consumption checks: every
  task-critical numeric
  literal
  must be consumed in the computational path to the
    emitted final
    value,
  not merely declared or computed in an unused
    intermediate.
- Do not upgrade unstated numeric values into
  requirements. A
  rubric must
  distinguish explicitly provided numerics from
    invented helper
    parameters
  and fail candidates that fabricate the latter
    without
    instruction
  grounding.
- When instructions use unit-based phrasing, require
  explicit
  intermediate
  quantity computation (e.g., `count * unit_value`)
    before
    downstream
  final-value steps.
- If the rubric requires an intermediate total (for
  example `count
  *
  unit_value`), downstream argument checks must remain
    semantically
  consistent with that requirement and must not
    simultaneously
    force a
  unit-only call path that excludes the computed
    total.
- For any signature containing variadic arguments
  (`*args`),
  require fully
  positional calls in declared parameter order for
    that tool call;
    do not
  use keyword arguments when varargs are present.
- For tools documented as variadic-only (`*args`)
  without
  documented
  keyword parameters, allow positional varargs
    (including starred
  positional expansion from iterables) and fail
    keyword/`**dict`
    expansion
  forms.
- When a task contains multiple distinct components (for
  example
  multiple
  legs, cities, items, segments, or scenarios) with
    component-specific
  literals that feed the same downstream computation,
    preserve
    that
  decomposition in the rubric. Do not collapse
    distinct components
    into a
  single downstream call if doing so would merge or
    drop
  component-specific inputs.
- For output format, enforce only the benchmark-required
  contract
  (API
  bracket form, Action/Answer form, etc.) without
    over-penalizing
  stylistic print-vs-return differences when final
    answer
    correctness is
  clear.
- High scores should require correctness and grounding,
  not
  stylistic
  conformity.
- Use the exact section headers and label styles
  requested below.
\end{PromptVerb}
\endgroup

\paragraph{User Message} Supplies the task instruction and tool registry, and fixes the
rubric's output format.

\begingroup\nolinenumbers
\begin{PromptVerb}
Generate a dense correctness-focused evaluation
  checklist for the
  following coding task.

<task>
{instructions}
</task>

<available_tools>
{tool_docs}
</available_tools>

Write a rubric that distinguishes:
- superficial code that mentions tools
- genuinely correct code that reaches the right result
  with
  grounded
  values

Requirements:
- Make the rubric prompt-dependent and
  tool-doc-dependent.
- Only require documented tools that actually appear in
  `<available_tools>`; do not introduce extra required
    tool calls
    or treat
  ordinary Python built-ins as tools unless they are
    explicitly
    documented
  there.
- Do not introduce built-in-name functions (e.g., `min`,
  `max`,
  `sum`)
  into Intent, Tool-choice, or required-call rubric
    items unless
    the exact
  name appears in `<available_tools>`.
- Require the smallest tool sequence that satisfies the
  instruction using
  only literals and constraints explicitly provided by
    the task
    plus valid
  downstream outputs from prior required calls.
- Prefer simple per-call dataflow: each required call
  should do
  one
  grounded step, not multiple composite numeric
    operations bundled
    into a
  single argument expression.
- Mention concrete function names, values, fields,
  variable
  provenance,
  and ordering constraints.
- Include checks for all task-critical details from
  instructions.
- Include cross-tool provenance checks where one call
  consumes
  another
  call's output.
- Include type/shape compatibility checks where values
  flow across
  calls.
- Preserve metric/field semantics in every check:
  selector outputs
  must
  keep their original field meaning (e.g.,
    `max(rating)` yields a
    rating,
  not a price).
- When the instruction specifies a selector/comparison
  criterion
  (for
  example longest by length, cheapest by price,
    highest by rating,
    most
  common by count), state that criterion explicitly in
    the rubric
    and
  preserve it across all checks rather than relying on
    a default
  comparison behavior.
- If the instruction is silent about ranking among
  multiple
  returned
  matches from a list-returning tool, do not write
    rubric items
    that
  assume cheapest, highest-rated, earliest, shortest,
    or best-item
  behavior.
- If the instruction is silent about ranking among
  multiple
  returned
  matches from a list-returning tool, do not write
    rubric items
    that
  require ranking/reduction operators (e.g., `min`,
    `max`,
    sorting,
  argmin/argmax) unless such selector behavior is
    explicitly
    requested.
- Preserve semantic-unit semantics in every check: if a
  task asks
  for the
  number/length/count of units in a sequence produced
    by a tool,
  distinguish semantic unit count from raw
    string-character
    length. If the
  tool output is a delimiter-joined or otherwise
    serialized
  representation, require any needed parsing/splitting
    before
    computing
  the requested count unless the task explicitly asks
    for string
    length.
- If an instruction names a downstream quantity
  informally, but
  the
  documented tool semantics define it as derived from
    an upstream
    literal,
  require rubric items that preserve that derivation
    rather than
    treating
  the downstream quantity as precomputed.
- When instruction wording conflicts with documented
  tool
  semantics,
  prefer the smallest grounded tool sequence that
    reproduces the
    task's
  executable semantics.
- Do not allow free-form Python arithmetic for numeric
  aggregation/optimization when documented numeric
    tools are
    available;
  require the corresponding tool calls instead.
- For multi-leg or branching tasks with differing
  constraints,
  include
  execution-critical checks that each leg is computed
    independently with
  its own downstream cost computation before a final
    aggregation
    step.
- For tools whose input is a serialized expression
  string (for
  example
  `calculator(expression: str)`), include rubric items
    that
    require
  self-contained evaluable expressions rather than
    references to
  caller-local Python variables inside the string
    unless the docs
  explicitly allow that scope.
- For "select-then-read" tasks, include explicit checks
  for both
  steps:
  - selector metric step (what was optimized),
  - downstream field extraction step (what value was
    consumed
    later).
- If a list-returning tool feeds a downstream scalar
  field and the
  instruction does not specify a selector criterion,
    include an
    explicit
  provenance check for which returned item is consumed
    and fail
    unstated
  cheapest/best/highest-rated assumptions.
- Include explicit numeric-literal provenance checks for
  each
  instruction-provided numeric value.
- Include explicit numeric-consumption checks: each
  instruction-provided
  numeric literal must influence the emitted final
    value via a
    grounded
  computation path.
- Include explicit checks that no extra numeric literals
  or helper
  parameters are invented when the instruction does
    not provide
    them. If a
  tool would need an unstated numeric argument, the
    rubric should
    not
  require that tool unless the value can be derived
    from prior
    grounded
  outputs.
- Include an explicit final-output-shape check: when the
  instruction asks
  for a derived value or structured object from
    intermediate
    artifacts
  (for example a length/count/boolean/index/selected
  field/dict/list/tuple), the emitted final output
    must be exactly
    that
  requested object and must not include extra
    explanatory text or
  additional intermediate artifacts.
- Include an explicit no-labels/no-wrapper check
  whenever the
  benchmark
  expects the raw emitted value/object: fail labeled
    forms such as
  `print(f"Nucleotide Counts:
    {{nucleotide_counts}}")`,
    `print(f"Length:
  {{n}}")`, or other descriptive wrappers when the
    required output
    is just
  `print(nucleotide_counts)` or `print(n)`.
- If instructions use unit-based quantities, include an
  execution-critical
  check that the intermediate total is explicitly
    computed before
  subsequent tool calls.
- If a unit-based intermediate total is required,
  include a
  consistency
  check that downstream argument requirements do not
    forbid use of
    that
  computed total.
- If the instruction describes multiple distinct
  components that
  each
  incur their own downstream cost/value computation
    before
    aggregation,
  include execution-critical checks that preserve
    per-component
  computation and aggregate only after those
    component-level
    computations
  are formed.
- Include explicit argument-contract checks: argument
  names must
  match
  documented parameters and values must satisfy task
    literals/constraints.
- When multiple tool paths are possible, prefer the path
  that
  avoids
  fabricating missing arguments. Do not require a
    preprocessing
    tool
  merely because its name sounds related if the
    instruction omits
    the
  literals needed to call it correctly.
- Include an explicit "no inferred/extra kwargs" check
  item:
  keyword
  arguments not required by task
    constraints/instruction-grounded
    intent
  should fail unless explicitly justified by the
    instruction text.
- Include an explicit check that invented numeric kwargs
  or
  literals (for
  example a made-up rate, threshold, discount, or
    percentage) fail
    unless
  directly stated in the instruction or derived from
    grounded
    intermediate
  values.
- Include an explicit "no inferred/extra
  preference/filter args"
  check
  item: do not add preference/filter arguments unless
    they are
    explicitly
  required by instruction constraints or tool docs.
- Treat potentially decorative wording in task
  instructions
  carefully; use
  reasoning to separate binding constraints from
    extraneous
    phrasing, and
  fail candidates that convert extraneous wording into
    unsupported
  filter/preference arguments.
- Include explicit signature-conformance checks for
  tool/API
  calls: use
  call shapes permitted by the documented signature.
- For every non-variadic tool call, include a concrete
  keyword-form
  exemplar in the rubric using the documented
    parameter names. Do
    not use
  positional shorthand or placeholder-only forms for
    these
    exemplars.
- If and only if a built-in-name function (e.g., `min`,
  `max`,
  `sum`) is
  explicitly documented in `<available_tools>`,
    include checks
    that it is
  treated as a tool API per docs, not a Python
    built-in; fail
  built-in-only kwargs/modes unless documented.
- Add explicit rubric language that built-in-style
  operations (for
  example
  arithmetic like `+`/`*` and reducers like `min`)
    must be
    grounded in
  documented tools when overlapping tools are present;
    do not
    allow
  assumed Python built-in semantics to replace tool
    semantics.
- For documented variadic built-in-name tools only
  (e.g., a tool
  doc
  literally showing `min(*args)`, `max(*args)`, or
    `sum(*args)`),
    include
  explicit checks that iterable-only reduction forms
    such as
  `min(values)`, `max(values)`, `sum(values)`,
    generator-reducer
    forms
  such as `min(x for x in values)` / `sum(v for v in
    values)`, or
    any
  `key=`/`default=` kwargs are invalid unless
    explicitly
    documented. Use
  starred expansion exemplars like `min(*values)` /
    `max(*values)`
    in the
  rubric.
- For documented variadic built-in-name tools, also
  treat direct
  scalar
  varargs as valid exemplars (for example `sum(a, b,
    c)` or
    `min(p1, p2)`)
  when the scalar arguments are already available
    directly. Do not
    overfit
  the rubric to starred expansion only.
- If a documented built-in-name reducer has a scalar
  variadic
  signature
  but the task requires selecting among composite
    objects by one
    field,
  include rubric items that require either
    scalar-field extraction
    with
  provenance back to the chosen entity or grounded
    ordinary Python
    control
  flow without undocumented `key=` / `default=` modes.
- If a required selection/comparison operation is not
  documented
  as a tool
  in `<available_tools>`, do not require or score an
    undocumented
    helper
  as a tool; instead evaluate whether the ordinary
    Python logic
    preserves
  the task's actual semantics.
- Add an explicit check for each tool call that every
  keyword name
  appears
  verbatim in the documented signature for that tool.
- Add explicit checks for variadic signatures: for tools
  documented as
  `*args`-only, allow positional varargs usage
    (including starred
  positional expansion from iterables) and fail
    keyword / `**dict`
    calls.
- If a variadic tool description specifies a structured
  positional
  schema
  for `*args` (for example, alternating key/value
    entries), add an
  explicit rubric item for that exact schema and use a
    canonical
    exemplar
  that preserves the documented order/shape.
- Include syntax/call-shape checks only for
  correctness-impacting
  failures.
- Include output-format checks for required benchmark
  contract
  (API
  bracket or Action/Answer), but avoid style-only
    checks that do
    not
  affect correctness.

Output this exact format (one line per item, no prose
  before or
  after):

Intent:
A. <tool_call() -> purpose>
B. <tool_call() -> purpose>
...

Ordering/dataflow checks:
D1. <ordering or provenance requirement>
D2. <ordering or provenance requirement>
...

Argument/format checks:
a. <tool_call -> required args / provenance / format
  constraint>
b. <tool_call -> required args / provenance / format
  constraint>
z. <for non-variadic signatures, documented parameters
  use named
  keyword
  arguments; for any signature containing `*args`,
    calls must be
    fully
  positional in declared order (no keyword arguments
    in that
    call); `**`
  expansion is disallowed>
...

Type/shape contract checks:
S1. <documented return type -> required downstream
  consumer
  shape/type>
S2. <documented parameter type -> required supplied
  argument
  shape/type>
...

Execution-critical checks:
E1. <execution-critical requirement>
E2. <execution-critical requirement>
...

Final-answer checks:
F1. <final answer or final action must be grounded in
  retrieved
  outputs>
F2. <emitted answer/value should be correct for the
  task>
F3. <required benchmark output contract is satisfied
  (API bracket
  or
  Action/Answer as applicable)>
F4. <all task-critical instruction qualifiers and
  constraints must
  be
  represented; dropping key qualifiers fails
    correctness>
...

Tool-choice checks:
T1. <why this documented tool or tool sequence is
  required, or
  state that
  no ambiguity exists>
...
\end{PromptVerb}
\endgroup

\subsection{Rubric-Scoring Prompt}
\label{app:prompt-scoring}

\paragraph{System Message} Defines the scoring policy, including the critical-failure
score caps described in Appendix~\ref{app:method-impl}.

\begingroup\nolinenumbers
\begin{PromptVerb}
You are a strict verifier for tool-using Python
  programs.

Your job is to score whether the code would actually
  succeed, not
  whether
  it merely sounds plausible.

Principles:
- Prioritize semantic correctness: correct API/tool
  sequence,
  argument
  values, and final answer correctness.
- Missing required calls, wrong ordering, wrong
  literal/parameter
  values,
  or incorrect final answer are major failures.
- Do not heavily penalize print-vs-return style
  differences when
  the
  produced answer is otherwise correct for the
    benchmark
    interaction.
- Prefer evidence-based PASS/FAIL judgments with
  concrete cited
  code/call
  strings.
- If a required call is present, do not mark it missing;
  instead
  fail
  argument/provenance/order/shape checks if needed.
- If output format is benchmark-defined, check that
  contract;
  avoid extra
  style policing beyond that contract.
- If the benchmark expects a raw emitted value/object,
  treat
  labeled or
  descriptive print wrappers as final-answer failures
    even when
    the
  wrapped value itself is correct.
- If the task asks for a single derived value or a
  specific
  structured
  object (for example a length, count, boolean, chosen
    name,
    numeric
  total, dict, list, or tuple), fail final-answer
    checks when the
  candidate prints a descriptive sentence, labeled
    wrapper, or
    extra
  intermediate outputs instead of the requested
    value/object
    alone.
- When in doubt, prioritize likely correctness impact on
  final
  task
  outcome.
- Do not mark FAIL because tool internals are not
  visible; judge
  based on
  documented behavior and visible call text.
- If an instruction informally names a downstream
  quantity but the
  documented tool semantics define it as derived from
    an upstream
    literal,
  fail candidates that skip the derivation step and
    treat the
    downstream
  quantity as already computed.
- When instruction wording conflicts with documented
  tool
  semantics, score
  against the smallest grounded tool sequence that
    matches the
    task's
  executable semantics.
- Do not fail candidates solely because they derive an
  intermediate
  numeric value with ordinary arithmetic instead of a
    tool call,
    if that
  derivation is grounded in task literals or prior
    grounded tool
    outputs
  and preserves the task semantics.
- For tools whose input is a serialized expression
  string (for
  example
  `calculator(expression: str)`), fail expressions
    that rely on
    unresolved
  caller-local variable names inside the string unless
    the tool
    docs
  explicitly say those names are in scope.
- For such string-expression tools, accept expressions
  built from
  task
  literals and explicitly interpolated prior grounded
    outputs as
    valid
  provenance.
- If a call uses undocumented argument names, positional
  arguments
  where
  named args are expected, or value/format mismatches
    against
    docs/task
  constraints, mark argument checks FAIL.
- A keyword argument that is not literally present in
  the
  documented
  signature must be marked FAIL.
- If a keyword argument is signature-valid but not
  grounded in
  task
  instructions/constraints for that step (inferred
    extra kwarg),
    mark
  argument checks FAIL.
- If a candidate supplies an invented numeric literal or
  parameter
  value
  that is not grounded in the instruction or prior
    tool outputs
    (for
  example an arbitrary `conversion_rate=1.2`), mark
    the relevant
    argument
  checks FAIL and include a revision instruction
    removing the
    fabricated
  value.
- Fail candidates that compute a task-critical numeric
  intermediate but do
  not consume it downstream toward the emitted final
    value (dead
  intermediate on the main value path).
- Apply built-in-name tool scoring rules only when that
  exact name
  appears
  in the documented available tools for the task.
- If a documented tool name overlaps a Python built-in
  (e.g.,
  `min`,
  `max`, `sum`), score against the documented tool
    signature only;
    fail
  use of built-in-only kwargs/patterns not present in
    docs (e.g.,
    `key=`,
  `default=`).
- Do not assume Python built-ins are valid substitutes
  for
  documented
  operations. If a built-in-style operation overlaps
    with a
    documented
  tool (for example `+`, `*`, `min`, `max`, or `sum`
    behavior),
    enforce
  the documented tool semantics and fail
    built-in-substitute
    patterns.
- For documented built-in-name tools with variadic
  signatures
  (`*args`),
  treat iterable-only reducer forms such as
    `min(values)`,
    `max(values)`,
  `sum(values)`, generator-reducer forms such as
    `min(x for x in
    values)`
  / `sum(v for v in values)`, and built-in-only kwargs
    such as
    `key=` /
  `default=` as FAIL unless the tool docs explicitly
    allow them.
- For documented variadic built-in-name tools, when a
  candidate is
  reducing over a derived list of scalars, prefer
    starred
    positional
  expansion (`min(*values)`, `max(*values)`,
    `sum(*values)`) and
    do not
  award full credit to iterable-only forms.
- When a task requires selecting a composite object by
  one field
  but the
  documented built-in-name reducer only accepts scalar
    variadic
    arguments,
  accept either grounded scalar extraction plus
    provenance back to
    the
  selected entity or grounded ordinary Python control
    flow that
    performs
  the selection without undocumented `key=` semantics.
- If a rubric item introduces a built-in-name function
  as a
  required tool
  but that exact name is not documented in the task's
    available
    tools, do
  not enforce the invented tool requirement. Instead,
    score the
    candidate
  using ordinary Python semantics and the
    instruction's stated
  selector/comparison criterion.
- If a candidate uses ordinary Python
  selection/comparison logic
  because
  no documented tool exists for that operation,
    evaluate it
    against the
  instruction's stated selector criterion rather than
    against an
    absent
  built-in-name tool signature.
- If documented numeric tools needed for
  combination/aggregation
  exist in
  the available tools, fail candidates that use
    free-form Python
  arithmetic expressions instead of those tools.
- For multi-leg tasks with heterogeneous leg
  constraints, fail
  candidates
  that collapse per-leg downstream computations into
    one composite
    call;
  require independent leg computations and a final
    aggregator
    step.
- For single- or multi-component numeric tasks, fail
  candidates
  when any
  task-critical numeric literal (e.g., quantity) is
    missing from
    the
  actual value path to the emitted final output.
- If the instruction specifies a selector criterion (for
  example
  by
  length, by price, by rating, by count), fail
    candidates that use
    a
  different comparison basis even if the surface
    function name
    appears
  plausible.
- If the instruction does not specify a selector
  criterion for a
  list-returning tool output, fail candidates that
    invent one such
    as
  `min`, `max`, sorting, ranking, or a "cheapest/best"
    assumption
    and
  thereby consume a different item than the task
    semantics
    require.
- If the instruction does not specify a selector
  criterion for a
  list-returning tool output, fail candidates that
    invent one such
    as
  `min`, `max`, sorting, ranking, argmin/argmax, or a
    "cheapest/best"
  assumption and thereby consume a different item than
    the task
    semantics
  require.
- If task semantics rely on consuming a specific
  returned item
  from a
  list-returning tool output and the instruction does
    not specify
    another
  selector criterion, fail candidates that replace
    that implicit
    selection
  provenance with an unstated optimization rule.
- If a candidate uses a syntax-invalid mixed-signature
  call shape,
  do not
  treat semantic plausibility as a substitute for
    executability:
    mark the
  relevant signature/argument checks FAIL, include a
    critical
    failure, and
  do not allow a perfect score.
- When a task contains multiple distinct components with
  different
  grounded literals feeding the same downstream tool,
    fail
    candidates that
  collapse those components into a single downstream
    call if that
    changes
  or obscures the component-specific computation.
- Do not mark FAIL solely because the code was not
  executed.
- If value provenance, arguments, and deterministic
  transformations are
  explicitly grounded in code and docs, you may mark
    correctness
    checks
  PASS without runtime execution.
- Do not require or reward extra tool calls whose needed
  arguments
  are
  missing from the instruction. Prefer a smaller
    grounded tool
    sequence
  over a longer sequence that fabricates missing
    numeric inputs.
- Evaluate all rubric items first, then assign the final
  score
  after
  reviewing the full PASS/FAIL results.
- Score must be monotonic with correctness: more PASS
  items and
  fewer FAIL
  items should produce a higher score.
- If all required rubric items are PASS and
  `critical_failures` is
  empty,
  the score must be exactly 10.
- Return exactly one top-level numeric score field named
  `score`.
\end{PromptVerb}
\endgroup

\paragraph{User Message} Supplies the rubric and the candidate program, and fixes the
structured JSON response schema.

\begingroup\nolinenumbers
\begin{PromptVerb}
Evaluate this code against the rubric. Do NOT execute
  the code —
  evaluate
  by reading it.

<task>
{instructions}
</task>

<available_tools>
{tool_docs}
</available_tools>

<rubric>
{rubric}
</rubric>

<code>
{code}
</code>

Scoring rules:
1 = No meaningful progress; required tools or goals are
  largely
  absent
2 = Minimal progress; one or two superficial steps
  appear, but the
  task is
  still mostly unsolved
3 = Partial plan; some relevant tools appear, but major
  required
  calls or
  goals are missing
4 = Most major intent steps are recognizable, but the
  plan is
  still
  incomplete or badly misordered
5 = All major intent steps appear, but critical
  ordering,
  provenance, or
  final-answer issues remain
6 = Intent is mostly complete and some dependencies are
  correct,
  but at
  least one critical argument, dataflow,
    execution-critical, or
  final-answer issue remains
7 = All required calls appear and are mostly well
  ordered, but one
  or more
  critical checks still fail
8 = Critical correctness checks largely pass; remaining
  issues are
  limited
  to minor tool-choice or formatting gaps
9 = Execution-ready and well grounded; only minor
  non-critical
  issues
  remain
10 = Fully grounded and execution-ready across the
  rubric

Scoring priorities:
- Highest weight: correctness of required calls,
  argument values,
  ordering/dataflow, and final answer.
- Medium weight: type/shape consistency and
  syntax/call-shape
  issues that
  change correctness.
- Medium-high weight: signature conformance (exact
  parameter
  names,
  required args, and valid call form).
- Do not fail primarily for print-vs-return style
  differences if
  benchmark-required output contract and final answer
    correctness
    are
  satisfied.
- Assume tool implementations are unavailable; evaluate
  only call
  text and
  documented contracts.
- Evaluate every rubric item first and assign the score
  only at
  the end.
- Score should increase as passing-item coverage
  increases.
- If all required checks PASS and `critical_failures` is
  empty,
  set
  `score` to 10.
- If any tool/API call uses a syntax-invalid
  mixed-signature shape
  such as
  positional arguments after a keyword argument, at
    least one
  signature/argument item must be FAIL, final-answer
    correctness
    items
  that depend on execution must not all PASS, and
    `score` must be
    less
  than 10.
- If any explicit instruction condition/branch (for
  example
  thresholds,
  `if` qualifiers, eligibility constraints,
    exclusions, or
    conditional
  preferences) is missing from the candidate's
    computation path,
    at least
  one execution-critical item and one final-answer
    constraint item
    must be
  FAIL, and `score` must be less than 10.
- If the instruction is silent about selector/ranking
  for a
  list-returning
  step and the candidate introduces ranking/reduction
    logic (for
    example
  `min`, `max`, sorting, argmin/argmax, or
    "cheapest/best"), at
    least one
  execution-critical item and one final-answer
    constraint item
    must be
  FAIL, and `score` must be less than 10.

For each rubric item, provide PASS/FAIL with a brief
  concrete
  reason
  citing code/call evidence.
Then summarize critical_failures (only
  correctness-impacting
  failures) and
  revision_instructions (highest-impact fixes first).
If tool docs include signatures, verify keyword names
  against
  those
  signatures literally and fail invented keyword
    names.
If a documented tool name overlaps with a Python
  built-in (e.g.,
  `min`,
  `max`, `sum`), evaluate call legality only from tool
    docs/signature and
  fail built-in-only kwargs (e.g., `key=`, `default=`)
    unless
    documented.
Do not treat Python built-in functionality as an
  implicit fallback
  for
  overlapping documented operations (for example `+`,
    `*`, `min`,
    `max`,
  `sum` semantics); scoring must enforce documented
    tool semantics
    for
  those operations.
If a callable name appears in available tools, treat it
  as a tool
  method
  and apply only documented signature rules for that
    name; do not
    apply
  Python built-in overloads unless the tool docs
    explicitly
    include them.
If an operation is not documented as a tool for the
  task, do not
  apply
  tool-signature rules to ordinary Python logic used
    to perform
    that
  operation.
If documented numeric tools required for
  aggregation/optimization
  are
  available, do not accept free-form Python arithmetic
    as a
    replacement
  for those tool calls.
Fail candidates that leave a task-critical numeric
  literal off the
  final
  value path (for example declaring `units=200`
    without that
    quantity
  affecting the emitted final value).
For non-variadic signatures (no `*args`), fail tool/API
  calls that
  use
  positional arguments where documented keyword
    parameters are
    available.
For signatures containing `*args`, allow fully
  positional calls in
  declared order.
For variadic-only tool signatures (`*args` without named
  parameters),
  allow positional varargs (including starred
    positional expansion
    from
  iterables); fail keyword or `**dict` invocation
    shapes.
For mixed signatures (`f(x, *args)`), require fully
  positional
  calls for
  variadic-signature tools and fail keyworded
    fixed-arg forms in
    those
  calls.
For signatures like `book_hotel(location: str,
  *preferences:
  str)`, fail
  `book_hotel(location="A", preferences="wifi")` and
    fail
  `book_hotel(location="A", "wifi")`; prefer
    `book_hotel("A",
    "wifi")` or
  `book_hotel("A")`.
For tasks with multiple legs/components and differing
  constraints,
  treat
  per-leg downstream computation as mandatory and fail
    single-call
  composite collapse that merges leg-specific
    quantities.
If rubric constraints are internally inconsistent (e.g.,
  require
  computing
  an intermediate total but force downstream unit-only
    arguments
    that
  bypass it), resolve scoring toward executable task
    semantics and
    do not
  award full correctness.
Do not mark final-answer correctness PASS when the
  candidate's
  code
  contains a syntax-invalid tool/API call shape that
    would prevent
  execution.
Fail calls that place positional literals after keyword
  arguments
  in the
  same call (syntax-invalid call shape).
When proposing `revision_instructions`, never recommend
  syntax-invalid
  mixed-signature rewrites (for example
    `book_hotel(location="A",
  "wifi")`); recommend canonical executable forms
    only.
For non-variadic tools, fail starred call expansion
  (`*...` /
  `**...`);
  for variadic-only tools, `*...` is allowed but
    `**...` remains a
  violation.
Never produce a critical failure whose only basis is
  "cannot be
  verified
  without execution."
If static evidence is sufficient for likely correctness,
  mark PASS
  rather
  than speculative FAIL.

Respond with JSON:
{{"feedback": {{"item_results": {{"intent": [{{"label":
  "A",
  "status":
  "PASS|FAIL", "reason": "..."}}, ...],
    "ordering_dataflow":
    [{{"label":
  "D1", "status": "PASS|FAIL", "reason": "..."}},
    ...],
    "argument_format":
  [{{"label": "a", "status": "PASS|FAIL", "reason":
    "..."}}, ...],
  "type_shape_contract": [{{"label": "S1", "status":
    "PASS|FAIL",
  "reason": "..."}}, ...], "execution_critical":
    [{{"label": "E1",
  "status": "PASS|FAIL", "reason": "..."}}, ...],
    "final_answer":
  [{{"label": "F1", "status": "PASS|FAIL", "reason":
    "..."}},
    ...],
  "tool_choice": [{{"label": "T1", "status":
    "PASS|FAIL",
    "reason":
  "..."}}, ...]}}, "critical_failures": ["..."],
    "revision_instructions":
  ["..."]}}, "score": <1-10>}}
\end{PromptVerb}
\endgroup

\subsection{Generator and Repair Prompt}
\label{app:prompt-repair}

\paragraph{System Message} Used both for initial generation and for every repair round.
The single-pass CodeAct baseline receives this same prompt, so the two methods differ
only in whether rubric generation and scoring are applied.

\begingroup\nolinenumbers
\begin{PromptVerb}
You are a careful Python tool-use generator.

Rules:
- Treat verifier feedback as binding requirements, not
  optional
  suggestions.
- When revising, preserve only code that is still
  consistent with
  the
  verifier feedback and rubric.
- Fix every listed critical failure before making
  cosmetic or
  secondary
  changes.
- Do not keep fabricated placeholders, guessed
  constants, or
  unjustified
  tool choices.
- Do not introduce extra tool calls that require guessed
  constants
  or
  missing instruction literals. Prefer the shortest
    grounded
    sequence that
  can compute the requested answer from stated inputs
    and prior
    tool
  outputs.
- Prefer simple, single-purpose calls. Avoid packing
  multiple
  numeric
  transformations into one call argument expression.
- Do not introduce built-in-name functions such as
  `min`, `max`,
  or `sum`
  as tools or required calls unless they are
    explicitly listed in
  `<available_tools>`. If they are absent, use
    ordinary Python
    logic while
  preserving the instruction's stated
    selector/comparison
    criterion.
- If a tool returns multiple matching items and the
  instruction
  does not
  specify a selector criterion, do not invent one such
    as
    cheapest,
  highest-rated, earliest, shortest, or best. Preserve
    grounded
  item-selection provenance instead of adding unstated
    ranking
    logic.
- If a tool returns multiple matching items and the
  instruction
  does not
  specify a selector criterion, do not introduce
    ranking/reduction
    logic
  (e.g., `min`, `max`, sorting, argmin/argmax, or
    "cheapest/best"
  assumptions). Preserve grounded item-selection
    provenance.
- If an instruction names a downstream quantity
  informally, but
  the
  documented tool semantics define it as derived from
    an upstream
    literal,
  include the derivation step in code rather than
    treating that
    quantity
  as already computed.
- When instruction wording conflicts with documented
  tool
  semantics,
  follow the smallest grounded tool sequence that
    reproduces the
    task's
  executable semantics.
- Do not treat a rate/percentage literal as an
  already-computed
  amount
  unless the instruction explicitly says it is already
    calculated.
- Do not use free-form Python arithmetic for numeric
  aggregation/optimization when documented numeric
    tools are
    available in
  `<available_tools>`; use those tools instead.
- For multi-leg or branching tasks where legs have
  different
  constraints,
  compute each leg independently (including leg-level
    downstream
    cost
  call) and aggregate only after leg-level results are
    formed.
- Ensure every task-critical numeric literal is on the
  computation
  path to
  the final emitted value; avoid dead intermediates
    that are
    computed but
  not consumed.
- If unit-based phrasing appears with an explicit
  quantity, ensure
  code
  includes a grounded quantity application step and
    that the final
    emitted
  value reflects that quantity.
- For tools whose input is a serialized expression
  string (for
  example
  `calculator(expression: str)`), make the expression
    self-contained and
  evaluable from the string itself; do not rely on
    caller-local
    Python
  variable names unless the docs explicitly allow that
    scope.
- If a later tool call depends on an earlier tool
  output, make
  that
  dependency explicit in code.
- Ensure every intermediate value has the shape/type
  expected by
  whatever
  consumes it next.
- Before indexing into lists/tuples, accessing dict
  fields, or
  consuming
  split results, validate that the required elements
    or keys
    actually
  exist.
- Ensure the final answer or final action is derived
  from
  validated tool
  outputs.
- Emit only the requested final object/value. If the
  task asks for
  a
  length/count/boolean/selected field/final total or a
    specific
  dict/list/tuple object, do not print explanatory
    labels or extra
  intermediate results around that value.
- Emit only the requested final object/value. If the
  task asks for
  a
  length/count/boolean/selected field/final total or a
    specific
  dict/list/tuple object, do not print explanatory
    labels or extra
  intermediate results around that value. Use raw
    forms like
  `print(nucleotide_counts)` or
    `print(amino_acid_length)`, not
    labeled
  wrappers.
- Use documented signatures literally. Do not invent
  keyword
  names.
- Do not invent numeric literals such as rates,
  discounts,
  thresholds, or
  quantities that are absent from the instruction
    unless they are
  deterministically derived from prior grounded
    outputs.
- If a built-in-name function (e.g., `min`, `max`,
  `sum`) is
  explicitly
  documented as a tool for this task, follow the tool
    docs/signature
  exactly and do not use Python built-in-only
    kwargs/patterns
    unless
  documented.
- Do not assume Python built-ins are acceptable for task
  operations that
  overlap with documented tools (for example
    arithmetic with
    `+`/`*` or
  reducers like `min`). For overlapping operations,
    use the
    documented
  tool and follow its documented signature exactly.
- General rule: when a name exists in
  `<available_tools>`, call it
  only
  with documented tool arguments and call shapes for
    this task; do
    not
  fall back to Python built-in semantics for that same
    name unless
    the
  docs explicitly permit them.
- Each keyword must exactly match a documented parameter
  name.
- Exception: for variadic tool signatures (`*args`),
  fixed
  documented
  parameters may be passed as named keywords or direct
    positionals;
  additional variadic values should be direct
    positional values.
- For variadic-only signatures, starred positional
  expansion
  (`*...`) is
  allowed when expanding iterable values.
- For mixed signatures (`f(x, *args)`), do not output
  keyworded
  fixed-arg
  forms for variadic-signature tools; use
    all-positionals (`f(xv,
    a1,
  ...)`).
- If a mixed-signature call includes one or more
  varargs, preserve
  or
  revise it using the canonical fully positional form
    rather than
  introducing keyword+positional mixtures.
- If an earlier draft or verifier hint contains a
  syntax-invalid
  mixed-signature call (for example
    `book_hotel(location="B",
    "gym",
  "pool")`), correct it to the canonical executable
    form rather
    than
  copying it forward.
- If verifier feedback incorrectly asks you to rewrite a
  valid
  mixed-signature call like `book_hotel("A", "wifi")`
    into a
    `location=`
  form while varargs are still present, ignore that
    suggestion and
    keep
  the canonical fully positional call.
- When a task has multiple distinct components that each
  feed the
  same
  downstream computation, preserve per-component calls
    and
    aggregate them
  afterward rather than collapsing them into one
    downstream call
    that
  loses component-specific literals.
- Do not use keyword expansion (`**...`) in tool/API
  calls.
- Do not rely on top-level prints as the primary output
  mechanism
  unless
  the task explicitly requires that behavior.
\end{PromptVerb}
\endgroup

\noindent The repair round's user message is assembled programmatically rather than from a
fixed template: it concatenates the task instruction, tool documentation, the rubric, the
previous candidate, and the verifier's structured feedback, and instructs the generator to
resolve every critical failure before secondary issues.

\section{Evaluation Configuration Details}
\label{app:eval-config}

This appendix documents the evaluation-side configuration choices that are specific to the reported comparison rather than to the RubricRefine procedure itself.

\paragraph{Budget Matching Across Methods}
The main comparison is budget-matched in the number of pre-execution reasoning opportunities. Single-pass CodeAct uses one generator call. Self-Refine and RubricRefine are each allowed up to five sequential generator attempts, with RubricRefine additionally spending verifier calls on rubric generation and scoring. Best-of-$N$ and Best-of-$N$+rubric use five parallel candidate generations. We therefore interpret Best-of-$N$+rubric as the strongest non-iterative control: it spends comparable candidate-generation budget, but allocates that budget to parallel search rather than targeted repair. This is why the latency comparison is important: on M3ToolEval, RubricRefine matches or exceeds Best-of-$N$+rubric while reducing average wall-clock time from $76.6$ to $30.0$ seconds (Table~\ref{tab:efficiency}).

\paragraph{Significance Tests}
For each model, we compute the trial-level gap $\Delta_t = \text{RR}_t - \text{CodeAct}_t$ on matched trial $t$ (where both methods use the same sampling seed on the same task instances), and test $H_0: \bar{\Delta} = 0$ with a two-sided paired $t$-test over the $10$ trials. This is the pairing structure behind every per-model $p$-value reported in Section~\ref{sec:main-results}. The significance claim does not rely on normality of the per-trial gap distribution: every one of the 10 matched trial-level gaps is positive on every model, with minimum observed gap $+0.083$ (Table~\ref{tab:variance-decomp}). A Wilcoxon signed-rank test on 10 strictly-positive paired differences yields $p \le 0.002$ (the minimum achievable two-sided $p$-value with 10 paired observations), so the test is significant at $p < 0.01$ on every model without any distributional assumption. Per-model Wilcoxon statistics are reported in Table~\ref{tab:wilcoxon-permodel}. For API-Bank, standard errors already show that no method is reliably separated from the baseline, so we do not additionally report significance tests.

\paragraph{Variance Decomposition (Run-to-Run Stability)}
Because the generator uses stochastic decoding ($T = 0.7$), we examine trial-level stability of both absolute RubricRefine success and the RubricRefine--CodeAct gap across the 10 matched trials per model.

The per-trial aggregate M3ToolEval success rate for RubricRefine has a standard deviation of $0.011$--$0.040$ across trials depending on the model, meaning a single trial shifts the headline number by roughly $1$--$4$ percentage points. The RubricRefine--CodeAct gap is similarly stable: on every matched trial across all seven models, RubricRefine outperforms CodeAct, with a per-model gap standard deviation of $0.020$--$0.067$ and a minimum observed gap of $+0.083$ (Table~\ref{tab:variance-decomp}). No single run produces a result where the improvement disappears.

\begin{table}[t]
\centering
\footnotesize
\setlength{\tabcolsep}{3pt}
\resizebox{\columnwidth}{!}{%
\begin{tabular}{lcccccc}
\toprule
 & \multicolumn{3}{c}{\textit{RubricRefine}} & \multicolumn{3}{c}{\textit{RR$-$CodeAct gap}} \\
\cmidrule(lr){2-4}\cmidrule(lr){5-7}
Model & Mean & SD & Range & Mean & SD & Min \\
\midrule
\texttt{GPT-4.1-mini} & 0.858 & 0.036 & .792--.917 & +0.221 & 0.067 & +0.146 \\
\texttt{GPT-4o}       & 0.860 & 0.039 & .812--.917 & +0.194 & 0.056 & +0.125 \\
\texttt{o3-mini}      & 0.848 & 0.033 & .792--.917 & +0.190 & 0.044 & +0.125 \\
\texttt{GPT-4.1}      & 0.852 & 0.040 & .792--.938 & +0.198 & 0.064 & +0.104 \\
\texttt{Gemma-4-26B}  & 0.854 & 0.022 & .812--.875 & +0.358 & 0.047 & +0.292 \\
\texttt{Qwen3.6-27B}  & 0.842 & 0.011 & .833--.854 & +0.381 & 0.020 & +0.354 \\
\texttt{Sonnet-4.6}   & 0.875 & 0.035 & .833--.917 & +0.128 & 0.040 & +0.083 \\
\bottomrule
\end{tabular}%
}
\caption{Run-to-run stability analysis for RubricRefine on M3ToolEval across 10 independent trials per model. Left block: trial-level success-rate statistics for RubricRefine. Right block: matched-trial RubricRefine$-$CodeAct gap statistics (RubricRefine outperforms CodeAct on every matched trial for every model).}
\label{tab:variance-decomp}
\end{table}

\begin{table}[t]
\centering
\footnotesize
\resizebox{\columnwidth}{!}{%
\begin{tabular}{lccc}
\toprule
Model & $n$ positive / $n$ total & Wilcoxon $W$ & $p$ (two-sided) \\
\midrule
\texttt{GPT-4.1-mini} & 10 / 10 & $0$ & $\le 0.002$ \\
\texttt{GPT-4o}       & 10 / 10 & $0$ & $\le 0.002$ \\
\texttt{o3-mini}      & 10 / 10 & $0$ & $\le 0.002$ \\
\texttt{GPT-4.1}      & 10 / 10 & $0$ & $\le 0.002$ \\
\texttt{Gemma-4-26B}  & 10 / 10 & $0$ & $0.002$ \\
\texttt{Qwen3.6-27B}  & 10 / 10 & $0$ & $\le 0.002$ \\
\texttt{Sonnet-4.6}   & 10 / 10 & $0$ & $\le 0.002$ \\
\bottomrule
\end{tabular}%
}
\caption{Per-model Wilcoxon signed-rank test for RubricRefine vs.\ CodeAct on M3ToolEval (10 matched trials per model). All 10 trial-level gaps are positive on every model (see Table~\ref{tab:variance-decomp}, ``Min gap'' row), so the Wilcoxon rank-sum statistic $W$ is at its minimum ($0$) and the two-sided $p$-value is at its floor for $n=10$. For \texttt{Gemma-4-26B} we report the explicitly computed value; the remaining rows report the bound implied by all-positive gaps.}
\label{tab:wilcoxon-permodel}
\end{table}

\paragraph{Statistical Methods Summary}
For completeness, the statistical procedures used across the paper are: (i) two-sided paired $t$-tests over the 10 matched trials per model for headline gap claims (Section~\ref{sec:main-results}); (ii) Wilcoxon signed-rank tests on the same 10 matched trial gaps as a non-parametric robustness check (Table~\ref{tab:wilcoxon-permodel}); (iii) standard errors of the mean computed across the 10 trials and reported alongside every mean success rate; and (iv) calibration metrics, namely Expected Calibration Error (ECE) following \citet{10.5555/2888116.2888120} and AUROC, defined formally in Appendix~\ref{app:calibration}. We do not apply explicit multiple-comparison corrections: each per-model paired test is reported on its own and the cross-model claim is supported by the directionally consistent positive gap on every one of the seven models tested, so no model-level conclusion depends on a single significance threshold.

\section{Positioning}
\label{app:positioning}

\subsection{What Is Novel Relative to CodeAct}

CodeAct's core innovation is action representation: use executable code as the action interface \citep{wang2024executable}. RubricRefine targets a different axis, namely the reliability of generated code under strict tool contracts. The two contributions are therefore best understood as complementary rather than competing:
\begin{itemize}
    \item CodeAct: make code mode effective and measurable.
    \item RubricRefine: make code mode more execution-ready before runtime.
\end{itemize}

\subsection{Why This Matters in High-Cost Environments}

Pre-execution error detection is particularly valuable when failed actions carry high marginal cost:
\begin{itemize}
    \item rate-limited or paid APIs,
    \item operations with side effects (state updates, transactions, downstream triggers),
    \item regulated workflows requiring traceable decision logic,
    \item safety-sensitive operations that favor conservative execution policies.
\end{itemize}

These contexts include enterprise automation, healthcare operations, legal and financial workflows, and systems that interact with the physical world. We invoke them to motivate reliability-first design, not to claim direct deployment validation from our benchmark results.

\section{Detailed Benchmark Dossier: M3ToolEval}
\label{app:m3}

\subsection{Task Shape}

M3 tasks provide:
\begin{itemize}
    \item natural-language instruction,
    \item tool registry with signatures and descriptions,
    \item output contract (often deterministic expected value),
    \item action-format contract requiring an \texttt{Action: ... End Action} block.
\end{itemize}

\subsection{Evaluator Semantics in This Repository}

For each instance, generated code is executed and task-level correctness is logged via \texttt{is\_correct}. This is a genuinely end-to-end measure: call structure alone is not enough if the final program output is wrong.

\subsection{Why M3 Is Essential for This Paper}

RubricRefine's main claim is pre-execution reliability for executable code. M3 is therefore the most direct benchmark view for that claim, because success is defined at the level of full-task execution rather than isolated call correctness.

\section{Detailed Benchmark Dossier: API-Bank}
\label{app:apibank}

\subsection{Task Shape}

API-Bank evaluates stepwise tool-use behavior, where each step requires the correct next API call under the current dialogue or task context.

\subsection{Evaluator Semantics in This Repository}

The evaluator parses predicted API calls, validates API name agreement, executes calls against the tool manager, and computes correctness according to API-specific checks. Aggregate score is:
\begin{equation}
\mathrm{API\ Accuracy}=
\frac{\texttt{correct\_api\_calls}}{\texttt{total\_api\_calls}}.
\end{equation}

\paragraph{Package-level evaluation settings.}
The reported API-Bank results use API-call evaluation (\texttt{--api\_test\_enabled}) rather than dialog-response evaluation. The reported score is therefore stateful API-call correctness, with exact API-name matching followed by API-specific result checking. The ROUGE-L scoring function present in the API-Bank evaluator is used only for dialog-response evaluation and is not used for any API-Bank result reported in this paper. M3ToolEval is evaluated by executing the generated code and checking task-level correctness via the benchmark's \texttt{is\_correct} field; no NLTK, SpaCy, ROUGE, or BLEU-style package is used for the reported M3ToolEval metric.

\subsection{Why API-Bank Is Essential for This Paper}

API-Bank serves as the negative-control benchmark for the core hypothesis. RubricRefine targets inter-tool contract violations (output shape, tool routing, argument provenance across multiple calls), so the mechanism predicts a flat result where those failure modes are absent. API-Bank's predominantly single-step API-call evaluation provides exactly that condition: there is little inter-tool structure for the rubric to repair, and the evaluation unit is one call rather than a multi-step program. A null or negative result on API-Bank is therefore a substantive test of the mechanism, not a weakness of the benchmark --- it rules out the alternative that RubricRefine's M3 gains come from extra inference compute generically improving any tool-use output. Appendix~\ref{app:apibank-regression} further analyzes the source of the small negative effect (sample-dependent rubric over-specification on multi-turn dialogue context, which Fixed RubricRefine avoids).

\subsection{API-Bank Results}

Table~\ref{tab:apibank} reports API-Bank success rates across the four OpenAI frontier models. On API-Bank, all methods cluster within a narrow $0.68$--$0.74$ band, and RubricRefine slightly underperforms the CodeAct baseline on every model (mean gap $-0.03$). This is consistent with the mechanism that drives RubricRefine's M3ToolEval gains: inter-tool contract checks have little room to help on predominantly single-step tasks. A qualitative error analysis (Appendix~\ref{app:apibank-regression}) reveals that the sample-dependent rubric actively hurts in this setting: $69\%$ of regressions involve the rubric redirecting the generator to the wrong API, because the rubric over-specifies which API to call from the full dialogue context without knowledge of the current conversation step. Fixed RubricRefine, which uses a generic rubric without task-specific API enumeration, avoids most of these regressions and matches the baseline.

\begin{table}[t]
\centering
\resizebox{\columnwidth}{!}{%
\begin{tabular}{lcccc}
\toprule
Method & \texttt{GPT-4.1-mini} & \texttt{GPT-4o} & \texttt{o3-mini} & \texttt{GPT-4.1} \\
\midrule
CodeAct (Baseline) & $.72 \pm .01$ & $.72 \pm .00$ & $.72 \pm .00$ & $.72 \pm .01$ \\
Self-Refine & $.68 \pm .01$ & $.71 \pm .01$ & $.73 \pm .01$ & $.71 \pm .01$ \\
Best-of-$N$ & $.73 \pm .00$ & $.73 \pm .00$ & $.73 \pm .00$ & $.73 \pm .00$ \\
\midrule
BoN+fixed rubric (Ctrl) & $.72 \pm .00$ & $.72 \pm .01$ & $.73 \pm .01$ & $.72 \pm .01$ \\
BoN+rubric (Ctrl) & $.73 \pm .00$ & $.73 \pm .00$ & $.74 \pm .00$ & $.74 \pm .00$ \\
Fixed RubricRefine (Ctrl) & $.72 \pm .01$ & $.72 \pm .00$ & $.72 \pm .01$ & $.71 \pm .01$ \\
RubricRefine (Ours) & $.69 \pm .01$ & $.70 \pm .01$ & $.68 \pm .01$ & $.69 \pm .01$ \\
\bottomrule
\end{tabular}%
}
\caption{API-Bank success rates (mean $\pm$ SE across trials) for four models. Tasks are predominantly single-step API calls. Logprob-weighted scoring variants are reported in Table~\ref{tab:logprob-apibank} (Appendix~\ref{app:logprob-results}).}
\label{tab:apibank}
\end{table}

\subsection{API-Bank Regression Analysis}
\label{app:apibank-regression}

Section~\ref{sec:apibank-results} reports that RubricRefine scores below the CodeAct baseline on all four models. This subsection investigates why the full iterative loop with sample-dependent rubrics actively hurts on API-Bank rather than merely showing no improvement.

\paragraph{Regression Counts}
Across all four OpenAI frontier models and 10 trials, we identified 62 total regression instances (cases where CodeAct produced the correct API call but RubricRefine did not). This averages to $3.0$--$3.8$ regressions per trial per model, against $2.0$--$2.2$ improvements, yielding a net negative of roughly $1$--$2$ calls per trial out of $\approx 50$ predictions, consistent with the $\approx 3$-point accuracy drop observed in Table~\ref{tab:apibank}.

\paragraph{Root Cause: Rubric Over-Specification in Multi-Turn Dialogues}
Of the 62 regression instances, $69\%$ involve the rubric redirecting the generator to the \emph{wrong API entirely}, while $31\%$ involve parameter changes to the correct API. The dominant failure pattern is rubric over-specification in multi-turn dialogues. API-Bank instances are drawn from multi-turn conversations where several APIs are available (e.g., \texttt{ModifyRegistration}, \texttt{QueryHealthData}, \texttt{CancelRegistration}), and each evaluation step asks for the correct \emph{next} API call given the current dialogue context. The sample-dependent rubric, conditioned on the full task description and tool registry, enumerates all APIs mentioned in the dialogue context and can misidentify which API is appropriate at the current step.

\paragraph{Illustrative Examples}
Two recurring patterns account for the majority of regressions:
\begin{itemize}
    \item \textbf{CancelRegistration $\to$ ModifyRegistration} (17 of 62 instances): The ground truth requires \texttt{CancelRegistration}, and CodeAct correctly calls it. The sample-dependent rubric, seeing the full dialogue context that includes prior modification steps, scores the correct cancellation call as a failure and revises toward \texttt{ModifyRegistration}.
    \item \textbf{ForgotPassword $\to$ GetUserToken} (12 of 62 instances): The ground truth requires \texttt{ForgotPassword}, but the rubric, conditioned on a dialogue involving multiple account-management APIs, directs the generator toward \texttt{GetUserToken} as the expected first step.
\end{itemize}

\paragraph{Fixed RubricRefine Avoids Most Regressions}
Fixed RubricRefine, which uses a generic rubric without task-specific API enumeration, avoids $50$--$91\%$ of the regressions that affect the full RubricRefine method (depending on model), and scores at or above the CodeAct baseline on API-Bank (Table~\ref{tab:apibank}). This confirms that the regression is caused by sample-dependent rubric over-specification rather than by the iterative repair loop itself.

\paragraph{Interpretation}
The failure mode is specific to settings where (1)~the evaluation unit is a single step within a multi-turn dialogue, and (2)~the rubric generator does not have access to the current dialogue state (only the task description and tool registry). In such settings, the sample-dependent rubric cannot reliably determine which API is appropriate at the current step, and its task-specific criteria become counterproductive. This contrasts with M3ToolEval, where the full task is evaluated as a single executable program and the rubric's task-specific criteria correctly target inter-tool contract violations.

\section{Why Success Criteria Differ Across the Two Benchmarks}
\label{app:criteria}

\begin{center}
\footnotesize
\setlength{\tabcolsep}{3pt}
\begin{tabular}{p{0.20\linewidth}p{0.35\linewidth}p{0.35\linewidth}}
\toprule
Dimension & M3ToolEval & API-Bank \\
\midrule
Unit of scoring & Whole task instance & Next API step \\
Primary question & ``Did the program solve the task?'' & ``Was the API call correct?'' \\
Error visibility & Inter-tool dataflow + execution effects & Call-level schema \& sequencing \\
Inter-tool contract structure & Rich (multi-step composition) & Sparse (predominantly single-step) \\
Role in this paper & Where the mechanism should help & Negative control: where it should not \\
\bottomrule
\end{tabular}
\end{center}

The criteria are intentionally different, and the asymmetry is the point. RubricRefine targets inter-tool contract violations, so the mechanism predicts a sizable lift on M3ToolEval and a flat (or worse) result on API-Bank, where contract structure is sparse. Reporting both therefore tests \emph{whether} the mechanism operates on the dimension we claim, rather than averaging two views of ``reliability.'' Improvement on both would not constitute a stronger reliability claim under our hypothesis; it would in fact weaken the mechanism story by suggesting the gains come from generic inference-compute scaling rather than from contract-structured checking. The observed pattern (large lift on M3, null on API-Bank) is exactly what the mechanism predicts.

\section{Execution Accounting in the Single-Attempt Regime}
\label{app:turns}

\paragraph{Definition Used in This Paper}
One task instance permits at most one live execution attempt.

\paragraph{Implication}
If a method performs five refinement rounds and then executes once, that sequence still counts as a single execution attempt.

\paragraph{Why This Definition Is Important}
This accounting separates \emph{reasoning and editing compute} from \emph{environment interaction cost}, which is the more meaningful unit in stateful or expensive environments where retrying after failure is operationally unattractive.

\section{Rubric-Category Ablation Details}
\label{app:ablation-detail}

\begin{table}[t]
\centering
\footnotesize
\setlength{\tabcolsep}{3pt}
\resizebox{\columnwidth}{!}{%
\begin{tabular}{lcccccc}
\toprule
 & \multicolumn{3}{c}{\texttt{GPT-4o}} & \multicolumn{3}{c}{\texttt{GPT-4.1-mini}} \\
\cmidrule(lr){2-4}\cmidrule(lr){5-7}
Condition removed & Mean & $\Delta$ & 95\% CI & Mean & $\Delta$ & 95\% CI \\
\midrule
\textit{Full (none)}  & $.825$ & --- & --- & $.883$ & --- & --- \\
\midrule
Call-signature  & $.767$ & $-.058$ & $[-.081,-.035]$ & $.738$ & $\mathbf{-.146}$ & $[-.175,-.117]$ \\
Tool-choice     & $.813$ & $-.012$ & $[-.038,+.013]$ & $.725$ & $\mathbf{-.158}$ & $[-.182,-.135]$ \\
Output-contract & $.810$ & $-.015$ & $[-.047,+.018]$ & $.860$ & $-.023$ & $[-.056,+.010]$ \\
Data-provenance & $.840$ & $+.015$ & $[-.011,+.040]$ & $.854$ & $-.029$ & $[-.057,-.002]$ \\
\bottomrule
\end{tabular}%
}
\caption{Rubric-category ablation on M3ToolEval ($10$ matched trials per condition per model, $R=5$). $\Delta$ is the mean paired per-trial difference from the full condition, with $95\%$ confidence intervals on that paired difference over the 10 matched trials; an interval excluding zero indicates a reliable effect. \textbf{Every effect is larger on the weaker model}: removals cost \texttt{GPT-4.1-mini} up to $-0.158$ but \texttt{GPT-4o} at most $-0.058$, so explicit rubric scaffolding matters most where the generator is least able to infer contracts unaided.}
\label{tab:ablation-multi}
\end{table}

Table~\ref{tab:ablation-multi}, discussed in Section~\ref{sec:ablation}, reports the ablation on \texttt{GPT-4o} and \texttt{GPT-4.1-mini}, each with $10$ matched-seed trials per condition at $R=5$. Ablation is implemented by removing rule lines matching a category's pattern set from all three system prompts (rubric design, scoring, and repair), so a removed category is absent from rubric construction, from scoring, and from the repair directive simultaneously.

\paragraph{Interpretation and limits}
Only \emph{call-signature} rules are load-bearing on both models. \emph{Tool-choice} rules matter on \texttt{GPT-4.1-mini} but not \texttt{GPT-4o}; \emph{output-contract} and \emph{data-provenance} removals fall within noise on both. We read the capability gradient as running toward the weaker model: every ablation effect is larger on \texttt{GPT-4.1-mini}, which is consistent with stronger models satisfying contracts they can already infer from documentation without explicit enumeration. Two limits bound these conclusions. First, an ablation measures marginal contribution \emph{given the rest of the system}, so a null result indicates redundancy rather than irrelevance; the output-contract case is the clearest instance, since the repair prompt separately constrains the emitted final value (Appendix~\ref{app:prompts}). Second, two models do not establish a general capability gradient, and both are from a single provider family.

\section{Documentation-Degradation Protocol}
\label{app:doc-robustness}

This appendix documents the perturbation procedure behind Table~\ref{tab:doc-robustness}. All runs use \texttt{GPT-4o} on the full M3ToolEval task set under the same single-attempt protocol as the main results, with $10$ independent trials per condition and a same-model verifier. Only the tool documentation presented to the agent is perturbed; the underlying tool implementations, the ground-truth answers, and the evaluator are untouched, so a task remains solvable in every condition.

\paragraph{Truncation}
For each tool, we drop the trailing fraction $\sigma \in \{0.25, 0.50, 0.75\}$ of its natural-language description, keeping the function signature intact. This models documentation that is incomplete but not misleading: the agent still sees correct parameter names and arity, but loses semantic detail about behavior, units, and edge cases. Both CodeAct and RubricRefine see the identical degraded registry, and the rubric generator conditions on the same truncated docs, so no method receives privileged information.

\paragraph{Parameter scrambling}
We replace every parameter name in every documented signature with a positional generic (\texttt{arg0}, \texttt{arg1}, \ldots), leaving descriptions and arity intact. This models documentation that is confidently \emph{wrong}: the names are well-formed and plausible, but do not match what the tool actually accepts. This is the adversarial case for any document-grounded checker, because nothing in the registry signals that the contract is untrustworthy.

\paragraph{Interpreting the asymmetry}
Truncation degrades both methods' absolute success sharply (CodeAct falls from $0.713$ to $0.352$--$0.377$) while leaving the RubricRefine$-$CodeAct gap intact, which is what we would expect if the rubric's leverage comes primarily from the structural contract encoded in signatures rather than from prose detail. Scrambling costs CodeAct comparatively little ($0.713 \to 0.600$) but costs RubricRefine $0.410$ absolute ($0.831 \to 0.421$). The asymmetry in \emph{which} method is harmed more is the informative part: CodeAct absorbs the perturbation comparatively well because it can often infer correct usage despite unhelpful names, whereas RubricRefine promotes the erroneous names into binding PASS/FAIL criteria and then repairs \emph{toward} them. Checking machinery that faithfully enforces its input contract will faithfully enforce a wrong one, which is why we treat this as a bound on the method's trustworthy operating regime rather than a tuning problem.

\section{Cross-Provider Verifier Details}
\label{app:cross-verifier}

Section~\ref{sec:cross-verifier} reports one cross-provider pairing: a \texttt{GPT-4.1} generator with a \texttt{Claude-Sonnet-4.6} verifier, run for $3$ trials on M3ToolEval with rubric generation, scoring, and repair prompts, decoding settings, budgets ($R=5$, $P=2$), and evaluation semantics identical to the main-text protocol. The generator is accessed through the OpenAI-compatible endpoint used elsewhere in the paper and the verifier through the Anthropic API; only the verifier's backing model differs from the corresponding row of Table~\ref{tab:main-m3-success}.

\paragraph{Result and scope}
Per-trial success rates are $0.896$, $0.854$, and $0.833$ (mean $0.861$, SD $0.032$, SE $0.018$), against $0.852$ (SD $0.040$, SE $0.013$, $n=10$) for the same-model verifier. The two are statistically indistinguishable, so a shared generator--verifier backbone is not necessary for the gain. We deliberately do not draw a stronger conclusion. With one pairing at $3$ trials we cannot speak to whether a \emph{weaker} verifier would suffice (the deployment-relevant cheap-verifier question), whether the direction of the pairing matters, or whether correlated errors would appear in pairings we did not run. Establishing that would require a generator $\times$ verifier matrix at trial counts comparable to Table~\ref{tab:main-m3-success}, which we flag as future work rather than claim as evidence here.

\section{Static Checker Baselines}
\label{app:static-checker}

The static checker is a deterministic, non-LLM auditor that scores candidate code by parsing it with \texttt{ast.parse} and walking the AST to check tool-call sites against documented signatures extracted from the prompt. Specifically, it checks: (1)~response-format conformance (\texttt{Action:/End Action} structure, bracket format); (2)~Python syntax validity of the action block; (3)~call-shape conformance for each tool call (positional vs.\ keyword usage matching the documented signature, no undocumented keyword arguments, no starred or dict expansion in non-variadic calls, no missing required parameters); and (4)~ambiguous binding patterns (keyword followed by positional). Each violation incurs a penalty that is subtracted from the maximum score. Crucially, the checker makes no judgment about semantic correctness: it does not verify whether the correct tools are called, whether argument \emph{values} are correct, whether call ordering or dataflow is sound, or whether the final answer is grounded in tool outputs.

Table~\ref{tab:static-checker} reports M3ToolEval success rates for two baselines using this checker across all four OpenAI frontier models: (1)~\textbf{Static Refine}, which uses the static score and its issue descriptions as the revision signal in the same iterative repair loop as RubricRefine; and (2)~\textbf{BoN+Static}, which uses static scores to select among five candidates.

\begin{table}[t]
\centering
\resizebox{\columnwidth}{!}{%
\begin{tabular}{lcccc}
\toprule
Method & \texttt{GPT-4.1-mini} & \texttt{GPT-4o} & \texttt{o3-mini} & \texttt{GPT-4.1} \\
\midrule
CodeAct (Baseline) & $.64$ & $.67$ & $.66$ & $.65$ \\
\midrule
Static Refine & $.71 \pm .01$ & $.72 \pm .01$ & $.73 \pm .01$ & $.70 \pm .02$ \\
BoN+Static & $.54 \pm .01$ & $.56 \pm .01$ & $.57 \pm .01$ & $.55 \pm .01$ \\
\midrule
RubricRefine (Ours) & $.86 \pm .01$ & $.86 \pm .01$ & $.85 \pm .01$ & $.85 \pm .01$ \\
\bottomrule
\end{tabular}%
}
\caption{Static checker baselines on M3ToolEval. Static Refine uses AST parsing and signature matching without LLM semantic reasoning. BoN+Static uses static scores for candidate selection. RubricRefine included for reference.}
\label{tab:static-checker}
\end{table}

Static Refine improves over CodeAct by $+0.06$--$0.08$ across models, capturing a fraction of RubricRefine's $+0.18$--$0.22$ gain. BoN+Static is worse still, scoring $0.54$--$0.57$ against the $0.64$--$0.67$ CodeAct baseline and the $0.75$--$0.77$ reached by rubric-guided selection.

\paragraph{Why this comparison is informative}
The selection setting isolates the variable of interest. Both BoN+Static and BoN+rubric draw the same $N=5$ candidates and apply the same rule --- take the highest-scoring one --- so they differ only in the scorer. Substituting a deterministic checker for the LLM verifier therefore measures the contribution of the checking mechanism itself, with the candidate pool and selection policy held fixed.

\paragraph{The result is that call-shape checking succeeds when done by an LLM and fails when done syntactically}
This is a striking result, because the comparison is stacked in the static checker's favour. Call-shape conformance is mechanically decidable from an AST: whether an argument name appears in a documented signature, whether a required parameter is missing, whether a call passes positionally where keywords are required. These are the checks our static auditor performs, and the rubric-category ablation (Section~\ref{sec:ablation}) identifies call-signature rules as the one category load-bearing on both models tested. The dimension that matters most is thus the dimension a deterministic tool should verify most reliably --- and yet enforcing it syntactically recovers only a third of the gain under iterative repair, and \emph{inverts} it under selection, landing below the unrefined baseline.

We read this as evidence that the mechanism of checking, not the choice of dimension, is what carries the improvement. A syntactic checker can confirm that a call is well-formed against a signature; it cannot judge whether the argument \emph{values} are the ones the task requires, whether the call ordering respects cross-tool dataflow, or whether a consumed value traces to a legitimate source. It also cannot express its verdict as a directive: BoN+Static's below-baseline score indicates that a penalty-based static score actively misranks candidates, whereas the rubric's itemized PASS/FAIL output doubles as a repair instruction. Two caveats bound the claim. It concerns the checker we implement, and a more capable static analyzer with type inference or dataflow tracking might narrow the gap. And the two methods' check sets coincide only on the signature dimension, so this is not a general statement about static analysis --- it is a statement about the one dimension where the two approaches are directly comparable. BoN+Static scores below CodeAct on every model ($-0.08$--$0.12$), consistent with static scores being a noisy selection signal that actively misleads the ranker in the multi-step setting.

\section{Per-Task Per-Model Breakdown}
\label{app:frontier-per-task}

Table~\ref{tab:frontier-per-task} reports the per-task M3ToolEval success rates for the narrative-critical comparison between baseline CodeAct and RubricRefine across all seven evaluated models. This view makes the heterogeneity behind the aggregate results explicit. Travel itinerary planning shows the largest and most consistent gains, improving by $+0.41$ to $+0.48$ on the four frontier OpenAI models, by $+0.81$ on \texttt{Gemma-4-26B}, by $+0.77$ on \texttt{Qwen3.6-27B}, and by $+0.06$ on \texttt{Sonnet-4.6} (where the CodeAct baseline on travel is already $0.73$). On the frontier models, message decoder and DNA sequencer show moderate positive gains ($+0.14$ to $+0.26$) and trade calculator is flat or within noise (range $-0.01$ to $+0.02$). On the two open-weight models, the per-task pattern is similar in shape but larger in magnitude on every family (\texttt{Gemma-4-26B}: message $+0.24$, DNA $+0.17$, trade $+0.12$; \texttt{Qwen3.6-27B}: message $+0.48$, DNA $+0.08$, trade $+0.11$), reflecting the weaker CodeAct baselines on these models rather than a qualitatively different mechanism. On \texttt{Sonnet-4.6}, lifts are smaller on the easier families (message $-0.02$, DNA $+0.10$) but the largest single-task-family gain comes on trade calculator ($+0.27$), the only family where the CodeAct baseline is weakest for this model. This pattern is consistent with the main-text interpretation that RubricRefine helps most where multi-step inter-tool contracts are richest, while leaving cases with already-strong baselines relatively unchanged.

\begin{table}[t]
\centering
\footnotesize
\setlength{\tabcolsep}{3pt}
\resizebox{\columnwidth}{!}{%
\begin{tabular}{llcccc}
\toprule
Model & Method & \shortstack{Travel\\($n{=}15$)} & \shortstack{Message\\($n{=}8$)} & \shortstack{DNA\\($n{=}8$)} & \shortstack{Trade\\($n{=}17$)} \\
\midrule
\texttt{GPT-4.1-mini} & CodeAct & 0.45 & 0.74 & 0.61 & 0.77 \\
\texttt{GPT-4.1-mini} & RubricRefine & 0.93 & 0.93 & 0.86 & 0.76 \\
\texttt{GPT-4.1-mini} & $\Delta$ & +0.48 & +0.19 & +0.25 & $-0.01$ \\
\midrule
\texttt{GPT-4o} & CodeAct & 0.43 & 0.78 & 0.80 & 0.76 \\
\texttt{GPT-4o} & RubricRefine & 0.91 & 0.94 & 0.88 & 0.78 \\
\texttt{GPT-4o} & $\Delta$ & +0.47 & +0.16 & +0.08 & +0.02 \\
\midrule
\texttt{o3-mini} & CodeAct & 0.46 & 0.75 & 0.75 & 0.75 \\
\texttt{o3-mini} & RubricRefine & 0.87 & 0.89 & 0.94 & 0.77 \\
\texttt{o3-mini} & $\Delta$ & +0.41 & +0.14 & +0.19 & +0.02 \\
\midrule
\texttt{GPT-4.1} & CodeAct & 0.47 & 0.75 & 0.69 & 0.75 \\
\texttt{GPT-4.1} & RubricRefine & 0.89 & 0.89 & 0.95 & 0.75 \\
\texttt{GPT-4.1} & $\Delta$ & +0.42 & +0.14 & +0.26 & +0.00 \\
\midrule
\texttt{Gemma-4-26B} & CodeAct & 0.03 & 0.54 & 0.75 & 0.77 \\
\texttt{Gemma-4-26B} & RubricRefine & 0.83 & 0.78 & 0.90 & 0.89 \\
\texttt{Gemma-4-26B} & $\Delta$ & +0.81 & +0.24 & +0.15 & +0.12 \\
\midrule
\texttt{Qwen3.6-27B} & CodeAct & 0.03 & 0.52 & 0.70 & 0.71 \\
\texttt{Qwen3.6-27B} & RubricRefine & 0.80 & 1.00 & 0.78 & 0.82 \\
\texttt{Qwen3.6-27B} & $\Delta$ & +0.77 & +0.48 & +0.08 & +0.11 \\
\midrule
\texttt{Sonnet-4.6} & CodeAct & 0.73 & 0.90 & 0.75 & 0.69 \\
\texttt{Sonnet-4.6} & RubricRefine & 0.79 & 0.88 & 0.85 & 0.96 \\
\texttt{Sonnet-4.6} & $\Delta$ & +0.06 & $-0.02$ & +0.10 & +0.27 \\
\bottomrule
\end{tabular}%
}
\caption{Per-task M3ToolEval success rates for baseline CodeAct versus RubricRefine across all seven evaluated models. $\Delta$ denotes RubricRefine minus CodeAct. \texttt{Gemma-4-26B} and \texttt{Qwen3.6-27B} show the largest absolute gains on travel planning ($+0.81$ and $+0.77$, vs.\ $+0.41$--$+0.48$ on the frontier models), reflecting much weaker CodeAct baselines on this family and RubricRefine's effectiveness at repairing inter-tool contract violations.}
\label{tab:frontier-per-task}
\end{table}

\section{Additional Scaling Analyses}
\label{app:scaling}

\begin{table}[t]
\centering
\resizebox{\columnwidth}{!}{%
\begin{tabular}{lccc}
\toprule
Method & \shortstack{Succ.\\Rate} & \shortstack{Wall-Clk\\(s/task)} & \shortstack{Tokens\\(/task)} \\
\midrule
CodeAct (Baseline) & 0.65 & 1.7 & 814 \\
Self-Refine & 0.60 & 29.6 & 18{,}375 \\
Self-Debug & 0.75 & 3.1 & 2{,}483 \\
Best-of-$N$ & 0.62 & 10.1 & 5{,}987 \\
\midrule
BoN+fixed rubric (Ctrl) & 0.73 & 55.7 & 42{,}282 \\
BoN+rubric (Ctrl) & 0.76 & 76.6 & 53{,}476 \\
Fixed RubricRefine (Ctrl) & 0.80 & 14.1 & 12{,}748 \\
RubricRefine (Ours) & 0.85 & 30.0 & 27{,}877 \\
\bottomrule
\end{tabular}%
}
\caption{Efficiency on M3ToolEval (\texttt{GPT-4.1}). RubricRefine outperforms all methods while requiring $2.6\times$ lower latency and $48\%$ fewer tokens than BoN+rubric. At April 2026 \texttt{GPT-4.1} API pricing (\$2/M input, \$8/M output tokens, assuming a roughly 1:1 input/output split), RubricRefine costs approximately \$0.14/task vs.\ \$0.27/task for BoN+rubric and \$0.004/task for single-pass CodeAct.}
\label{tab:efficiency}
\end{table}

\subsection{Success vs.\ Wall-Clock Latency}

\begin{figure}[t]
\centering
\includegraphics[width=0.75\linewidth]{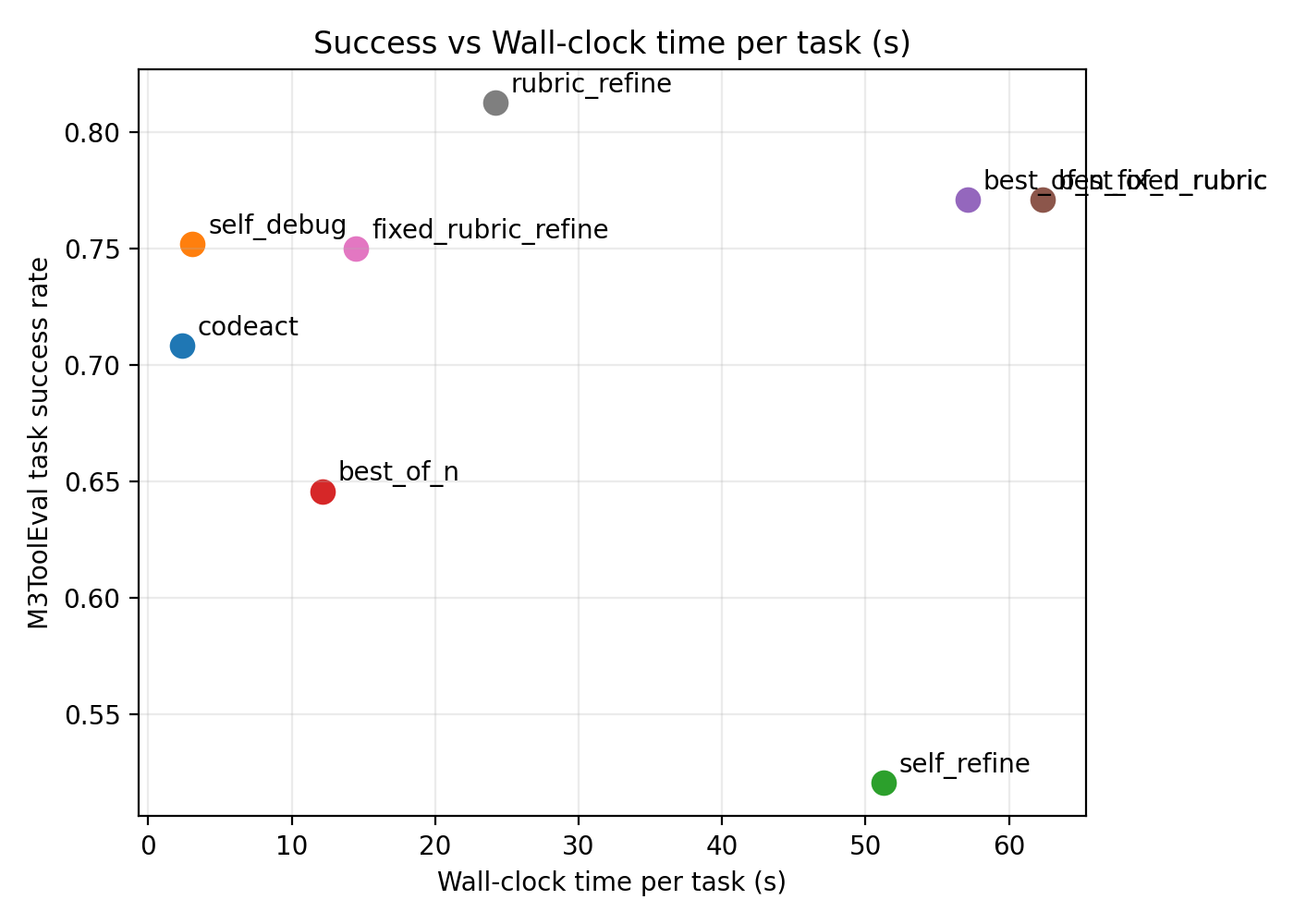}
\caption{Success rate vs.\ wall-clock latency on M3ToolEval (\texttt{GPT-4.1}; same \texttt{method\_eval\_fixed\_story} run as the main tables).}
\label{fig:appendix-wall-time}
\end{figure}

Figure~\ref{fig:appendix-wall-time} provides the appendix wall-clock comparison with the full method set and any additional budget sweeps. We use this plot to check that the main-paper comparison between RubricRefine and Self-Refine is not driven by a single reporting point. If RubricRefine traces a stronger success--latency frontier across multiple operating points, then the improvement reflects a better use of inference time rather than merely more inference time.

\subsection{Success vs.\ LM Calls}

\begin{figure}[t]
\centering
\includegraphics[width=0.75\linewidth]{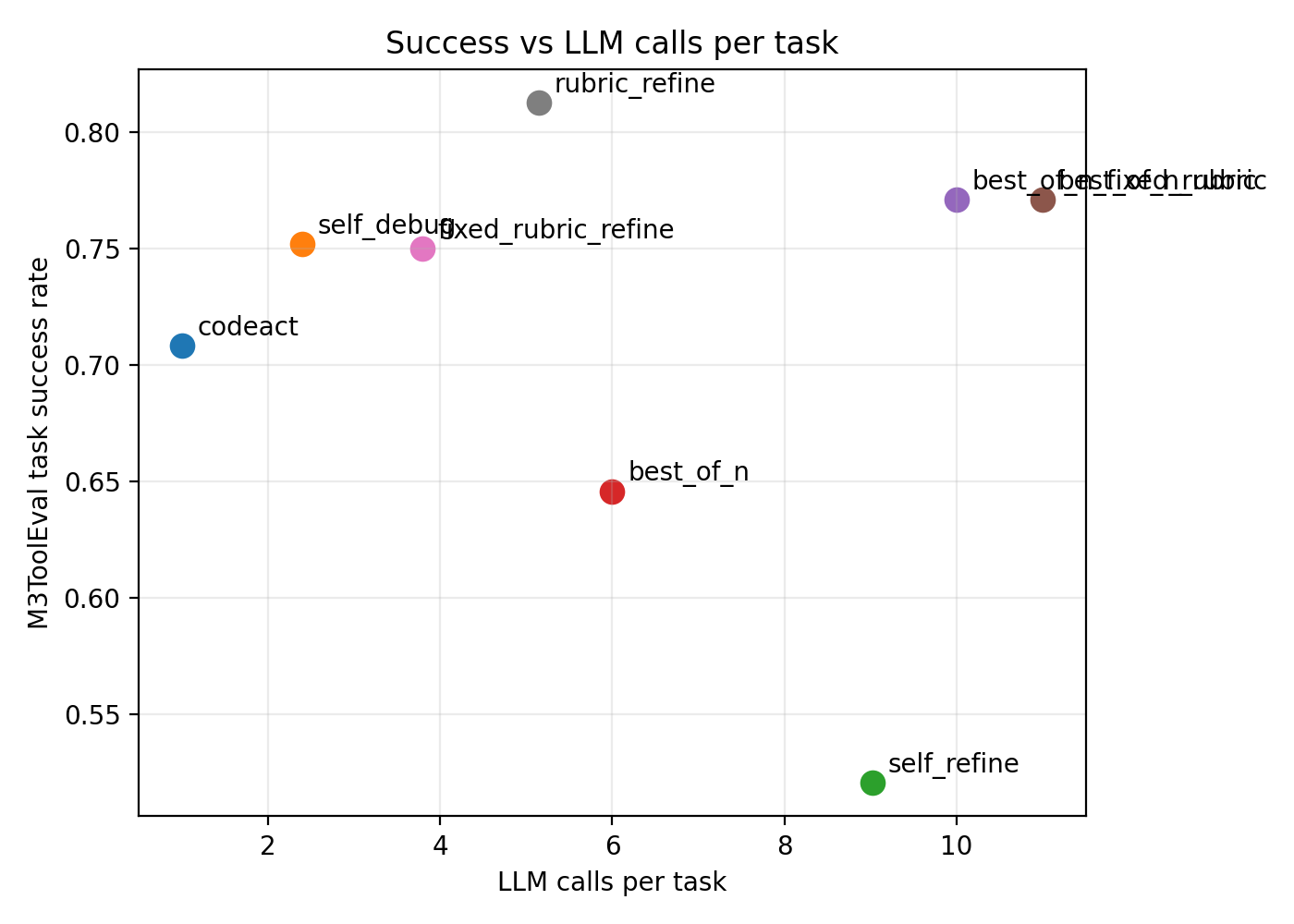}
\caption{Success rate vs.\ total LM calls on M3ToolEval (\texttt{GPT-4.1}; same \texttt{method\_eval\_fixed\_story} run as the main tables).}
\label{fig:appendix-llm-calls}
\end{figure}

Figure~\ref{fig:appendix-llm-calls} reports success against total LM calls. This controls for a different notion of budget than wall-clock latency: call count abstracts away from serving noise and asks how effectively each method turns model invocations into successful trajectories.

\subsection{Success vs.\ Total Tokens}

\begin{figure}[t]
\centering
\includegraphics[width=0.75\linewidth]{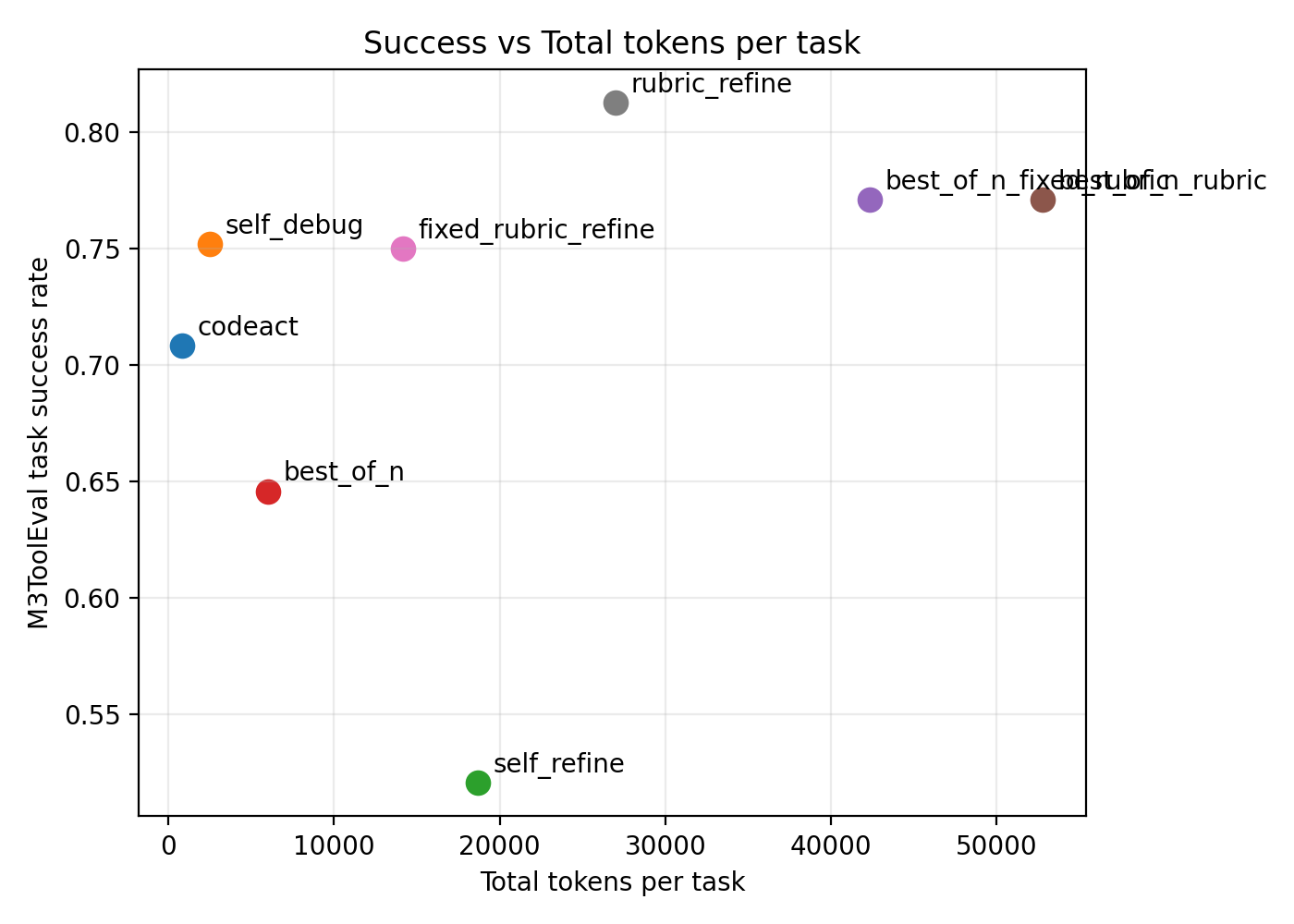}
\caption{Success rate vs.\ total tokens on M3ToolEval (\texttt{GPT-4.1}; same \texttt{method\_eval\_fixed\_story} run as the main tables).}
\label{fig:appendix-tokens}
\end{figure}

Figure~\ref{fig:appendix-tokens} reports success against total token usage. This is the cleanest accounting view for prompt-length and verifier-overhead concerns because it directly measures how much language-model computation each method consumes.

\subsection{Other Inference-Cost Views}

The main paper reports success versus wall-clock latency because that is the cleanest deployment-facing measure. We additionally report success against total LM calls and total tokens in the appendix using the same \texttt{method\_eval\_fixed\_story} run as the main tables. These plots help distinguish whether a method's gains arise from a better use of comparable inference budget or simply from much larger token expenditure.

\subsection{Inference Tradeoffs on \texttt{Gemma-4-26B}}
\label{app:gemma4-tradeoffs}

\begin{figure}[t]
\centering
\includegraphics[width=0.75\linewidth]{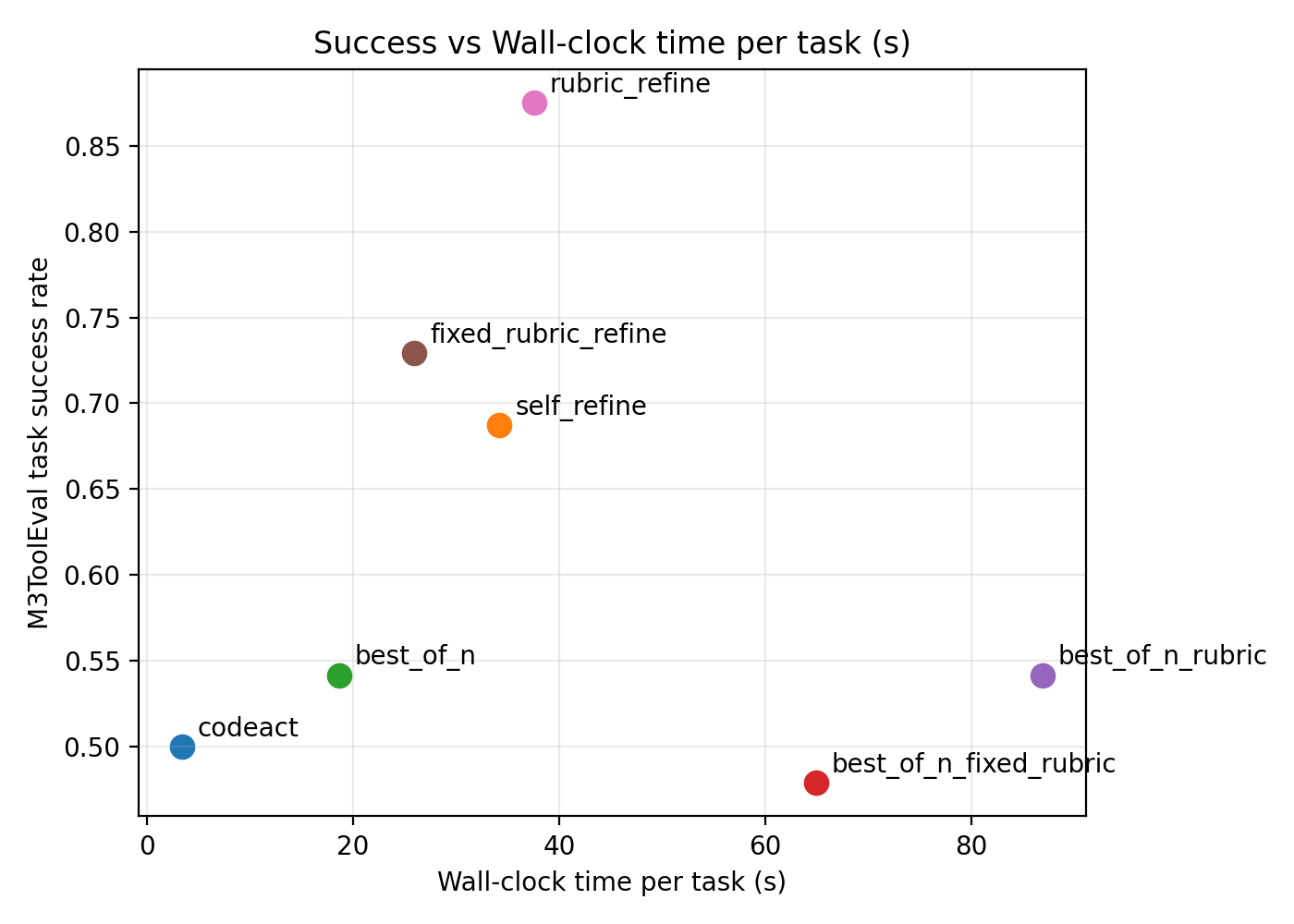}\\[4pt]
\includegraphics[width=0.75\linewidth]{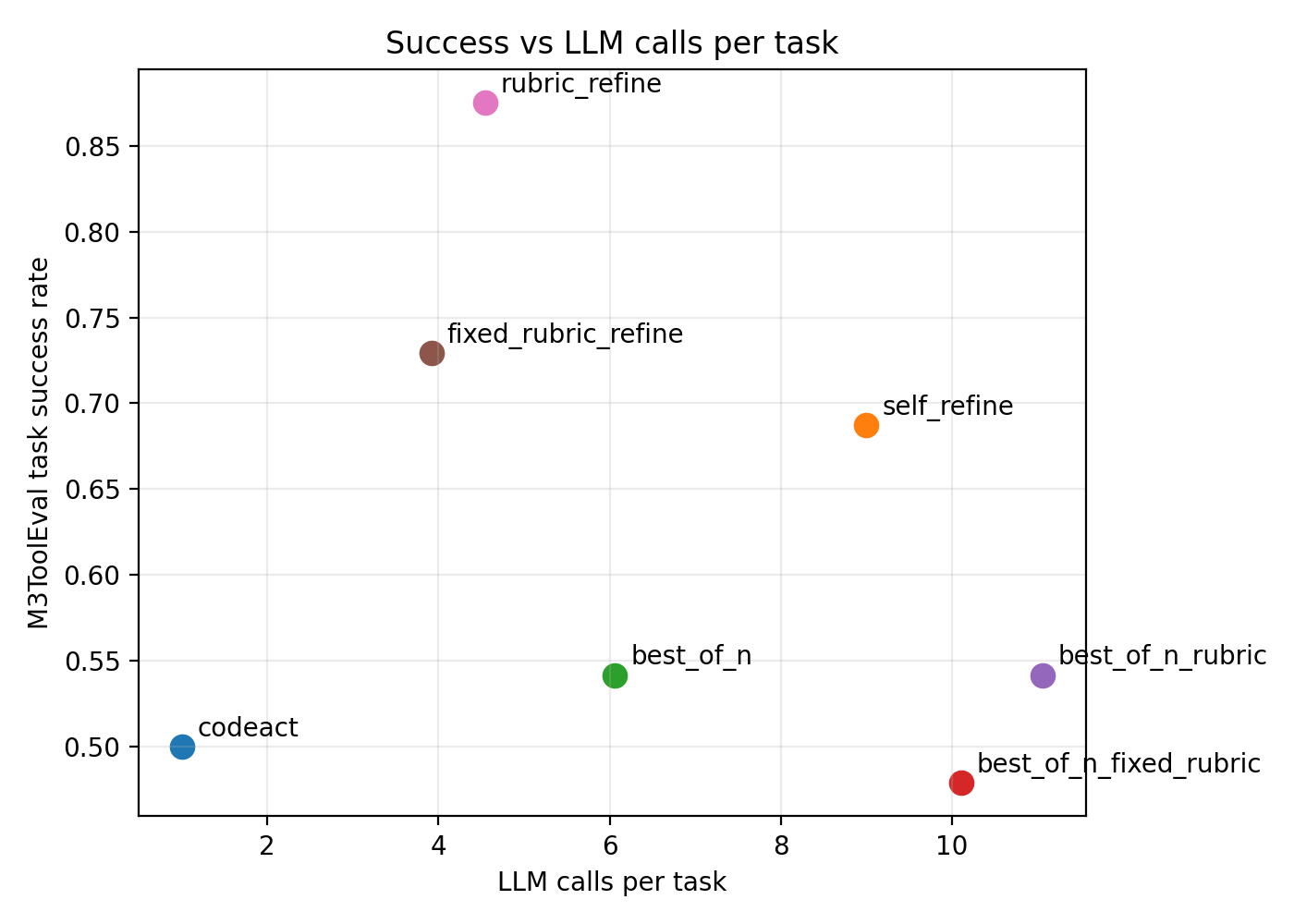}\\[4pt]
\includegraphics[width=0.75\linewidth]{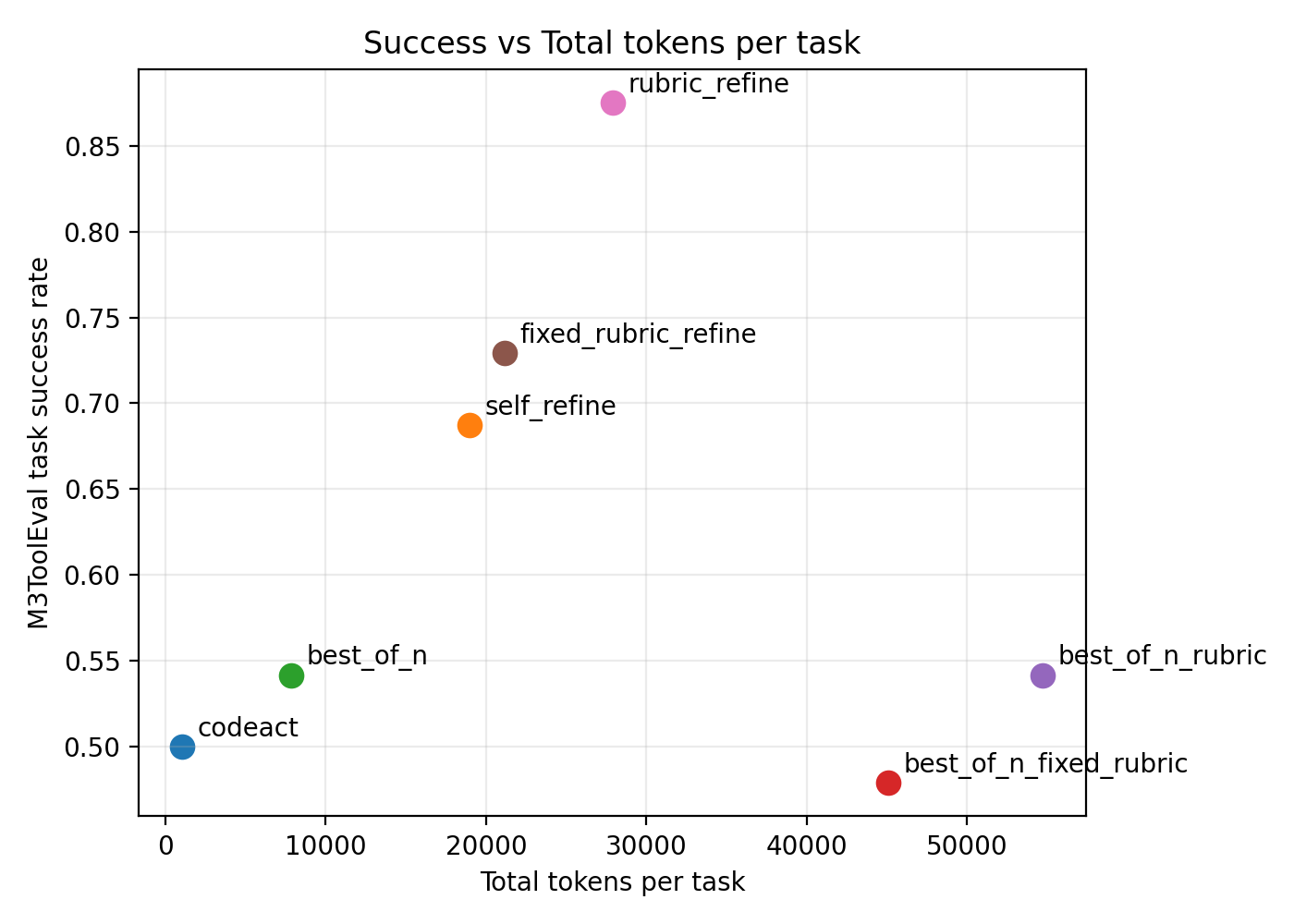}
\caption{Inference-cost tradeoffs on M3ToolEval for \texttt{Gemma-4-26B} (served locally via vLLM). Top: success vs.\ wall-clock latency per task. Middle: success vs.\ LM calls per task. Bottom: success vs.\ total tokens per task. RubricRefine achieves the highest success rate while consuming strictly less of each inference-cost axis than Best-of-$N$+rubric, matching the qualitative pattern seen on the frontier API models.}
\label{fig:gemma4-tradeoff}
\end{figure}

The frontier-model tradeoff plots in the previous subsections use API latency, which mixes model compute with network overhead. To check that RubricRefine's efficiency advantage transfers to a different serving regime, we also report the same three views for \texttt{Gemma-4-26B} served locally via vLLM (Figure~\ref{fig:gemma4-tradeoff}). The qualitative pattern matches the frontier-model plots exactly: RubricRefine achieves the highest success rate while using substantially fewer wall-clock seconds, LM calls, and tokens than Best-of-$N$+rubric. Quantitatively, RubricRefine averages ${\approx}38$s and ${\approx}27.9$k tokens per task against ${\approx}87$s and ${\approx}54.7$k for Best-of-$N$+rubric, a $2.3\times$ wall-clock reduction and ${\approx}49\%$ fewer tokens --- closely tracking the $2.6\times$ and $48\%$ measured on \texttt{GPT-4.1} through a provider API (Table~\ref{tab:efficiency}) despite the entirely different serving path. Combined with the round-stopping statistics in Table~\ref{tab:round-stopping}, where \texttt{Gemma-4-26B} terminates at round 2 in $65.5\%$ of tasks and never exceeds round 3, this confirms that the early-stopping mechanism that drives RubricRefine's efficiency is not specific to the OpenAI family.

\subsection{Budget-Parameter Scaling}
\label{app:budget-scaling}

The main experiments fix the two primary budget parameters (maximum refinement rounds $R=5$ for RubricRefine and number of candidates $N=5$ for Best-of-$N$+rubric) at a single operating point. To assess sensitivity to these choices, we reconstruct per-round and per-candidate trajectories from saved candidate-level data across the four OpenAI frontier models.

\begin{figure}[t]
\centering
\includegraphics[width=\linewidth]{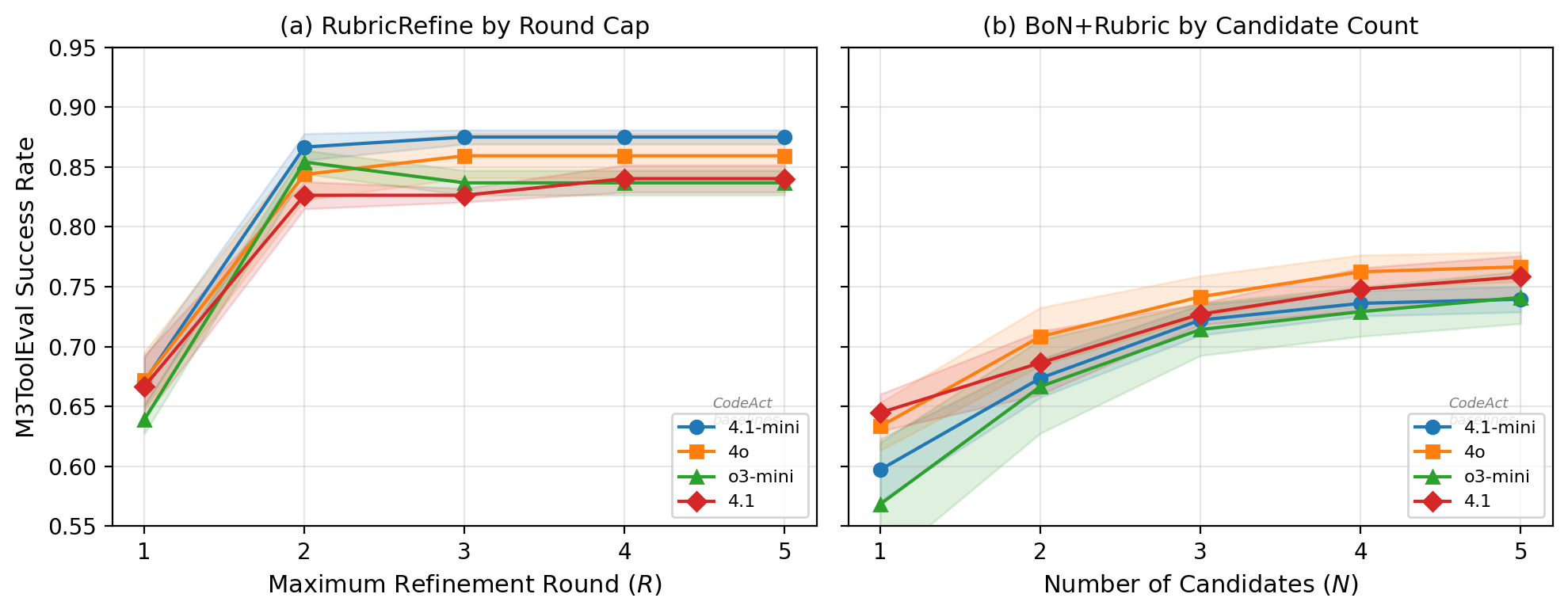}
\caption{Budget-parameter scaling on M3ToolEval. (a)~RubricRefine success rate by maximum refinement round~$R$. (b)~Best-of-$N$+rubric success rate by number of candidates~$N$. Shaded regions show $\pm$1 SE across available trials. Dotted lines indicate per-model CodeAct baselines.}
\label{fig:budget-scaling}
\end{figure}

\begin{figure}[t]
\centering
\includegraphics[width=\linewidth]{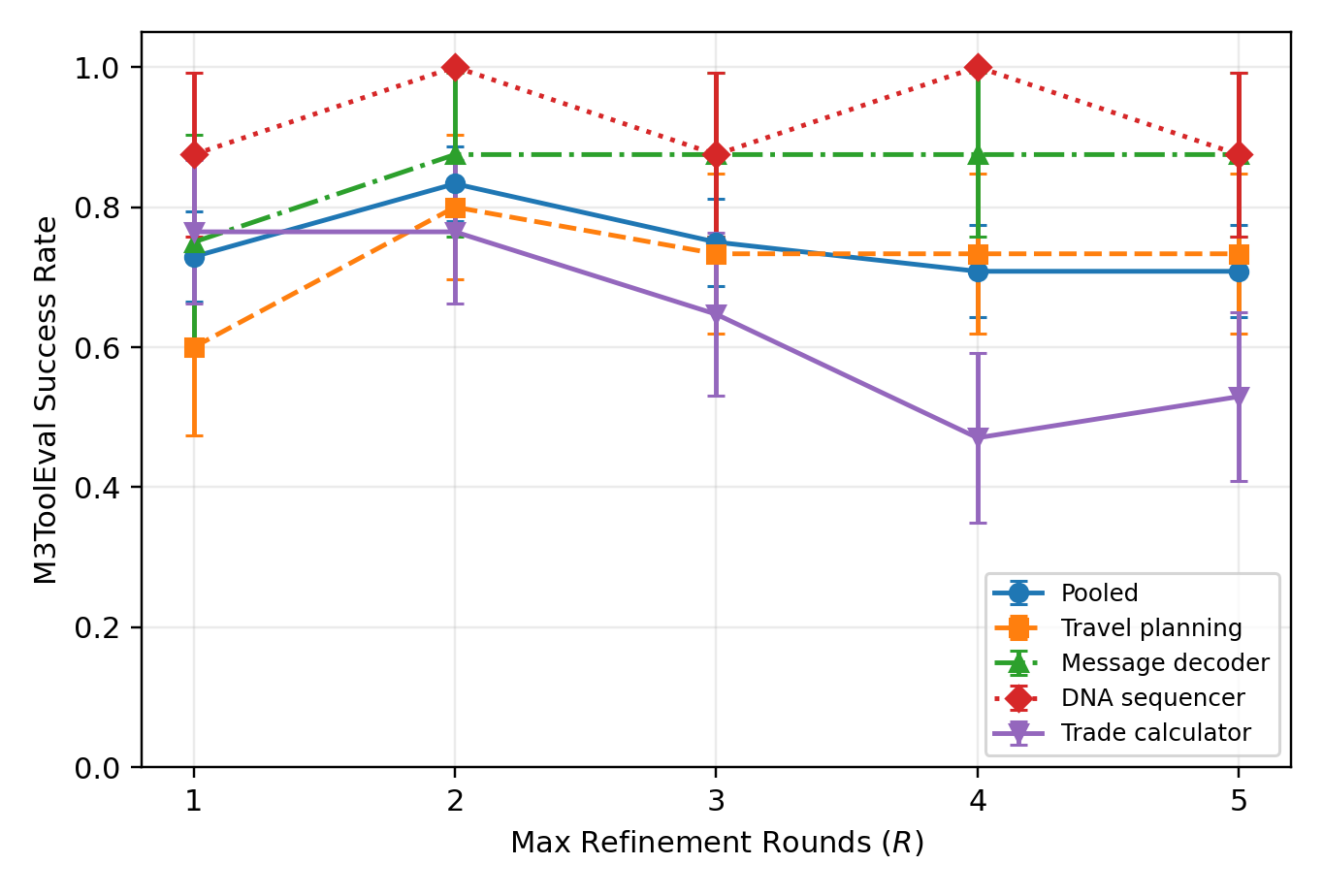}
\caption{Per-task-family breakdown of RubricRefine success rate by maximum refinement round~$R$ (\texttt{o3-mini}). Travel planning and message decoder saturate at $R=2$. Trade calculator does not benefit from additional rounds and degrades beyond $R=2$, consistent with rubric-guided contract checks being less effective for arithmetic-heavy tasks. Error bars show $\pm$1 SE.}
\label{fig:budget-scaling-family}
\end{figure}

Figure~\ref{fig:budget-scaling} reports the results. Panel~(a) shows RubricRefine success rate as a function of the maximum refinement round $R$. Nearly all gains materialize in the first refinement round ($R=1 \to R=2$): averaged across models, success jumps by $+0.17$ absolute in a single round and then plateaus. This confirms that the early-stopping mechanism (Section~\ref{sec:calibration}) captures most of the available improvement and that increasing $R$ beyond $2$--$3$ yields negligible additional benefit. One alternative interpretation of the R=2 saturation is that the generator's revision capacity is exhausted after one repair round regardless of verifier feedback quality, meaning the model cannot produce meaningfully different code on subsequent rounds. We cannot fully rule this out from the scaling curve alone; however, the calibration result (Section~\ref{sec:calibration}) provides indirect evidence against it: when the rubric score already reaches $10$ at $R=1$ (early stopping), the candidate is genuinely execution-ready at a higher rate than lower-scoring candidates, suggesting that the verifier's signal is informative and that saturation reflects successful repair rather than revision fatigue. A per-task-family breakdown of the $R$-scaling curve (Figure~\ref{fig:budget-scaling-family}, Appendix~\ref{app:budget-scaling}) further shows that saturation is not uniform: travel planning and message decoder saturate at $R=2$, while trade calculator (the most arithmetic-heavy family) does not improve with additional rounds and in fact degrades, consistent with the verifier's contract checks being less effective for arithmetic-intensive tasks.

Panel~(b) shows Best-of-$N$+rubric success rate as a function of the number of candidates $N$. In contrast to the sharp saturation of iterative refinement, rubric-guided selection improves more gradually: each additional candidate helps, but the curve is still rising at $N=5$ and has not yet reached the level that RubricRefine achieves at $R=2$. This comparison illustrates why iterative repair is a more compute-efficient use of rubric-structured checking than parallel selection in this setting: one refinement round with targeted feedback achieves a larger gain than four additional independently sampled candidates scored by the same rubric.

\subsection{Operational LM-Call Statistics}
\label{app:early-stopping}

RubricRefine's main efficiency claim can be stated directly in terms of observed inference usage. Table~\ref{tab:early-stopping} compares the observed average total LM calls per M3ToolEval task for RubricRefine against the strongest non-iterative rubric-guided baseline (Best-of-$N$+rubric with $N=5$) across all seven evaluated models, averaged over 10 trials.

\begin{table}[t]
\centering
\footnotesize
\setlength{\tabcolsep}{2pt}
\resizebox{\columnwidth}{!}{%
\begin{tabular}{lccc}
\toprule
Model & \shortstack{RubricRefine\\Avg.\ LM Calls} & \shortstack{BoN+rubric\\Avg.\ LM Calls} & \shortstack{Relative\\Reduction} \\
\midrule
\texttt{GPT-4.1-mini} & 8.71 & 15.16 & 42.6\% \\
\texttt{GPT-4o} & 7.26 & 16.79 & 56.8\% \\
\texttt{o3-mini} & 7.88 & 16.07 & 51.0\% \\
\texttt{GPT-4.1} & 8.13 & 17.33 & 53.1\% \\
\texttt{Gemma-4-26B} & 4.47 & 11.10 & 59.7\% \\
\texttt{Qwen3.6-27B} & 5.07 & 11.33 & 55.3\% \\
\texttt{Sonnet-4.6} & 5.02 & 11.95 & 58.0\% \\
\bottomrule
\end{tabular}%
}
\caption{Observed average total LM calls per M3ToolEval task for RubricRefine versus Best-of-$N$+rubric across the seven evaluated models, averaged over 10 trials per model.}
\label{tab:early-stopping}
\end{table}

\begin{table}[t]
\centering
\footnotesize
\setlength{\tabcolsep}{2pt}
\resizebox{\columnwidth}{!}{%
\begin{tabular}{lcccc}
\toprule
Model & Trials & \shortstack{RubricRefine\\Tokens/task} & \shortstack{BoN+rubric\\Tokens/task} & \shortstack{Relative\\Reduction} \\
\midrule
\texttt{GPT-4.1-mini} & 10 & 28{,}157 & 53{,}662 & 47.5\% \\
\texttt{GPT-4o}       & 1  & 26{,}272 & 52{,}831 & 50.3\% \\
\texttt{o3-mini}      & 10 & 28{,}028 & 53{,}648 & 47.8\% \\
\texttt{GPT-4.1}      & 10 & 27{,}676 & 53{,}687 & 48.4\% \\
\texttt{Sonnet-4.6}   & 8  & 46{,}230 & 81{,}610 & 43.4\% \\
\bottomrule
\end{tabular}%
}
\caption{Observed average total tokens per M3ToolEval task for RubricRefine versus Best-of-$N$+rubric, for the five models accessed through provider APIs. Trial counts differ by model and are stated per row; the \texttt{GPT-4o} row is a single trial and is reported for completeness rather than as a variance estimate. Token accounting for the two locally served open-weight models is reported in Figure~\ref{fig:gemma4-tradeoff}.}
\label{tab:token-reduction}
\end{table}

\paragraph{Token usage}
The same early-stopping behaviour shows up in token accounting. Table~\ref{tab:token-reduction} reports observed average total tokens per task for the five API-served models. RubricRefine uses $43$--$50\%$ fewer tokens than rubric-guided reranking on every model measured, and the four OpenAI models cluster tightly ($27.7$k--$28.2$k tokens per task) despite spanning a range of capabilities. We report tokens rather than wall-clock across models deliberately: token counts are a property of the method, whereas wall-clock across providers is dominated by serving latency and would not isolate method behaviour. Wall-clock is therefore reported for a single API model (Table~\ref{tab:efficiency}) and, in a controlled local-serving regime, for \texttt{Gemma-4-26B} (Appendix~\ref{app:gemma4-tradeoffs}).

Across models, RubricRefine reduces observed LM-call usage by $43$--$60\%$ relative to rubric-guided reranking while also improving or matching task success. The open-weight \texttt{Gemma-4-26B} shows the largest relative reduction ($59.8\%$), indicating that the early-stopping behavior underlying RubricRefine's inference efficiency is not specific to any particular model family. This direct LM-call reduction, together with the corresponding token and wall-clock reductions, is the operational evidence behind RubricRefine's inference-efficiency advantage; the per-round breakdown driving it is reported in Appendix~\ref{app:round-stopping}.

\subsection{Round-by-Round Stopping Distribution}
\label{app:round-stopping}

Table~\ref{tab:round-stopping} reports the fraction of M3ToolEval tasks that terminate at each refinement round for RubricRefine, averaged across 10 trials per model. The method's early-stopping behavior is tight across every model tested: $\approx\!90\%$ or more of tasks terminate by round~2, mean round counts fall in $[1.73, 2.03]$, and no model averages more than 2.03 rounds per task. Tasks stopping at rounds 1--2 achieve score $10$ in nearly all cases, confirming that early stopping at score $10$ is tightly aligned with the verifier's maximum-confidence judgment. Fewer than $10\%$ of tasks require three or more rounds on any model. This indicates that rubric-guided early stopping is a generic property of the method rather than an artifact of a particular backbone.

\begin{table}[t]
\centering
\footnotesize
\setlength{\tabcolsep}{2pt}
\resizebox{\columnwidth}{!}{%
\begin{tabular}{lccccccc}
\toprule
Round & \texttt{GPT-4.1-mini} & \texttt{GPT-4o} & \texttt{o3-mini} & \texttt{GPT-4.1} & \texttt{Gemma-4-26B} & \texttt{Qwen3.6-27B} & \texttt{Sonnet-4.6} \\
\midrule
1 & 33\% & 38\% & 35\% & 35\% & 31\% &  0\% &  0\% \\
2 & 57\% & 52\% & 56\% & 56\% & 66\% & 97\% & 99\% \\
3 &  7\% &  6\% &  6\% &  6\% &  4\% &  3\% &  1\% \\
4+ &  2\% &  4\% &  2\% &  2\% &  0\% &  0\% &  0\% \\
\midrule
Mean rounds & 1.79 & 1.76 & 1.76 & 1.76 & 1.73 & 2.03 & 2.01 \\
\bottomrule
\end{tabular}%
}
\caption{Fraction of M3ToolEval tasks terminating at each RubricRefine round, averaged across 10 trials. $\approx\!90\%$ of tasks reach score $10$ by round~2.}
\label{tab:round-stopping}
\end{table}

\section{Calibration Analysis: Why Ranking and Stopping Place Different Demands on the Verifier}
\label{app:calibration-analysis}

This appendix consolidates the calibration evidence behind two main-text claims: that RubricRefine remains effective on \texttt{Gemma-4-26B} despite poor aggregate verifier calibration (Section~\ref{sec:calibration}), and that this fact is structural rather than incidental. Section~\ref{app:gemma4-calibration} reports the empirical dissociation between rubric-guided ranking and rubric-guided stopping on \texttt{Gemma-4-26B}; Section~\ref{app:middle-bin-calibration} states the underlying calibration asymmetry independent of any specific model.

\subsection{Why \texttt{Gemma-4-26B} Fails at Rubric-Guided Ranking but Succeeds at Rubric-Guided Repair}
\label{app:gemma4-calibration}

The main-text results in Table~\ref{tab:main-m3-success} show a striking dissociation on \texttt{Gemma-4-26B}: rubric-guided \emph{selection} (BoN+rubric) drops to $0.50$, below unstructured Best-of-$N$, while rubric-guided \emph{iterative repair} (RubricRefine) reaches $0.85$, matching the strongest backbones evaluated. The two methods use the same model, the same rubric-generation prompt, and the same PASS/FAIL scoring protocol; they differ only in how the verifier's output is consumed. This subsection reports the calibration evidence that explains the dissociation.

Figure~\ref{fig:reliability-main} (right panel) shows the reliability diagram for Gemma-4's normalized RubricRefine scores on M3ToolEval (pooled across 10 trials; $n=580$ candidate trajectories). The overall calibration is materially worse than on frontier models: $\text{ECE} = 0.165$ versus $0.106$--$0.115$ on \texttt{GPT-4.1-mini} and \texttt{GPT-4.1} (Section~\ref{sec:calibration}). This is consistent with prior findings that smaller models tend to produce less calibrated self-assessments~\citep{kadavath2022language}; \texttt{Gemma-4-26B} is the smallest model we evaluate. The middle bins are the culprit. Scores in the $0.5$--$0.8$ range show non-monotone accuracy: the $0.6$ bin has accuracy $0.60$, but the $0.8$ bin has accuracy $0.00$. A candidate scored $0.8$ is therefore not reliably better than a candidate scored $0.6$; mid-range scores convey ordinal noise rather than ordinal signal.

The top bin of the RubricRefine reliability diagram tells a different story. The score-$=10$ bin contains $329$ of the $580$ pooled trajectories ($57\%$) and has accuracy $0.87$ despite confidence $1.0$ (a $0.13$ calibration gap), worse than the frontier models' top-bin alignment but still far better than any middle bin, and well-separated from the second-highest bin (accuracy $0.60$ at $0.65$). The corresponding AUROC is $0.795$, essentially identical to the frontier models ($0.796$--$0.764$). On these trajectories, the verifier is a capable ranker.

\begin{figure}[t]
\centering
\includegraphics[width=0.9\linewidth]{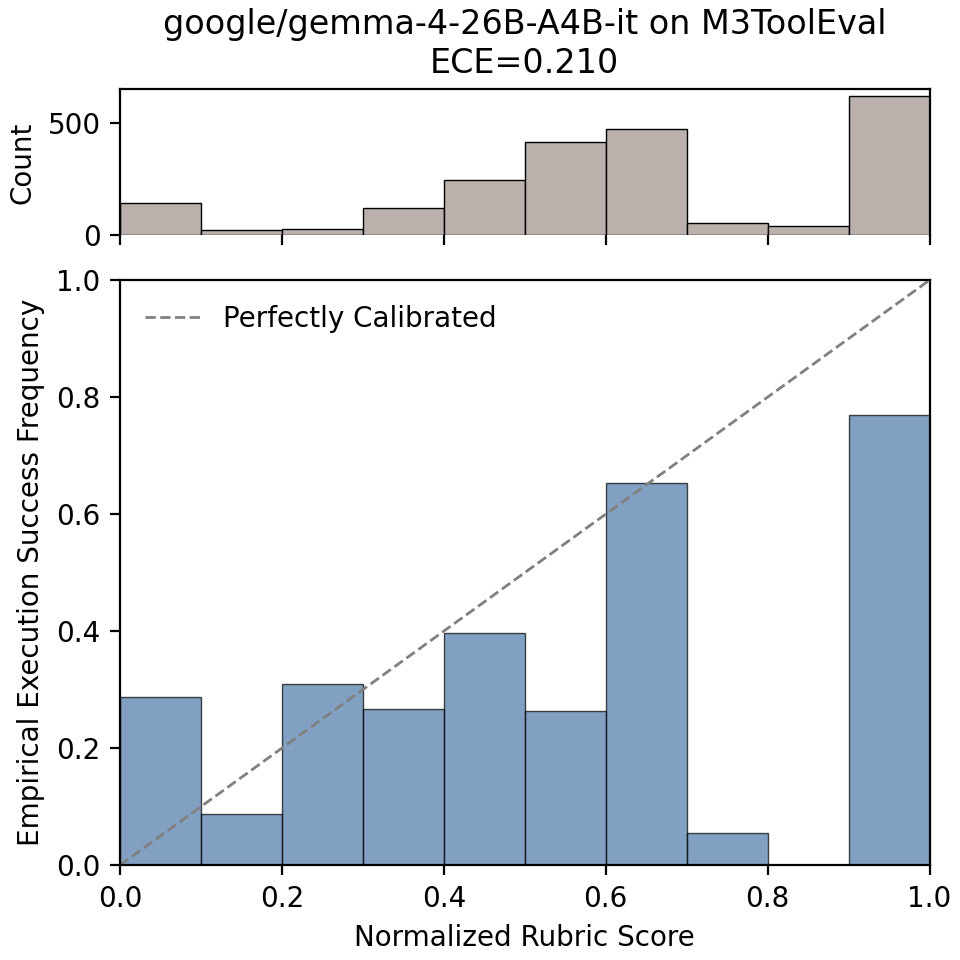}
\caption{Reliability diagram for \texttt{Gemma-4-26B}'s normalized rubric scores on \emph{BoN+rubric} trajectories on M3ToolEval (10 trials; $n=2{,}160$). Compare with Figure~\ref{fig:reliability-main} (right panel), which shows the same model and verifier on \emph{RubricRefine} trajectories. Overall $\text{ECE} = 0.210$, AUROC $= 0.700$, and top-bin accuracy is only $0.77$: strictly worse on every metric than the RubricRefine-trajectory version.}
\label{fig:gemma4-bon-reliability}
\end{figure}

\paragraph{But the AUROC on RubricRefine trajectories overstates what the verifier can do in general.}
To see why, we run the same calibration analysis on \emph{BoN+rubric} trajectories from the same model: same verifier, same rubric-generation prompt, same PASS/FAIL protocol, but the scored candidates come from 5 parallel samples of CodeAct rather than from iterative refinement. Figure~\ref{fig:gemma4-bon-reliability} shows the resulting reliability diagram. The verifier's AUROC drops from $0.795$ on RubricRefine trajectories to $0.700$ on BoN+rubric trajectories, and ECE worsens from $0.165$ to $0.210$. The top bin is also less reliable: the accuracy in the score-$=10$ bin is $0.77$ on BoN+rubric trajectories versus $0.87$ on RubricRefine trajectories, and on the hardest task family (travel itinerary planning) the BoN+rubric top bin is wrong $100\%$ of the time ($0/13$ across all trials; see inspection below).

This is not a property of the verifier alone (the verifier is identical in both experiments). It is a property of \emph{what candidates the verifier is scoring}. Inspection of the BoN+rubric trajectories on travel itinerary planning makes the mechanism concrete: across 135 BoN+rubric task instances (9 trials $\times$ 15 tasks), only $10\%$ of 5-candidate sets contain any candidate that reaches score $10$, and even among those 13 sets the selected top-scored candidate is never correct at execution time. The underlying reason is that \texttt{Gemma-4}'s CodeAct baseline on travel planning is $0.02$ (Appendix~\ref{app:frontier-per-task}): five parallel CodeAct samples almost never contain a correct program, so no ranking strategy, however well-calibrated, can recover one. On easier task families (dna sequencer, trade calculator), where CodeAct already produces mostly-correct candidates, the BoN+rubric top bin is more reliable ($100\%$ and $82\%$ accuracy when score$=10$ respectively).

\paragraph{The mechanism argument, stated carefully.}
The apparent dissociation between BoN+rubric and RubricRefine on \texttt{Gemma-4} therefore has two components, not one. First, the \emph{calibration-ranking component}: middle-bin scores are non-monotone as a ranker (e.g., the $0.8$ bin has accuracy $0.00$ while the $0.6$ bin has accuracy $0.60$ on RubricRefine trajectories), so any method that consumes the full score distribution as a ranking signal inherits this noise. Second, the \emph{candidate-supply component}: parallel sampling from a weak-baseline generator rarely contains a correct candidate on hard tasks, so selection cannot recover what was never sampled. Iterative repair sidesteps both components because it does not rank across candidates: it uses the verifier's PASS/FAIL decomposition to \emph{construct} a candidate that meets each criterion. When a RubricRefine trajectory reaches score $10$, the candidate has been shaped by the rubric's item-level directives, not merely selected from a pre-existing pool. This is why the top-bin accuracy is higher on RubricRefine trajectories than on BoN+rubric trajectories even though the verifier is the same: the distribution of candidates being scored is qualitatively different.

This observation also explains why AUROC alone is not a sufficient metric for the calibration thesis. The AUROC of $0.795$ on RubricRefine trajectories reflects the verifier's ability to rank \emph{the specific candidates RubricRefine produces}, many of which have been refined to the top bin by construction. The AUROC of $0.700$ on BoN+rubric trajectories reflects the same verifier's ability to rank \emph{cold-sampled CodeAct candidates}, which it handles less well because the signal is dominated by middle-bin miscalibration on flawed candidates. Section~\ref{sec:calibration}'s claim that RubricRefine needs only top-bin reliability is therefore correct but narrower than it sounds: the top-bin reliability it relies on is \emph{top-bin reliability on refinement-loop trajectories specifically}, which is produced jointly by the verifier's recognition capability and the refinement loop's ability to drive candidates toward rubric-passing forms.

The practical implication is that a verifier need not be a good general-purpose ranker to support effective rubric-guided repair; it needs only to reliably recognize candidates that the refinement loop has shaped to pass its own criteria. This is a weaker and more plausible requirement than global calibration, and is what makes RubricRefine effective on models whose aggregate verifier calibration is poor.

\subsection{Why Middle-Bin Calibration Matters for Ranking but Not for Top-Bin Stopping}
\label{app:middle-bin-calibration}

This subsection makes explicit why rubric-guided selection (Best-of-$N$+rubric) and rubric-guided stopping (RubricRefine's score-$=10$ trigger) place fundamentally different demands on the verifier's calibration profile, independent of any specific model.

\paragraph{What rubric-guided selection needs.}
Best-of-$N$+rubric samples $N$ candidates, scores each against a rubric, and picks the highest-scoring one. The selection is correct whenever the candidate with the highest execution accuracy also has the highest rubric score. Formally, for any pair of candidates $(c_i, c_j)$ with rubric scores $(s_i, s_j)$, the verifier must satisfy
\[
\mathbb{P}[\text{correct}(c_i)] > \mathbb{P}[\text{correct}(c_j)] \;\Longrightarrow\; s_i > s_j
\]
everywhere the two candidates' scores land, including the middle bins. This is a \emph{monotone ranking} requirement: the score function must preserve the order of candidate quality across the full score range. When the middle bins are miscalibrated (i.e., when two candidates scored $0.6$ and $0.8$ have accuracies $0.60$ and $0.00$ respectively, as on Gemma-4 in this evaluation), the monotone-ranking assumption is violated in exactly the region where most candidates' scores actually fall. Selection then systematically picks worse candidates and can underperform random selection.

\paragraph{What RubricRefine's early-stopping needs.}
RubricRefine does not rank candidates against each other. It generates a single candidate per round and asks the verifier one binary question: is the rubric score at its maximum? If yes, stop; if no, use the itemized PASS/FAIL feedback to drive a revision and produce the next candidate. The only calibration property required is that the \emph{top bin} be accurate: $\mathbb{P}[\text{correct}(c) \mid s(c) = s_{\max}]$ must be high enough to make stopping safe.
No assumption is made about scores below the maximum, because no decision depends on them. Scores of $0.5$, $0.6$, or $0.8$ all trigger the same action: revise. A candidate scored $0.4$ and a candidate scored $0.8$ are treated identically by the early-stopping rule, so the fact that their true accuracies are swapped relative to their scores (on Gemma-4) has no effect on the method's behavior.

\paragraph{The asymmetry in one sentence.}
Selection is a ranking problem and requires higher rubric scores to indicate higher actual correctness across the full score range; early stopping is a thresholding problem at the maximum score and requires only that the maximum-score bin be reliable. This is why the same verifier can fail as a ranker while still serving as an effective stopping signal, and is the structural reason RubricRefine remains effective on models like \texttt{Gemma-4-26B} whose aggregate calibration is poor.

\paragraph{Itemized PASS/FAIL versus scalar score.}
A second, complementary property helps iterative repair but not selection. RubricRefine's revision step consumes the verifier's per-criterion PASS/FAIL judgments rather than the scalar score; even when the aggregate score is noisy, the specific failing criteria can be informative (``output is a string but should be a float'' is a useful directive regardless of whether the overall score is $5$ or $7$). Selection, by contrast, collapses the full rubric output into a single scalar for ranking and therefore cannot exploit criterion-level diagnostic information. This compounds the calibration asymmetry above: iterative repair recovers useful signal from exactly the dimension that selection throws away.

\section{Qualitative Example: A Rubric-Caught Silent Contract Failure}
\label{app:qualitative-example}

This appendix shows a concrete case drawn from our \texttt{gpt-4.1} M3ToolEval run where RubricRefine's rubric diagnoses a contract violation that would be invisible to execution-based feedback. The failure is not an exception or crash; the initial candidate runs to completion and produces a wrong answer silently. Execution feedback therefore offers no diagnostic signal, which is the failure mode Section~\ref{sec:main-results} identifies as the dominant one in this setting.

\paragraph{Task.}
The task (\texttt{message\_decoder/}\allowbreak\texttt{hex\_caesar\_combined\_decoding}) asks the agent to ``decode a message that was first converted to hexadecimal, then encoded with a Caesar cipher with a shift of 2'' from the hex-and-Caesar-shifted input \texttt{'4d4f5252'}. The tool registry exposes \texttt{convert\_hex\_to\_ascii} and \texttt{caesar\_decode}, among others. Ground-truth output: \texttt{KMPP}.

\paragraph{Initial candidate (round 0, rubric score 3/10).}
The generator reverses the required operation order:
\begin{verbatim}
Action:
decoded_caesar = caesar_decode(
    '4d4f5252', 2)
ascii_message = convert_hex_to_ascii(
    decoded_caesar)
print(ascii_message)
End Action
\end{verbatim}
This code is \emph{executable}: every call has a valid signature, every argument has the right type, and the program terminates normally, producing a garbage string rather than \texttt{KMPP}. An execution-feedback system (Self-Debug) sees a successful run that emits a wrong answer it cannot re-diagnose; there is no stack trace or error message to repair from.

\paragraph{Rubric feedback (verbatim PASS/FAIL items).}
The task-specific rubric generated for this instance explicitly encodes the inter-tool ordering and dataflow constraints. The verifier's round-0 feedback marks:
\begin{itemize}\itemsep0pt
    \item \textbf{intent~A} (\textsc{fail}): ``The candidate reverses the order of operations, attempting to decode the Caesar cipher first instead of decoding the hex-encoded message into ASCII.''
    \item \textbf{ordering\_dataflow~D1} (\textsc{fail}): ``The candidate calls \texttt{caesar\_decode} before \texttt{convert\_hex\_to\_ascii}, violating the required order of operations.''
    \item \textbf{ordering\_dataflow~D2} (\textsc{fail}): ``The output of \texttt{convert\_hex\_to\_ascii} is not passed as the \texttt{'message'} argument to \texttt{caesar\_decode}.''
    \item \textbf{execution\_critical~E1/E2} (\textsc{fail}): ``The hex-encoded message '4d4f5252' is not decoded using \texttt{convert\_hex\_to\_ascii} first, as required by the task.''
    \item \textbf{final\_answer~F2} (\textsc{fail}): ``The emitted answer is not the correct decoded message derived from the hex-encoded, Caesar-shifted input.''
\end{itemize}
Several other items (type-shape contracts, Caesar shift value, \texttt{print} format) are correctly marked \textsc{pass}, localizing the failure to the ordering and dataflow dimensions rather than to syntactic or surface concerns.

\paragraph{Repair (round 1, rubric score 10/10).}
The repair prompt receives the item-level PASS/FAIL structure and rewrites the candidate so that the hex-to-ASCII step precedes the Caesar decoding, with the output of the first call feeding the \texttt{message} argument of the second:
\begin{verbatim}
Action:
ascii_message = convert_hex_to_ascii(
    hex_string='4d4f5252')
decoded_message = caesar_decode(
    message=ascii_message, shift=2)
print(decoded_message)
End Action
\end{verbatim}
The verifier scores this 10/10, triggering early stopping. On execution, the program emits \texttt{KMPP}, matching the ground truth.

\paragraph{Why execution feedback cannot catch this.}
The round-0 program is contract-violating but not exception-raising: it calls documented tools with documented signatures and produces a printed string that is merely semantically wrong. Self-Debug observes the execution trace (a successful run and a printed answer) but not the ground-truth comparison. Because Self-Debug retries only in response to runtime exceptions, silent wrong-answer failures like this one terminate with no repair attempt: there is no error for execution feedback to act on. The rubric, in contrast, encodes the ordering and dataflow contract before execution and surfaces the violation as a specific item-level directive that the generator can act on. This is the mechanism behind the main-text finding that pre-execution contract-structured feedback recovers failures that execution feedback cannot.

\section{Logprob-Weighted Scoring Variants}
\label{app:logprob-results}

This appendix reports the full results for logprob-weighted scoring variants, which use the token-probability-weighted expected-value scoring mechanism of LLM-as-a-Verifier~\citep{kwok2026llmverifier}. Because \texttt{o3-mini} does not expose token-level logprobs, these variants are evaluated on three models.

\paragraph{Method Descriptions}
\begin{itemize}
\item \textbf{Best-of-$N$ (logprob)} samples $N$ candidates and scores each using token-probability-weighted expected-value scoring (from top-$k$ token logprobs) over a discrete 20-point letter-grade scale (A--T). The verifier is prompted to emit a single letter grade; the score is computed as the probability-weighted mean of all valid grade tokens in the top-$k$ logprobs, normalized to $[0,1]$. This variant isolates the effect of logprob-based scoring without any rubric structure.
\item \textbf{Best-of-$N$+fixed rubric (logprob)} combines the fixed, task-independent rubric with logprob-weighted scoring. This isolates the effect of rubric specificity within the logprob scoring family.
\item \textbf{Best-of-$N$+rubric (logprob)} combines sample-dependent rubric scoring with token-probability-weighted aggregation: each candidate is scored against the task-specific rubric using the same expected-value mechanism derived from top-$k$ token logprobs. This tests whether rubric structure provides additional value beyond improved scoring calibration.
\end{itemize}

\paragraph{Results}
Tables~\ref{tab:logprob-m3} and~\ref{tab:logprob-apibank} report success rates for the logprob variants alongside the non-logprob baselines and RubricRefine for reference. On M3ToolEval, logprob scoring consistently improves over plain Best-of-$N$, and adding rubric structure on top of logprob scoring yields further gains. On API-Bank, logprob variants cluster near the baseline, consistent with the main-text pattern.

\begin{table}[t]
\centering
\footnotesize
\setlength{\tabcolsep}{2pt}
\resizebox{\columnwidth}{!}{%
\begin{tabular}{lccc}
\toprule
Method & \texttt{GPT-4.1-mini} & \texttt{GPT-4o} & \texttt{GPT-4.1} \\
\midrule
Best-of-$N$ & $.65 \pm .02$ & $.65 \pm .02$ & $.62 \pm .01$ \\
BoN (logprob) & $.70 \pm .02$ & $.66 \pm .01$ & $.68 \pm .01$ \\
BoN+fixed rubric logprob & $.74 \pm .01$ & $.76 \pm .01$ & $.73 \pm .02$ \\
BoN+rubric logprob & $.78 \pm .01$ & $.76 \pm .01$ & $.76 \pm .01$ \\
\midrule
RubricRefine (Ours) & $.86 \pm .01$ & $.86 \pm .01$ & $.85 \pm .01$ \\
\bottomrule
\end{tabular}%
}
\caption{M3ToolEval success rates (mean $\pm$ SE across trials) for logprob-weighted scoring variants. Plain Best-of-$N$ and RubricRefine are included for reference. Logprob variants are not available for \texttt{o3-mini} (no token-level logprob support).}
\label{tab:logprob-m3}
\end{table}

\begin{table}[t]
\centering
\footnotesize
\setlength{\tabcolsep}{2pt}
\resizebox{\columnwidth}{!}{%
\begin{tabular}{lccc}
\toprule
Method & \texttt{GPT-4.1-mini} & \texttt{GPT-4o} & \texttt{GPT-4.1} \\
\midrule
Best-of-$N$ & $.73 \pm .00$ & $.73 \pm .00$ & $.73 \pm .00$ \\
BoN (logprob) & $.75 \pm .01$ & $.76 \pm .01$ & $.77 \pm .01$ \\
BoN+fixed rubric logprob & $.75 \pm .01$ & $.77 \pm .01$ & $.77 \pm .01$ \\
BoN+rubric logprob & $.76 \pm .01$ & $.76 \pm .01$ & $.77 \pm .01$ \\
\midrule
RubricRefine (Ours) & $.69 \pm .01$ & $.70 \pm .01$ & $.69 \pm .01$ \\
\bottomrule
\end{tabular}%
}
\caption{API-Bank success rates (mean $\pm$ SE across trials) for logprob-weighted scoring variants. Plain Best-of-$N$ and RubricRefine are included for reference.}
\label{tab:logprob-apibank}
\end{table}

\section{Artifacts and Licenses}
\label{app:artifacts}

This work uses the following existing artifacts. \textbf{M3ToolEval} is used through the CodeAct evaluation ecosystem~\citep{wang2024executable}, and \textbf{API-Bank} is used through its released benchmark data, API implementations, and evaluation code~\citep{li2023apibank}. We adapt the benchmark runners locally to support code-as-action outputs and method comparisons, while preserving the benchmark-level correctness semantics described in Appendices~\ref{app:m3} and~\ref{app:apibank}. Both benchmarks are used under their original public release terms for their intended research-evaluation purpose. The benchmark task instructions and tool documentation in both M3ToolEval and API-Bank are in English. \textbf{Gemma-4-26B-A4B-it}~\citep{googledeepmind2026gemma4} is used under the Gemma Terms of Use; \textbf{Qwen3.6-27B-FP8}~\citep{qwen3.6-27b} is released under the Apache 2.0 license; \textbf{GPT-4.1-mini}, \textbf{GPT-4o}, \textbf{o3-mini}, and \textbf{GPT-4.1} are accessed through the OpenAI API under the OpenAI Services Agreement; and \textbf{Claude-Sonnet-4.6} is accessed through the Anthropic API under Anthropic's Commercial Terms of Service. We do not redistribute any of these artifacts and use them only for research evaluation, consistent with their stated intended use. We release no new datasets or model weights with this submission.

\paragraph{Prompt and Code Release}
RubricRefine is an inference-time harness: a generator--verifier control loop with a defined state (the rubric), a scoring function, a stopping rule, and a repair policy. Reproducing it therefore requires both the control logic and the instructions that instantiate each role, so we release both rather than describing them. The complete rubric-generation, rubric-scoring, and repair prompts --- together with the evaluation harness used to produce the reported M3ToolEval results --- are available for review at:

\begin{center}
\url{https://anonymous.4open.science/r/RubricRefinePrivate-89FD/}
\end{center}

\noindent The released prompts are the exact versions used for the reported experiments and are reproduced in full in Appendix~\ref{app:prompts}. Upon acceptance we will additionally deposit the same contents in a permanent, versioned archive with a DOI (e.g., Zenodo) under the MIT license, so that the released artifact is citable and immutable rather than dependent on a mutable hosting URL.

\section{AppWorld Multi-Turn Details}
\label{app:appworld}

\paragraph{Scale and sampling}
We evaluate on the AppWorld train split ($90$ tasks: $36$ at difficulty~1, $36$ at difficulty~2, $18$ at difficulty~3). Figure~\ref{fig:appworld} pools $900$ paired episodes from 10 trials at temperature $0.7$; the difficulty-stratified breakdown is left to future work.

\paragraph{Protocol}
Each episode permits up to eight executed turns. At every turn the generator emits one block of code; under RubricRefine that block is scored against a freshly generated rubric and repaired before it is executed, so no check consumes an interaction. Both arms receive an identical task-scoped registry, obtained by querying AppWorld's \texttt{api\_docs} search with the verbatim task instruction (BM25 over the documentation index) and appending the login and supervisor endpoints that every task requires procedurally and that lexical retrieval never surfaces. Retrieval reads documentation only and mutates no state.

\paragraph{Namespace exposure}
In a stateful interpreter each block inherits the variables bound by earlier turns. A refiner that rewrites a block in isolation can silently rename or drop those bindings, which surfaces as a \texttt{NameError} at execution time. We therefore query the live interpreter for its user-defined bindings, supply them to the rubric, scoring, and repair prompts as ground truth, and statically reject any revision that references a name which is neither live nor bound within the block itself. Without this, refinement introduced roughly twice as many execution failures as it repaired and RubricRefine underperformed the baseline; with it, refinement-induced \texttt{NameError}s are eliminated by construction.

\paragraph{Significance testing}
Significance is assessed on the paired data, since both methods run identical tasks. A paired test is more powerful than comparing per-method intervals, so the Wilson bands in Figure~\ref{fig:appworld} may overlap at budgets where the paired difference is reliable.

\paragraph{Reading the budget curve}
The curve is a re-analysis of episodes run with an eight-turn cap: a point at $k$ counts episodes that completed within $k$ executed turns and passed evaluation. The agent was not told its budget in advance, so an agent optimising against a known cap of $k$ turns could plausibly do better by batching more work per block. The curves therefore bound a budget-constrained agent from below rather than measuring one directly.

\section{Personal Data and Human Subjects}
\label{app:personal-data}

We do not collect new data from people, use human annotators, or conduct human-subject experiments. The empirical study uses existing executable tool-use benchmark tasks and model/API outputs generated for this evaluation. We do not curate or release personally identifying information, and our contribution does not introduce a new dataset containing human-subject data.

\section{Computing Infrastructure}
\label{app:compute}

Open-weight models (\texttt{Gemma-4-26B-A4B-it} and \texttt{Qwen3.6-27B-FP8}) were served locally via vLLM on a node with $8\times$ NVIDIA L4 GPUs. Frontier models (\texttt{GPT-4.1-mini}, \texttt{GPT-4o}, \texttt{o3-mini}, \texttt{GPT-4.1}, and \texttt{Claude-Sonnet-4.6}) were accessed through the OpenAI and Anthropic APIs. Because RubricRefine is a training-free inference-time method, we report inference-time compute budgets in the units that determine deployment cost: wall-clock latency, total LM calls, and total tokens (Section~\ref{sec:efficiency}, Table~\ref{tab:efficiency}, and Appendix~\ref{app:scaling}).

\section{Use of AI Assistants}
\label{app:ai-assistance}

The authors used AI assistance (Claude, Anthropic) for editing and revision of manuscript text and for coding assistance in the implementation of experiments. All scientific claims, experimental design, analysis, and final manuscript content were reviewed and verified by the authors, who take full responsibility for the work.

\end{document}

%% file: references.tex
{\small
\bibliographystyle{acl_natbib}

}

%% file: codemode_paper.bbl
\begin{thebibliography}{99}

\bibitem[He et~al.(2026)]{he2026sdzero}
Yinghui He, Simran Kaur, Adithya Bhaskar, Yongjin Yang, Jiarui Liu, Narutatsu Ri, Liam Fowl, Abhishek Panigrahi, Danqi Chen, and Sanjeev Arora.
\newblock Self-distillation zero: Self-revision turns binary rewards into dense supervision.
\newblock arXiv:2604.12002, 2026.
\newblock URL: \url{https://arxiv.org/abs/2604.12002}.

\bibitem[Wang et~al.(2024)]{wang2024executable}
Xingyao Wang, Yangyi Chen, Lifan Yuan, Yizhe Zhang, Yunzhu Li, Hao Peng, and Heng Ji.
\newblock Executable code actions elicit better {LLM} agents.
\newblock In \emph{Proceedings of ICML}, 2024.
\newblock arXiv:2402.01030.
\newblock URL: \url{https://proceedings.mlr.press/v235/wang24h.html}.

\bibitem[Li et~al.(2023)]{li2023apibank}
Minghao Li, Yingxiu Zhao, Bowen Yu, Feifan Song, Hangyu Li, Haiyang Yu, Zhoujun Li, Fei Huang, and Yongbin Li.
\newblock {API-Bank}: A comprehensive benchmark for tool-augmented {LLMs}.
\newblock In \emph{Proceedings of EMNLP}, 2023.
\newblock DOI: \url{https://doi.org/10.18653/v1/2023.emnlp-main.187}.
\newblock URL: \url{https://aclanthology.org/2023.emnlp-main.187/}.

\bibitem[Madaan et~al.(2023)]{madaan2023selfrefine}
Aman Madaan et~al.
\newblock Self-refine: Iterative refinement with self-feedback.
\newblock In \emph{Proceedings of NeurIPS}, 2023.
\newblock URL: \url{https://openreview.net/forum?id=S37hOerQLB}.

\bibitem[Kadavath et~al.(2022)]{kadavath2022language}
Saurav Kadavath et~al.
\newblock Language models (mostly) know what they know.
\newblock arXiv:2207.05221, 2022.
\newblock URL: \url{https://arxiv.org/abs/2207.05221}.

\bibitem[Niculescu-Mizil and Caruana(2005)]{niculescumizil2005predicting}
Alexandru Niculescu-Mizil and Rich Caruana.
\newblock Predicting good probabilities with supervised learning.
\newblock In \emph{Proceedings of ICML}, 2005.
\newblock DOI: \url{https://doi.org/10.1145/1102351.1102430}.

\bibitem[Guo et~al.(2017)]{guo2017calibration}
Chuan Guo, Geoff Pleiss, Yu Sun, and Kilian~Q. Weinberger.
\newblock On calibration of modern neural networks.
\newblock In \emph{Proceedings of ICML}, 2017.
\newblock URL: \url{https://proceedings.mlr.press/v70/guo17a.html}.

\bibitem[Kull et~al.(2017)]{kull2017beta}
Meelis Kull, Telmo Silva Filho, and Peter Flach.
\newblock Beta calibration: a well-founded and easily implemented improvement on logistic calibration for binary classifiers.
\newblock In \emph{Proceedings of AISTATS}, 2017.
\newblock URL: \url{https://proceedings.mlr.press/v54/kull17a.html}.

\bibitem[Roucher et~al.(2025)]{roucher2024smolagents}
Aymeric Roucher, Albert Villanova del Moral, Thomas Wolf, Leandro von Werra, and Erik Kaunism\"aki.
\newblock smolagents: a smol library to build great agentic systems.
\newblock GitHub repository, 2025.
\newblock URL: \url{https://github.com/huggingface/smolagents}.

\bibitem[Varda and Pai(2025)]{cloudflare2025codemode}
Kenton Varda and Sunil Pai.
\newblock Code Mode: the better way to use {MCP}.
\newblock Cloudflare blog, 2025.
\newblock URL: \url{https://blog.cloudflare.com/code-mode/}.

\bibitem[Snell et~al.(2024)]{snell2024scaling}
Charlie Snell, Jaehoon Lee, Kelvin Xu, and Aviral Kumar.
\newblock Scaling {LLM} test-time compute optimally can be more effective than
scaling model parameters.
\newblock arXiv:2408.03314, 2024.
\newblock URL: \url{https://arxiv.org/abs/2408.03314}.

\bibitem[Kamoi et~al.(2024)]{kamoi2024selfcorrection}
Ryo Kamoi, Yusen Zhang, Nan Zhang, Jiawei Han, and Rui Zhang.
\newblock When can {LLMs} actually correct their own mistakes? A critical survey of self-correction of {LLMs}.
\newblock \emph{TACL}, 2024.
\newblock DOI: \url{https://doi.org/10.1162/tacl_a_00713}.
\newblock URL: \url{https://aclanthology.org/2024.tacl-1.78/}.

\bibitem[Wang et~al.(2026)]{wang2026preflect}
Hanyu Wang, Yuanpu Cao, Lu Lin, and Jinghui Chen.
\newblock {PreFlect}: From retrospective to prospective reflection in large language model agents.
\newblock arXiv:2602.07187, 2026.
\newblock URL: \url{https://arxiv.org/abs/2602.07187}.

\bibitem[Gou et~al.(2024)]{gou2024critic}
Zhibin Gou et~al.
\newblock {CRITIC}: Large language models can self-correct with tool-interactive critiquing.
\newblock In \emph{Proceedings of ICLR}, 2024.
\newblock URL: \url{https://openreview.net/forum?id=Sx038qxjek}.

\bibitem[Liu et~al.(2024)]{liu2024toolace}
Weiwen Liu et~al.
\newblock {ToolACE}: Winning the points of {LLM} function calling.
\newblock arXiv:2409.00920, 2024.
\newblock URL: \url{https://arxiv.org/abs/2409.00920}.

\bibitem[Chen et~al.(2025)]{chen2025button}
Mingyang Chen et~al.
\newblock Facilitating multi-turn function calling for {LLMs} via compositional instruction tuning.
\newblock In \emph{Proceedings of ICLR}, 2025.
\newblock URL: \url{https://openreview.net/forum?id=owP2mymrTD}.

\bibitem[Ma et~al.(2025)]{ma2025toolmvr}
Zhiyuan Ma et~al.
\newblock Advancing tool-augmented large language models via meta-verification and reflection learning.
\newblock In \emph{Proceedings of KDD}, 2025.
\newblock DOI: \url{https://doi.org/10.1145/3711896.3736835}.

\bibitem[Hao et~al.(2025)]{hao2025funreason}
Bingguang Hao et~al.
\newblock {FunReason}: Enhancing large language models' function calling via self-refinement multiscale loss and automated data refinement.
\newblock arXiv:2505.20192, 2025.
\newblock URL: \url{https://arxiv.org/abs/2505.20192}.

\bibitem[Zhang et~al.(2025)]{zhang2025tooln1}
Shaokun Zhang et~al.
\newblock Nemotron-Research-Tool-N1: Exploring tool-using language models with reinforced reasoning.
\newblock arXiv:2505.00024, 2025.
\newblock URL: \url{https://arxiv.org/abs/2505.00024}.

\bibitem[Feng et~al.(2025)]{feng2025retool}
Jiazhan Feng et~al.
\newblock {ReTool}: Reinforcement learning for strategic tool use in {LLMs}.
\newblock arXiv:2504.11536, 2025.
\newblock URL: \url{https://arxiv.org/abs/2504.11536}.

\bibitem[Lu et~al.(2024)]{lu2024gear}
Yining Lu, Haoping Yu, and Daniel Khashabi.
\newblock {GEAR}: Augmenting language models with generalizable and efficient tool resolution.
\newblock In \emph{Proceedings of EACL}, 2024.
\newblock URL: \url{https://aclanthology.org/2024.eacl-long.7/}.

\bibitem[Wu et~al.(2025)]{wu2025chainoftools}
Mengsong Wu et~al.
\newblock Chain-of-Tools: Utilizing massive unseen tools in the {CoT} reasoning of frozen language models.
\newblock arXiv:2503.16779, 2025.
\newblock URL: \url{https://arxiv.org/abs/2503.16779}.

\bibitem[Lumer et~al.(2025)]{lumer2025graphragtoolfusion}
Elias Lumer et~al.
\newblock Graph {RAG}-Tool Fusion.
\newblock arXiv:2502.07223, 2025.
\newblock URL: \url{https://arxiv.org/abs/2502.07223}.

\bibitem[Qin et~al.(2024)]{qin2024toolllm}
Yujia Qin, Shihao Liang, Yining Ye, Kunlun Zhu, Lan Yan, Yaxi Lu, Yankai Lin, Xin Cong, Xiangru Tang, Bill Qian, Sihan Zhao, Lauren Hong, Runchu Tian, Ruobing Xie, Jie Zhou, Mark Gerstein, Dahai Li, Zhiyuan Liu, and Maosong Sun.
\newblock ToolLLM: Facilitating large language models to master 16000+ real-world {APIs}.
\newblock In \emph{Proceedings of ICLR}, 2024.
\newblock URL: \url{https://openreview.net/forum?id=dHng2O0Jjr}.

\bibitem[Patil et~al.(2025)]{berkeley-function-calling-leaderboard}
Shishir~G. Patil, Huanzhi Mao, Fanjia Yan, Charlie Cheng-Jie Ji, Vishnu Suresh, Ion Stoica, and Joseph~E. Gonzalez.
\newblock The Berkeley function calling leaderboard ({BFCL}): From tool use to agentic evaluation of large language models.
\newblock In \emph{Proceedings of ICML}, 2025.
\newblock URL: \url{https://proceedings.mlr.press/v267/patil25a.html}.

\bibitem[Bai et~al.(2022)]{bai2022constitutional}
Yuntao Bai et~al.
\newblock Constitutional {AI}: Harmlessness from {AI} feedback.
\newblock arXiv:2212.08073, 2022.
\newblock URL: \url{https://arxiv.org/abs/2212.08073}.

\bibitem[Zheng et~al.(2023)]{zheng2023judging}
Lianmin Zheng et~al.
\newblock Judging {LLM}-as-a-judge with {MT-Bench} and {Chatbot Arena}.
\newblock In \emph{Proceedings of NeurIPS}, 2023.
\newblock URL: \url{https://proceedings.neurips.cc/paper_files/paper/2023/hash/91f18a1287b398d378ef22505bf41832-Paper-Datasets_and_Benchmarks.pdf}.

\bibitem[Kim et~al.(2024)]{kim2024prometheus}
Seungone Kim et~al.
\newblock Prometheus: Inducing fine-grained evaluation capability in language models.
\newblock In \emph{Proceedings of ICLR}, 2024.
\newblock URL: \url{https://openreview.net/forum?id=8euJaTveKw}.

\bibitem[Sharma et~al.(2025)]{sharma2025researchrubrics}
Manasi Sharma et~al.
\newblock {ResearchRubrics}: A benchmark of prompts and rubrics for evaluating deep research agents.
\newblock arXiv:2511.07685, 2025.
\newblock URL: \url{https://arxiv.org/abs/2511.07685}.

\bibitem[Gunjal et~al.(2025)]{gunjal2025rubricsrewards}
Anisha Gunjal et~al.
\newblock Rubrics as Rewards: Reinforcement learning beyond verifiable domains.
\newblock arXiv:2507.17746, 2025.
\newblock URL: \url{https://arxiv.org/abs/2507.17746}.

\bibitem[Raghavendra et~al.(2026)]{raghavendra2026agenticrubrics}
Mohit Raghavendra et~al.
\newblock Agentic Rubrics as contextual verifiers for {SWE} agents.
\newblock arXiv:2601.04171, 2026.
\newblock URL: \url{https://arxiv.org/abs/2601.04171}.

\bibitem[DeGroot and Fienberg(1983)]{degroot1983comparing}
Morris~H. DeGroot and Stephen~E. Fienberg.
\newblock The comparison and evaluation of forecasters.
\newblock \emph{Journal of the Royal Statistical Society: Series D (The Statistician)}, 32(1-2):12--22, 1983.
\newblock DOI: \url{https://doi.org/10.2307/2987588}.

\bibitem[Naeini et~al.(2015)]{10.5555/2888116.2888120}
Mahdi~Pakdaman Naeini, Gregory~F. Cooper, and Milos Hauskrecht.
\newblock Obtaining well calibrated probabilities using {Bayesian} binning.
\newblock In \emph{Proceedings of AAAI}, 2015.
\newblock URL: \url{https://ojs.aaai.org/index.php/AAAI/article/view/9602}.

\bibitem[LeVine et~al.(2023)]{levine2023calibration}
Will LeVine, Benjamin Pikus, Pranav Raja, and Fernando Amat~Gil.
\newblock Enabling calibration in the zero-shot inference of large vision-language models.
\newblock In \emph{Proceedings of ICLR (Tiny Papers)}, 2023.
\newblock arXiv:2303.12748.
\newblock URL: \url{https://openreview.net/forum?id=na1T7ZGYb4}.


\bibitem[Rajendran and LeVine(2019)]{rajendran2019accurate}
Vickram Rajendran and William LeVine.
\newblock Accurate layerwise interpretable competence estimation.
\newblock \emph{Advances in Neural Information Processing Systems}, 32, 2019.
\newblock URL: \url{https://proceedings.neurips.cc/paper_files/paper/2019/file/a11da6bd58b95b334f8cd49f00918f16-Paper.pdf}.

\bibitem[Kull et~al.(2019)]{kull2019beyond}
Meelis Kull, Miquel Perello~Nieto, Markus K\"angsepp, Telmo Silva~Filho, Hao Song, and Peter Flach.
\newblock Beyond temperature scaling: Obtaining well-calibrated multi-class probabilities with {Dirichlet} calibration.
\newblock \emph{Advances in Neural Information Processing Systems}, 32, 2019.
\newblock URL: \url{https://proceedings.neurips.cc/paper_files/paper/2019/file/8ca01ea920679a0fe3728441494041b9-Paper.pdf}.

\bibitem[Minderer et~al.(2021)]{minderer2021revisiting}
Matthias Minderer, Josip Djolonga, Rob Romijnders, Frances Hubis, Xiaohua Zhai, Neil Houlsby, Dustin Tran, and Mario Lucic.
\newblock Revisiting the calibration of modern neural networks.
\newblock \emph{Advances in Neural Information Processing Systems}, 34:15682--15694, 2021.
\newblock URL: \url{https://proceedings.neurips.cc/paper_files/paper/2021/file/8420d359404024567b5aefda1231af24-Paper.pdf}.

\bibitem[Chen et~al.(2024)]{chen2023selfdebug}
Xinyun Chen, Maxwell Lin, Nathanael Sch\"arli, and Denny Zhou.
\newblock Teaching large language models to self-debug.
\newblock In \emph{Proceedings of ICLR}, 2024.
\newblock arXiv:2304.05128.
\newblock URL: \url{https://openreview.net/forum?id=KuPixIqPiq}.

\bibitem[Chen et~al.(2022)]{chen2022codet}
Bei Chen, Fengji Zhang, Anh Nguyen, Daoguang Zan, Zeqi Lin, Jian-Guang Lou, and Weizhu Chen.
\newblock {CodeT}: Code generation with generated tests.
\newblock \emph{arXiv preprint}, 2022.
\newblock arXiv:2207.10397.
\newblock URL: \url{https://arxiv.org/abs/2207.10397}.

\bibitem[Ridnik et~al.(2024)]{ridnik2024alphacodium}
Tal Ridnik, Dedy Kredo, and Itamar Friedman.
\newblock Code generation with {AlphaCodium}: From prompt engineering to flow engineering.
\newblock \emph{arXiv preprint}, 2024.
\newblock arXiv:2401.08500.
\newblock URL: \url{https://arxiv.org/abs/2401.08500}.

\bibitem[Shinn et~al.(2023)]{shinn2023reflexion}
Noah Shinn, Federico Cassano, Ashwin Gopinath, Karthik Narasimhan, and Shunyu Yao.
\newblock Reflexion: Language agents with verbal reinforcement learning.
\newblock In \emph{Proceedings of NeurIPS}, 2023.
\newblock URL: \url{https://proceedings.neurips.cc/paper_files/paper/2023/file/1b44b878bb782e6954cd888628510e90-Paper-Conference.pdf}.

\bibitem[Yao et~al.(2023)]{yao2023tree}
Shunyu Yao, Dian Yu, Jeffrey Zhao, Izhak Shafran, Thomas~L. Griffiths, Yuan Cao, and Karthik Narasimhan.
\newblock Tree of thoughts: Deliberate problem solving with large language models.
\newblock In \emph{Proceedings of NeurIPS}, 2023.
\newblock URL: \url{https://proceedings.neurips.cc/paper_files/paper/2023/file/271db9922b8d1f4dd7aaef84ed5ac703-Paper-Conference.pdf}.

\bibitem[Google DeepMind(2026)]{googledeepmind2026gemma4}
Google DeepMind.
\newblock Gemma 4.
\newblock 2026.
\newblock URL: \url{https://deepmind.google/models/gemma/gemma-4/}.

\bibitem[Qwen Team(2026)]{qwen3.6-27b}
Qwen Team.
\newblock {Qwen3.6-27B}: Flagship-Level Coding in a {27B} Dense Model.
\newblock April 2026.
\newblock URL: \url{https://qwen.ai/blog?id=qwen3.6-27b}.

\bibitem[Huang et~al.(2024)]{huang2024selfcorrect}
Jie Huang, Xinyun Chen, Swaroop Mishra, Huaixiu Steven Zheng, Adams~Wei Yu, Xinying Song, and Denny Zhou.
\newblock Large language models cannot self-correct reasoning yet.
\newblock In \emph{Proceedings of ICLR}, 2024.
\newblock URL: \url{https://openreview.net/forum?id=IkmD3fKBPQ}.

\bibitem[Zhang et~al.(2026)]{zhang2026chasingtail}
Junkai Zhang, Zihao Wang, Lin Gui, Swarnashree Mysore~Sathyendra, Jaehwan Jeong, Victor Veitch, Wei Wang, Yunzhong He, Bing Liu, and Lifeng Jin.
\newblock Chasing the tail: Effective rubric-based reward modeling for large language model post-training.
\newblock In \emph{Proceedings of ICLR}, 2026.
\newblock arXiv:2509.21500.
\newblock URL: \url{https://arxiv.org/abs/2509.21500}.

\bibitem[Kwok et~al.(2026)]{kwok2026llmverifier}
Jacky Kwok, Shulu Li, Pranav Atreya, Yuejiang Liu, Marco Pavone, Ion Stoica, and Azalia Mirhoseini.
\newblock {LLM}-as-a-Verifier: A general-purpose verification framework.
\newblock Project page, 2026.
\newblock URL: \url{https://llm-as-a-verifier.notion.site/}.

\bibitem[Le et~al.(2022)]{le2022coderl}
Hung Le, Yue Wang, Akhilesh~Deepak Gotmare, Silvio Savarese, and Steven~C.H. Hoi.
\newblock {CodeRL}: Mastering code generation through pretrained models and deep reinforcement learning.
\newblock In \emph{Proceedings of NeurIPS}, 2022.
\newblock URL: \url{https://proceedings.neurips.cc/paper_files/paper/2022/file/8636419dea1aa9fbd25fc4248e702da4-Paper-Conference.pdf}.

\bibitem[Ni et~al.(2023)]{ni2023lever}
Ansong Ni, Srini Iyer, Dragomir Radev, Veselin Stoyanov, Wen-tau Yih, Sida~I. Wang, and Xi~Victoria Lin.
\newblock {LEVER}: Learning to verify language-to-code generation with execution.
\newblock In \emph{Proceedings of ICML}, 2023.
\newblock URL: \url{https://proceedings.mlr.press/v202/ni23b.html}.

\bibitem[Li et~al.(2022)]{li2022competition}
Yujia Li et~al.
\newblock Competition-level code generation with {AlphaCode}.
\newblock \emph{Science}, 378(6624):1092--1097, 2022.
\newblock DOI: \url{https://doi.org/10.1126/science.abq1158}.

\bibitem[Zhou et~al.(2023)]{zhou2023language}
Andy Zhou, Kai Yan, Michal Shlapentokh-Rothman, Haohan Wang, and Yu-Xiong Wang.
\newblock Language agent tree search unifies reasoning, acting, and planning in language models.
\newblock arXiv:2310.04406, 2023.
\newblock URL: \url{https://arxiv.org/abs/2310.04406}.

\bibitem[LeVine and Varjavand(2025)]{levine2025rebel}
Will LeVine and Bijan Varjavand.
\newblock Relevance isn't all you need: Scaling {RAG} systems with inference-time compute via multi-criteria reranking.
\newblock arXiv:2504.07104, 2025.
\newblock URL: \url{https://arxiv.org/abs/2504.07104}.

\bibitem[Trivedi et~al.(2024)]{trivedi2024appworld}
Harsh Trivedi, Tushar Khot, Mareike Hartmann, Ruskin Manku, Vinty Dong, Edward Li, Shashank Gupta, Ashish Sabharwal, and Niranjan Balasubramanian.
\newblock {AppWorld}: A controllable world of apps and people for benchmarking interactive coding agents.
\newblock In \emph{Proceedings of ACL}, 2024.
\newblock URL: \url{https://arxiv.org/abs/2407.18901}.

\bibitem[Lightman et~al.(2024)]{lightman2024verify}
Hunter Lightman, Vineet Kosaraju, Yura Burda, Harri Edwards, Bowen Baker, Teddy Lee, Jan Leike, John Schulman, Ilya Sutskever, and Karl Cobbe.
\newblock Let's verify step by step.
\newblock In \emph{Proceedings of ICLR}, 2024.
\newblock arXiv:2305.20050.
\newblock URL: \url{https://openreview.net/forum?id=v8L0pN6EOi}.

\bibitem[Uesato et~al.(2022)]{uesato2022process}
Jonathan Uesato, Nate Kushman, Ramana Kumar, Francis Song, Noah Siegel, Lisa Wang, Antonia Creswell, Geoffrey Irving, and Irina Higgins.
\newblock Solving math word problems with process- and outcome-based feedback.
\newblock arXiv:2211.14275, 2022.
\newblock URL: \url{https://arxiv.org/abs/2211.14275}.

\end{thebibliography}
